\documentclass[twoside]{article}
\usepackage[accepted]{aistats2026}
\usepackage{amsmath, amsthm, amssymb}

\usepackage{expl3}  
\ExplSyntaxOn
_to_str:N
\ExplSyntaxOff
\usepackage{ifthen}
\usepackage{pgffor}

\usepackage[acronym,nowarn,section,nonumberlist]{glossaries}
\setacronymstyle{long-sc-short}
\glsdisablehyper  
\makeglossaries

\setacronymstyle{long-short}

\newacronym{auc}{AUC}{Area Under the Curve}
\newacronym{asr}{ASR}{Automatic Speech Recognition}
\newacronym{ai}{AI}{Artificial Intelligence}

\newacronym{bert}{BERT}{Bidirectional Encoder Representations from Transformers}
\newacronym{bow}{BOW}{Bag of Words}
\newacronym{boe}{BOE}{Bag of Embeddings}
\newacronym{borep}{BOREP}{Bag of Random Embedding Projections}
\newacronym{bn}{BN}{Bayesian Network}
\newacronym{byol}{BYOL}{Bootstrap Your Own Latent}

\newacronym{cdf}{CDF}{Cumulative Distribution Function}
\newacronym{cnn}{CNN}{Convolutional Neural Network}
\newacronym{cka}{CKA}{Centered Kernel Alignment}
\newacronym{ctc}{CTC}{Connectionist Temporal Classification}
\newacronym{cvae}{CVAE}{Conditional Variational Auto-Encoder}
\newacronym{cv}{CV}{Computer Vision}

\newacronym{dag}{DAG}{Directed Acyclic Graph}
\newacronym{ddpg}{DDPG}{Deep Deterministic Policy Gradient}
\newacronym{dino}{DINO}{DIstillation with NO labels}
\newacronym{dgm}{DGM}{Directed Graphical Model}
\newacronym{dpi}{DPI}{Data Processing Inequality}
\newacronym{dnn}{DNN}{Deep Neural Network}

\newacronym{ece}{ECE}{Expected Calibration Error}
\newacronym{ecfp4}{ECFP4}{Extended Connectivity Fingerprint}
\newacronym{ehr}{EHR}{Electronic Health Record}
\newacronym{ema}{EMA}{Exponential Moving Average}
\newacronym{esn}{ESN}{Echo State Network}
\newacronym{esr}{ESR}{Exponential Scaling Rule}
\newacronym{erm}{ERM}{Empirical Risk Minimization }

\newacronym{fft}{FFT}{Fast Fourier Transform}
\newacronym{fid}{FID}{Fr\'echet Inception Distance}
\newacronym{fs}{FS}{FastSent}
\newacronym{ft}{FT}{Fourier Transform}
\newacronym{flop}{FLOP}{Floating Operation}
\newacronym{flops}{FLOPs}{Floating Operations}

\newacronym{gat}{GAT}{Graph Attention Network}
\newacronym{gcn}{GCN}{Graph Convolutional Network}
\newacronym{gd}{GD}{Gradient Descent}
\newacronym{gdl}{GDL}{Geometric Deep Learning}
\newacronym{ggnn}{GGNN}{Gated Graph Neural Network}
\newacronym{glove}{GloVe}{Global Vectors for Word Representation}
\newacronym{gnn}{GNN}{Graph Neural Network}
\newacronym{gqa}{GQA}{Group Query Attention}
\newacronym{gru}{GRU}{Gated Recurrent Unit}

\newacronym{hmm}{HMM}{Hidden Markov Model}

\newacronym{ica}{ICA}{Independent Component Analysis}
\newacronym{idf}{IDF}{Inverse Document Frequency}
\newacronym{ig}{IG}{Information Gain}
\newacronym{iid}{i.i.d.}{Independent and Identically Distributed}
\newacronym{isan}{ISAN}{Input Switched Affine Network}

\newacronym{jsd}{JSD}{Jensen-Shannon Divergence}

\newacronym{kld}{KLD}{Kullback-Leibler Divergence}
\newacronym{knn}{KNN}{$K$-Nearest Neighbour}

\newacronym{lhc}{LHC}{Large Hadron Collider}
\newacronym{lhs}{L.H.S.}{left hand side}
\newacronym{llm}{LLM}{Large Language Model}
\newacronym{lm}{LM}{Language Model}
\newacronym{lstm}{LSTM}{Long Short Term Memory Network}
\newacronym{lsr}{LSR}{Linear Scaling Rule}

\newacronym{map}{MAP}{Maximum a Posteriori}
\newacronym{mc}{MC}{Monte Carlo}
\newacronym{mdl}{MDL}{Minimum Description Length}
\newacronym{ml}{ML}{Machine Learning}
\newacronym{mlp}{MLP}{Multi-Layer Perceptron}
\newacronym{mle}{MLE}{Maximum Likelihood Estimation}
\newacronym{mnist}{MNIST}{Modified National Institute of Standards and Technology}
\newacronym{mi}{MI}{Mutual Information}
\newacronym{mup}{$\mu$P}{Maximal Update Parameterization}

\newacronym{nlp}{NLP}{Natural Language Processing}
\newacronym{nll}{NLL}{Negative Log Likelihood}
\newacronym{nn}{NN}{Neural Network}
\newacronym{ntp}{NTP}{Next Token Prediction}

\newacronym{ode}{ODE}{Ordinary Differential Equation}
\newacronym{oh}{OH}{One-Hot}

\newacronym{pdf}{PDF}{Probability Density Function}
\newacronym{pgm}{PGM}{Probabilistic Graphical Model}
\newacronym{pl}{PL}{Pseudo-Label}

\newacronym{relu}{ReLU}{Rectified Linear Unit}
\newacronym{rdf}{RDF}{Resource Description Framework}
\newacronym{rgb}{RGB}{Red Green Blue}
\newacronym{rgc}{RGC}{Relational Graph Convolution}
\newacronym{rgat}{RGAT}{Relational Graph Attention Network}
\newacronym{rgcn}{RGCN}{Relational Graph Convolutional Network}
\newacronym{rhs}{R.H.S.}{Right Hand Side}
\newacronym{rl}{RL}{Reinforcement Learning}
\newacronym{rnn}{RNN}{Recurrent Neural Network}
\newacronym{roc}{ROC}{Receiver Operating Characteristic}
\newacronym{rope}{RoPE}{Rotary Position Embedding}
\newacronym{rv}{RV}{Random Variable}
\newacronym{rkhs}{RKHS}{Reproducing Kernel Hilbert Space}

\newacronym{sde}{SDE}{Stochastic Differential Equation}
\newacronym{sgd}{SGD}{Stochastic Gradient Descent}
\newacronym{sg}{SG}{SkipGram}
\newacronym{simclr}{SimCLR}{Simple framework for Contrastive Learning of visual Representations}
\newacronym{slsqp}{SLSQP}{Sequential Least SQuares Programming}
\newacronym{ssl}{SSL}{Self-Supervised Learning}
\newacronym{st}{ST}{SkipThought}
\newacronym{sts}{STS}{Semantic Textual Similarity}
\newacronym{swa}{SWA}{Stochastic Weight Averaging}
\newacronym{swag}{SWAG}{SWA-Gaussian}

\newacronym{termfreq}{TF}{Term Frequency}
\newacronym{tfidf}{TF-IDF}{Term Frequency-Inverse Document Frequency}
\newacronym[plural=TPPs, descriptionplural={Temporal Point Processes}]{tpp}{TPP}{Temporal Point Process}

\newacronym{vae}{VAE}{Variational Auto-Encoder}
\newacronym{vit}{ViT}{Vision Transformer}

\newacronym{wer}{WER}{Word Error Rate}
\newacronym{wirgat}{WIRGAT}{Within-Relation Graph Attention}
\newacronym{wl}{WL}{Weisfeiler-Lehman}

\usepackage{bm}
\usepackage{mathtools}

\usepackage{hyperref}
\usepackage{xcolor}
\definecolor{hrefblue}{RGB}{84,151,193}
\definecolor{hrefred}{RGB}{227,94,105}
\definecolor{hrefgreen}{RGB}{83,172,121}
\definecolor{hreforange}{RGB}{250,151,92}
\hypersetup{
	colorlinks   = true, 
	urlcolor     = hrefblue, 
	linkcolor    = hreforange, 
	citecolor   = hrefblue, 
}
\usepackage{graphicx}
\usepackage[capitalize,noabbrev]{cleveref}
\creflabelformat{equation}{#2\textup{#1}#3}

\usepackage{booktabs}       

\usepackage[font=small]{caption}
\usepackage{xcolor}

\usepackage{enumitem}

\usepackage{placeins}
\usepackage{titletoc}
\usepackage[page,header]{appendix}

\usepackage{ifthen}
\usepackage{tikz}

\usepackage{listings}

\usepackage{bm}
\usetikzlibrary{positioning,decorations.pathmorphing,decorations.pathreplacing, fit}
\usepackage{amsfonts}

\usepackage{sidecap}
\usepackage{dashrule}
\usepackage{pdflscape}

\usepackage[textsize=tiny,disable]{todonotes}

\usepackage[utf8]{inputenc} 
\usepackage{csquotes}
\usepackage[UKenglish,USenglish]{babel}
\usepackage[table]{xcolor}
\usepackage{subcaption}

\usepackage{hyperref}       
\usepackage{url}            
\usepackage{booktabs}       
\usepackage{amsfonts}       
\usepackage{nicefrac}       
\usepackage{microtype}      
\usepackage{xcolor}         

\usepackage{xcolor}
\definecolor{AppleWhite}{RGB}{255,255,255}
\definecolor{ApplePrimaryCoolGrey}{RGB}{116,128,139}
\definecolor{AppleCoolGray1}{RGB}{199,209,214}
\definecolor{AppleCoolGray2}{RGB}{147,174,190}
\definecolor{AppleCoolGray3}{RGB}{124,147,160}
\definecolor{AppleCoolGray4}{RGB}{92,102,109}
\definecolor{AppleCoolGray5}{RGB}{78,93,100}
\definecolor{AppleCoolGray6}{RGB}{53,60,65}
\definecolor{AppleBlack}{RGB}{0,0,0}
\definecolor{AppleSecondaryChartGray}{RGB}{168,168,168}
\definecolor{AppleChartGrey2}{RGB}{233,233,233}
\definecolor{AppleChartGrey3}{RGB}{211,211,211}
\definecolor{AppleChartGrey4}{RGB}{190,190,190}
\definecolor{AppleChartGrey5}{RGB}{140,140,140}
\definecolor{AppleChartGrey6}{RGB}{102,102,102}
\definecolor{AppleChartGrey7}{RGB}{64,64,64}
\definecolor{ApplePrimaryChartBlue}{RGB}{84,151,193}
\definecolor{AppleBlue2}{RGB}{212,229,239}
\definecolor{AppleBlue3}{RGB}{169,202,223}
\definecolor{AppleBlue4}{RGB}{127,177,209}
\definecolor{AppleBlue5}{RGB}{71,130,166}
\definecolor{AppleBlue6}{RGB}{55,99,128}
\definecolor{AppleBlue7}{RGB}{45,72,89}
\definecolor{ApplePrimaryChartGreen}{RGB}{83,172,121}
\definecolor{AppleGreen2}{RGB}{212,234,221}
\definecolor{AppleGreen3}{RGB}{169,213,188}
\definecolor{AppleGreen4}{RGB}{126,193,155}
\definecolor{AppleGreen5}{RGB}{58,140,82}
\definecolor{AppleGreen6}{RGB}{39,102,54}
\definecolor{AppleGreen7}{RGB}{29,58,31}
\definecolor{ApplePrimaryChartYellow}{RGB}{253,195,93}
\definecolor{AppleYellow2}{RGB}{254,240,214}
\definecolor{AppleYellow3}{RGB}{254,224,174}
\definecolor{AppleYellow4}{RGB}{254,210,134}
\definecolor{AppleYellow5}{RGB}{230,168,69}
\definecolor{AppleYellow6}{RGB}{191,131,46}
\definecolor{AppleYellow7}{RGB}{153,107,54}
\definecolor{ApplePrimaryChartOrange}{RGB}{250,151,92}
\definecolor{AppleOrange2}{RGB}{254,229,214}
\definecolor{AppleOrange3}{RGB}{252,203,173}
\definecolor{AppleOrange4}{RGB}{252,178,133}
\definecolor{AppleOrange5}{RGB}{227,121,68}
\definecolor{AppleOrange6}{RGB}{191,87,46}
\definecolor{AppleOrange7}{RGB}{143,59,36}
\definecolor{ApplePrimaryChartRed}{RGB}{227,94,105}
\definecolor{AppleRed2}{RGB}{248,215,217}
\definecolor{AppleRed3}{RGB}{241,174,180}
\definecolor{AppleRed4}{RGB}{234,135,143}
\definecolor{AppleRed5}{RGB}{196,63,77}
\definecolor{AppleRed6}{RGB}{153,35,53}
\definecolor{AppleRed7}{RGB}{102,19,43}
\definecolor{ApplePrimaryChartPurple}{RGB}{161,150,204}
\definecolor{ApplePurple2}{RGB}{231,228,242}
\definecolor{ApplePurple3}{RGB}{208,202,229}
\definecolor{ApplePurple4}{RGB}{185,176,217}
\definecolor{ApplePurple5}{RGB}{128,113,171}
\definecolor{ApplePurple6}{RGB}{89,76,128}
\definecolor{ApplePurple7}{RGB}{62,46,101}
\definecolor{AppleCoolGrey}{RGB}{116,128,139}
\definecolor{AppleChartGray}{RGB}{168,168,168}
\definecolor{AppleBlue}{RGB}{84,151,193}
\definecolor{AppleGreen}{RGB}{83,172,121}
\definecolor{AppleYellow}{RGB}{253,195,93}
\definecolor{AppleOrange}{RGB}{250,151,92}
\definecolor{AppleRed}{RGB}{227,94,105}
\definecolor{ApplePurple}{RGB}{161,150,204}

\DeclareRobustCommand{\mathup}[1]{\begingroup\changegreek\mathrm{#1}\endgroup}
\DeclareRobustCommand{\mathbfup}[1]{\begingroup\changegreekbf\mathbf{#1}\endgroup}
\DeclareRobustCommand{\mathbit}[1]{\bm{\mathit{#1}}}

\DeclareMathAlphabet{\mathsfit}{\encodingdefault}{\sfdefault}{m}{sl}
\SetMathAlphabet{\mathsfit}{bold}{\encodingdefault}{\sfdefault}{bx}{n}
\newcommand{\tens}[1]{\bm{\mathsfit{#1}}}

\newcommand{\constantvector}{\bm}               

\newcommand{\constantmatrix}{\bm}               
\newcommand{\constantmatrixgreek}{\mathbit}

\newcommand{\randomscalar}{\textnormal}         
\newcommand{\randomscalargreek}{\mathup}

\newcommand{\randomvector}{\mathbf}             
\newcommand{\randomvectorgreek}{\mathbfup}

\newcommand{\randommatrix}{\mathbf}             
\newcommand{\randommatrixgreek}{\mathbfup}

\newcommand{\graphstyle}{\mathcal}              

\newcommand{\tensorstyle}{\tens}                

\newcommand{\setstyle}{\mathbb}                

\def\alphabet{a,b,c,d,e,f,g,h,i,j,k,l,m,n,o,p,q,r,s,t,u,v,w,x,y,z}
\def\Alphabet{A,B,C,D,E,F,G,H,I,J,K,L,M,M,O,P,Q,R,S,T,U,V,W,X,Y,Z}
\def\greekalphabet{alpha,beta,gamma,delta,epsilon,varepsilon,zeta,eta,theta,vartheta,iota,kappa,varkappa,lambda,mu,nu,xi,pi,varpi,rho,varrho,sigma,varsigma,tau,upsilon,phi,varphi,chi,psi,omega}
\def\GreekAlphabet{Gamma,Delta,Theta,Lambda,Xi,Pi,Sigma,Upsilon,Phi,Psi,Omega}

\makeatletter
\def\changegreek{\@for\next:=\greekalphabet
	\do{\expandafter\let\csname\next\expandafter\endcsname\csname\next up\endcsname}}
\def\changegreekbf{\@for\next:=\greekalphabet
	\do{\expandafter\def\csname\next\expandafter\endcsname\expandafter{%
			\expandafter\bm\expandafter{\csname\next up\endcsname}}}}
\makeatother

\foreach \x in \alphabet {
	\expandafter\xdef\csname v\x\endcsname{\noexpand\ensuremath{\noexpand\constantvector{\x}}}

	\expandafter\xdef\csname ev\x\endcsname{\noexpand\ensuremath{\noexpand\x}}

	\ifthenelse{\equal{\x}{m}}{}{
		\expandafter\xdef\csname r\x\endcsname{\noexpand\ensuremath{\noexpand\randomscalar{\x}}}
	}

	\expandafter\xdef\csname rv\x\endcsname{\noexpand\ensuremath{\noexpand\randomvector{\x}}}
}

\foreach \x in \greekalphabet {
	\expandafter\xdef\csname v\x\endcsname{\noexpand\ensuremath{\noexpand\constantvector{\csname \x\endcsname}}}

	\expandafter\xdef\csname ev\x\endcsname{\noexpand\ensuremath{\noexpand{\csname \x \endcsname}}}

	\expandafter\xdef\csname r\x\endcsname{\noexpand\ensuremath{\noexpand\randomscalargreek{\csname \x\endcsname}}}

	\expandafter\xdef\csname rv\x\endcsname{\noexpand\ensuremath{\noexpand\randomvectorgreek{\csname \x\endcsname}}}
}

\foreach \x in \Alphabet {
	\expandafter\xdef\csname m\x\endcsname{\noexpand\ensuremath{\noexpand\constantmatrix{\x}}}

	\expandafter\xdef\csname em\x\endcsname{\noexpand\ensuremath{\noexpand\x}}

	\expandafter\xdef\csname rm\x\endcsname{\noexpand\ensuremath{\noexpand\randommatrix{\x}}}

	\expandafter\xdef\csname t\x\endcsname{\noexpand\ensuremath{\noexpand\tensorstyle{\x}}}

	\expandafter\xdef\csname g\x\endcsname{\noexpand\ensuremath{\noexpand\graphstyle{\x}}}

	\ifthenelse{\equal{\x}{E}}{}{
		\expandafter\xdef\csname s\x\endcsname{\noexpand\ensuremath{\noexpand\setstyle{\x}}}
	}

}

\foreach \x in \GreekAlphabet {
	\expandafter\xdef\csname m\x\endcsname{\noexpand\ensuremath{\noexpand\constantmatrixgreek{\csname \x\endcsname}}}

	\expandafter\xdef\csname rm\x\endcsname{\noexpand\ensuremath{\noexpand\randommatrixgreek{\csname \x\endcsname}}}
}

\DeclareMathOperator*{\argmin}{arg\,min}

\DeclareRobustCommand{\wass}[3]{\ensuremath{\mathcal{W}_{#1}\left(#2\;,#3\right)}}

\newcommand{\E}{\mathbb{E}}
\newcommand{\X}{\mathcal{X}}
\newcommand{\Ell}{\mathcal{L}}
\newcommand{\G}{\mathcal{G}}
\newcommand{\Sc}{\mathcal{S}}
\newcommand{\cH}{\mathcal{H}}
\newcommand{\R}{\mathbb{R}}

\newtheorem{definition}{Definition}[section]

\newtheorem{theorem}{Theorem}[section]
\newtheorem{corollary}{Corollary}[theorem]
\newtheorem{lemma}[theorem]{Lemma}

\newtheorem{assumption}{Assumption}[section]

\newtheorem{manthmin}{Theorem}
\newenvironment{manualthm}[1]{%
	\renewcommand\themanthmin{#1}%
	\manthmin
}{\endmanthmin}

\newtheorem{manlemmin}{Lemma}
\newenvironment{manuallemma}[1]{%
	\renewcommand\themanlemmin{#1}%
	\manlemmin
}{\endmanlemmin}

\newtheorem{mancorin}{Corollary}
\newenvironment{manualcor}[1]{%
	\renewcommand\themancorin{#1}%
	\mancorin
}{\endmancorin}

\newcommand{\indep}{\perp \!\!\! \perp}

\newcommand{\tr}{\textnormal{tr}}

\newcommand{\Supp}{\mathrm{Supp}}

\newcommand{\Y}{\mathcal{Y}}

\usepackage[round]{natbib}
\renewcommand{\bibname}{References}
\renewcommand{\bibsection}{\subsubsection*{\bibname}}

\begin{document}

%
\runningtitle{Beyond Real Data}

%
\runningauthor{Amitis Shidani, Tyler Farghly, Yang Sun, Habib Ganjgahi, George Deligiannidis}
\twocolumn[

\aistatstitle{Beyond Real Data: \\ Synthetic Data Through The Lens Of Regularization}
\vspace{-0.5cm}
\aistatsauthor{
  Amitis Shidani$^{1,2}$ \And
  Tyler Farghly$^{2}$  \AND
  Yang Sun$^{3}$  \And
  Habib Ganjgahi$^{3,\dagger}$  \And
  George Deligiannidis$^{2, \dagger}$ \AND
}

\aistatsaddress{$^1$Apple \qquad $^2$University of Oxford \qquad $^3$Big Data Institute\\ $^\dagger$ Joint senior supervision.\\ Correspondence to: amitis\_shidani@apple.com} 
]

\begin{abstract}
  Synthetic data can improve generalization when real data is scarce, but excessive reliance may introduce distributional mismatches that degrade performance. In this paper, we present a learning-theoretic framework to quantify the trade-off between synthetic and real data. Our approach leverages algorithmic stability to derive generalization error bounds, characterizing the optimal synthetic-to-real data ratio that minimizes expected test error as a function of the Wasserstein distance between the real and synthetic distributions. We motivate our framework in the setting of kernel ridge regression with mixed data, offering a detailed analysis that may be of independent interest. Our theory predicts the existence of an optimal ratio, leading to a U-shaped behavior of test error with respect to the proportion of synthetic data. Empirically, we validate this prediction on CIFAR-10 and a clinical brain MRI dataset. Our theory extends to the important scenario of domain adaptation, showing that carefully blending synthetic target data with limited source data can mitigate domain shift and enhance generalization. We conclude with practical guidance for applying our results to both in-domain and out-of-domain scenarios. \looseness=-1
\end{abstract}

\section{INTRODUCTION}

\looseness=-1The success of modern \gls{ml} and \gls{ai} heavily depends on the availability of large-scale training datasets \citep{sun2017revisiting, radford2021learning}. However, in many critical domains such as healthcare, data collection is often prohibitively expensive, time-consuming, or constrained by privacy concerns \citep{esteva2019guide, kaissis2021secure}. Similar challenges arise in scientific domains where obtaining labeled data requires high-fidelity physical simulations or specialized experimental setups. For instance, generating data in molecular dynamics \citep{hollingsworth2018molecular,hansson2002molecular} often demands significant computational resources, or structural biology techniques like cryo-electron microscopy \citep{MURATA2018324,milne2013cryo} involve costly and complex instrumentation. In these scenarios, \gls{ml} models are trained on small datasets and as a result frequently suffer from poor generalization, limiting their practical applicability \citep{recht2019imagenet,maleki2022generalizability,NEURIPS2018_f708f064,brigato2021close}.

To address this challenge, several strategies have been proposed, including data augmentation \citep{shorten2019survey, cubuk2020randaugment} and the use of synthetic data \citep{fridadar2018gan, karras2020analyzing, lu2023machine, jordon2022synthetic}. Although these methods can improve model accuracy, their success depends critically on how well the synthetic data approximates the real data distribution \citep{bowles2018gan}. With the emergence of powerful generative models such as diffusion models \citep{ho2020denoising, song2021score,lipman2022flow,NEURIPS2021_940392f5}, there is renewed interest in using synthetic data to supplement limited real data \citep{trabucco2023effective, voetman2023big, alemohammad2024self}. Empirical evidence suggests that, when properly generated, synthetic data can substantially boost the downstream model performance in low-data regimes \citep{azizi2023synthetic,feng2024beyond}.\looseness=-1


However, the integration of synthetic data introduces a critical trade-off as synthetic data may deviate from the true data distribution. If the synthetic dataset grows disproportionately large, the training algorithm may overlook the real data, introducing bias \citep{DBLP:conf/iclr/AlemohammadCLHB24, DBLP:journals/corr/abs-2311-16822, betzalel2022study,dohmatob2025strong,DBLP:conf/iclr/BertrandBDJG24}. See \Cref{sec:expanded-related} for an extensive literature review on this topic. This issue motivates a central question:
\vspace{-0.5mm}
\begin{center}
    \textit{"What is the optimal balance between real and synthetic data to minimize generalization error?"}
\end{center}

In this work, we address this question from a learning-theoretic perspective, establishing that an optimal ratio of synthetic to real data exists for maximizing generalization performance. In \Cref{app:traditional-bound}, we show that the traditional formalization of the problem fails to capture this trade-off, yielding a loose bound whose optimum lies at either excluding synthetic data entirely or using it without limit. Since this behaviour does not align with empirical observations (\Cref{subsec:brain-experiments}), we propose a modified formalization that more accurately reflects the practical balance. We first motivate our analysis through a simple yet insightful case study in kernel ridge regression \citep{singh2023kernel}, which may be of independent interest. We then extend our theoretical framework to more general settings, deriving generalization bounds via stability analysis \citep{DBLP:journals/jmlr/BousquetE02, DBLP:books/daglib/0033642, hardt2016train}. Our theoretical insights are empirically validated on two distinct datasets: CIFAR-10 (a standard benchmark) \citep{krizhevsky2009learning} and a real-world brain imaging dataset for Multiple Sclerosis (MS) \citep{DBLP:journals/neuroimage/CarassRJCSGBNPS17}.


Furthermore, we extend our framework to domain adaptation settings \citep{ben2010theory, ganin2016domain, wilson2020survey}, where synthetic data from a target domain is used to enhance limited real data from a source domain. This broadens the scope of our approach, highlighting its relevance for data-scarce scenarios in diverse \gls{ml} applications like healthcare \citep{zhuang2021comprehensive}.
\vspace{-0.25cm}
\paragraph{Contribution}
Our main contributions are as follows:
\vspace{-0.65cm}
\begin{itemize}[leftmargin=0.75cm]
    \item We provide a learning-theoretic analysis demonstrating the existence of an optimal ratio between synthetic and real data that minimizes generalization error. Our approach is grounded in stability-based generalization bounds and is first illustrated through a tractable kernel ridge regression model. See \Cref{sec:kernel-ridge,sec:generalization}.
    
    \item We empirically validate our theoretical predictions using both benchmark (CIFAR-10, \Cref{app:cifar-experiments}) and real-world (brain MRI for Multiple Sclerosis, \Cref{subsec:brain-experiments}) datasets, confirming that an appropriate balance of synthetic data improves performance in low-data regimes.
    
    \item We extend our framework to domain adaptation, showing how synthetic data from a target domain can be effectively combined with limited real data from a source domain, thereby broadening the applicability of our results (\Cref{sec:domain-shift}). We also provide practical guidance for applying our theory to both in-domain and out-of-domain generalization tasks (\Cref{sec:practical}).
\end{itemize}



\section{MOTIVATION: SYNTHETIC DATA IN KERNEL REGRESSION}
\label{sec:kernel-ridge}

We study the effect of incorporating synthetic data into kernel regression \citep{singh2023kernel, allerbo2023solving, JMLR:v23:21-0570, SMALE2005285} as a simple yet illustrative setting to gain insight into the key factors influencing the generalization bound, i.e., when synthetic data improves or degrades generalization.


We consider kernel regression, where a function is learned by minimizing a regularized empirical risk over a \gls{rkhs} denoted by $\mathcal{H}_K$. Given training data $\{(\rx_n, \ry_n)\}_{n = 1}^N$ with $\rx_n \in \mathcal{X} \sim p_\rvx$ and $\ry_n \in \R$, the objective is to find a function $f \in \mathcal{H}_K$ that best fits the data while controlling complexity through a regularization term. We assume that $\ry_n = f_\star(\rx_n) + \varepsilon_n$, where $\varepsilon_n$ are \gls{iid} samples from a zero-mean Gaussian distribution with variance $\sigma^2$. The \gls{erm} normally takes the following form:\looseness=-1
\vspace{-0.25cm}
\begin{align*}
    f_N = \argmin_{f \in \mathcal{H}_k}  \sum_{n = 1}^N \left(\ry_n - f(\rx_n)\right)^2 + \sum_{m=1}^M \left(\tilde{\ry}_m - f(\tilde{\rx}_m)\right)^2\,.
\end{align*}
As shown in \Cref{app:traditional-bound}, classical generalization arguments yield a loose bound on the test error of this formulation. The looseness arises mainly from the sample noise of synthetic data, which complicates the analysis. To address this, we approximate it with the following ansatz, where, unlike the standard setup, we regularize towards a synthetic data generator \(g \in \mathcal{H}_K\), effectively corresponding to the case of having an infinite number of synthetic samples:
\begin{align}\label{eq:kernel-reg}
\hspace{-0.32cm}
    f_N = \argmin_{f \in \mathcal{H}_k} \frac{1-\tilde{\lambda}}{N} \sum_{n = 1}^N \left(\ry_n - f(\rx_n)\right)^2 + \tilde{\lambda} \|f - g\|^2
\end{align}
where \(\tilde{\lambda} > 0\) is the regularization strength. This perspective allows us to derive tighter bounds that align with empirical behavior. See \Cref{app:ansantz,sec:infinite-data-kernel} for a detailed analysis of the asymptotics and the connection to the finite-sample formulation.

By the \emph{Representer Theorem} \citep{kimeldorf1971some,DBLP:conf/colt/ScholkopfHS01}, the learned function takes the form \(f_N(\rx) = \sum_{n=1}^{N} \alpha_n K(\rx, \rx_n)\), 
where \(K\) is a positive definite kernel function, and \( \alpha_n \) are coefficients obtained from a regularized least squares problem. Let \(\Tilde{\mathcal{H}} = \text{span}\{K(\cdot, \rx_1), \ldots, K(\cdot, \rx_N)\}\), and 
\(\mathcal{H}_K = \Tilde{\mathcal{H}} \oplus \Tilde{\mathcal{H}}^\perp\), where \( \Tilde{\mathcal{H}}^\perp \) is the orthogonal complement in \( \Tilde{\mathcal{H}} \). By the Representer Theorem, the synthetic data generator \( g \in \mathcal{H}_K \) can be written as $g(\rx) = \sum_{n = 1}^N \beta_n K(\rx, \rx_n) + g_\perp(\rx)$, where $g_\perp \in \Tilde{\mathcal{H}}^\perp$. Setting \( g = 0 \) recovers the standard kernel ridge regression. We establish the following lemma (proof in \Cref{app:proof-kernel-reg}), which characterizes the solution to \Cref{eq:kernel-reg}. Let $\lambda := \tilde{\lambda}/(1-\tilde{\lambda})$.\looseness=-1

\begin{lemma}\label{lem:kernel-reg}
    Let \( K_N \in \R^{N \times N} \) be the empirical kernel matrix with entries \( (K_N)_{ij} = K(\rx_i, \rx_j) \). Define the integral operator \( T_K : L^2(p_x) \to L^2(p_x) \) by \((T_K f)(\rx) = \int K(\rx, x') f(x')\, dp_x(x') = \E_{\rx'}\left[K(\rx, \rx') f(\rx')\right]\).
    Let \( \lambda_N = N \lambda\). Then the solution to \Cref{eq:kernel-reg} has the closed-form representation:
    \vspace{-0.15cm}
    \begin{align*}
        \textstyle \boldsymbol{\alpha} = \left(K_N + \lambda_N I\right)^{-1} \left(K_N \boldsymbol{\alpha}_\star + \lambda_N \boldsymbol{\beta} + \boldsymbol{\varepsilon}\right)\,,
    \end{align*}
    where \( \boldsymbol{\alpha}_\star \), \( \boldsymbol{\beta} \), and \( \boldsymbol{\varepsilon} \) are the coefficients of \( f_\star \), \( g \), and the noise vector in the training basis. 
\end{lemma}
We now recall the Mercer decomposition \citep{Mercer:1909dea} of the kernel. The operator \( T_K \) defined in \Cref{lem:kernel-reg} is compact, self-adjoint, and positive semi-definite, and thus admits a spectral decomposition. That is, there exist eigenfunctions \( \{\phi_j\}_{j=1}^\infty \) forming an orthonormal basis of \( L^2(p_x) \) and corresponding non-negative eigenvalues \( \mu_1 \geq \mu_2 \geq \cdots \to 0 \) such that \(T_K \phi_j = \mu_j \phi_j\). The eigenfunctions \( \phi_j \) can be interpreted as the natural coordinates of the function space with respect to the kernel, and the eigenvalues \( \mu_j \) encode their relative importance. In this basis, we can write
\vspace{-0.22cm}
\begin{align}\label{eq:mercer}
    &f_\star = \sum_{j = 1}^\infty \theta_j \phi_j\,, \qquad
    g = \sum_{j = 1}^\infty \omega_j \phi_j\,,
\end{align}
where $\theta_j = \langle f_\star, \phi_j \rangle$, and $\omega_j = \langle g, \phi_j \rangle$.
\begin{assumption}[Polynomial eigendecay and smoothness]\label{ass:eigendecay}
    We assume the kernel \( K \) exhibits \( 2r \)-polynomial eigendecay for some \( r \geq \frac{1}{2} \). Given the expansions in \Cref{eq:mercer}, we assume for some \( s, s' > 0 \):
    \vspace{-0.15cm}
    \begin{center}
        \textit{(a)} \( \theta_j^2 \asymp \mu_j^s \asymp j^{-2rs} \), \textit{(b)} \( \omega_j^2 \asymp \mu_j^{s'} \asymp j^{-2rs'} \).
    \end{center}
\end{assumption}
\looseness=-1 \Cref{ass:eigendecay} quantifies how well \( f_\star \) and \( g \) align with the eigenfunctions of \( T_K \). It ensures 
that \( \sum_{j} \theta_j^2/\mu_j^s < \infty\) and \(\sum_{j} \omega_j^2/ \mu_j^{s'} < \infty\), i.e.\ \(f_\star\) and \(g\) decay sufficiently fast in the eigenbasis; larger rate corresponds to greater smoothness. Such assumptions are standard in kernel regression analysis; see, e.g., \cite{DBLP:journals/corr/abs-2410-17796, DBLP:journals/corr/abs-1906-11300, DBLP:journals/corr/abs-2105-15004, DBLP:conf/icml/BarzilaiS24}.

\begin{figure}[t]
    \centering
     \begin{subfigure}[b]{0.5\textwidth}
         \centering
         \includegraphics[width=0.8\textwidth]{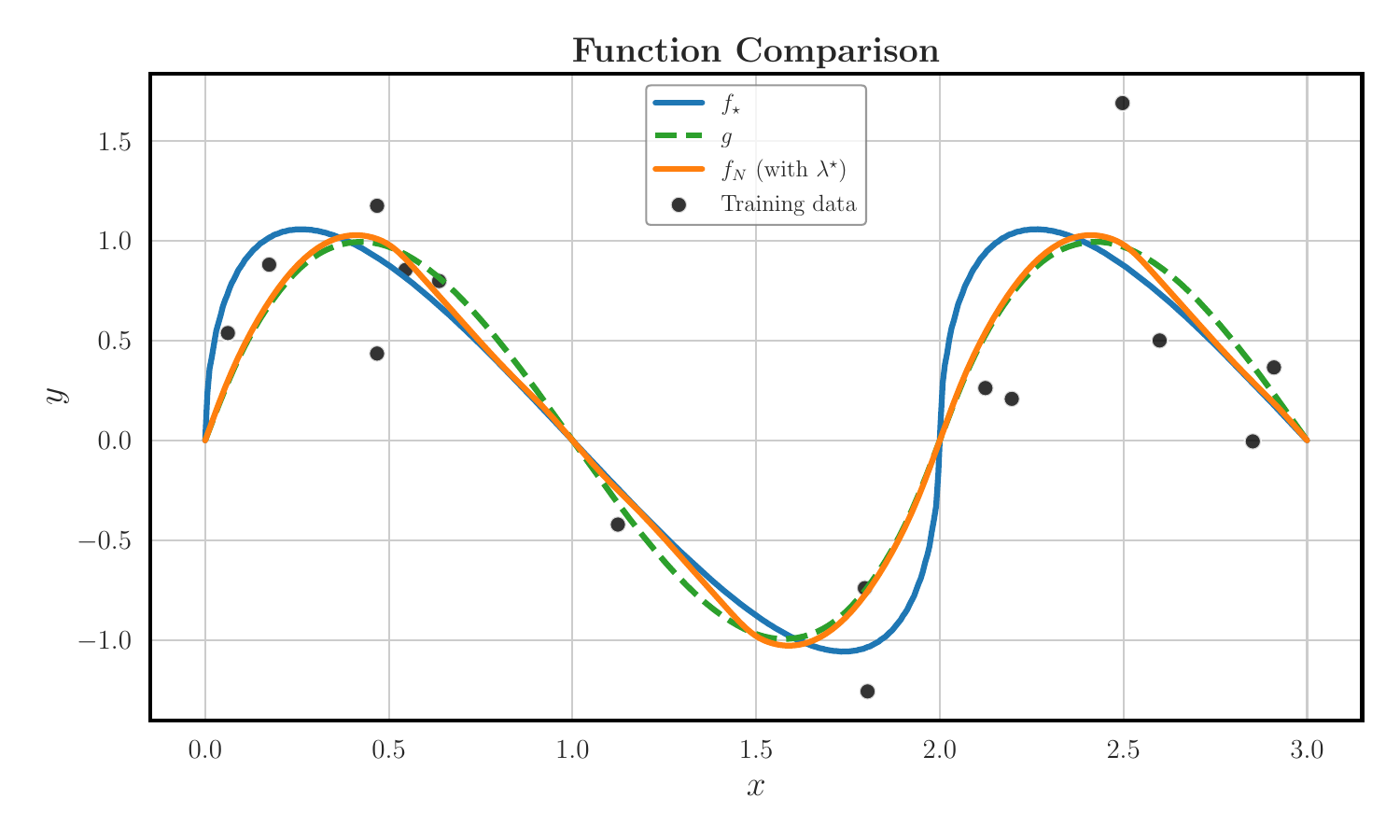}
         \caption{Function comparisons.}
         \label{fig:function-comparison}
     \end{subfigure}
     \hfill
     \begin{subfigure}[b]{0.5\textwidth}
         \centering
         \includegraphics[width=0.8\textwidth]{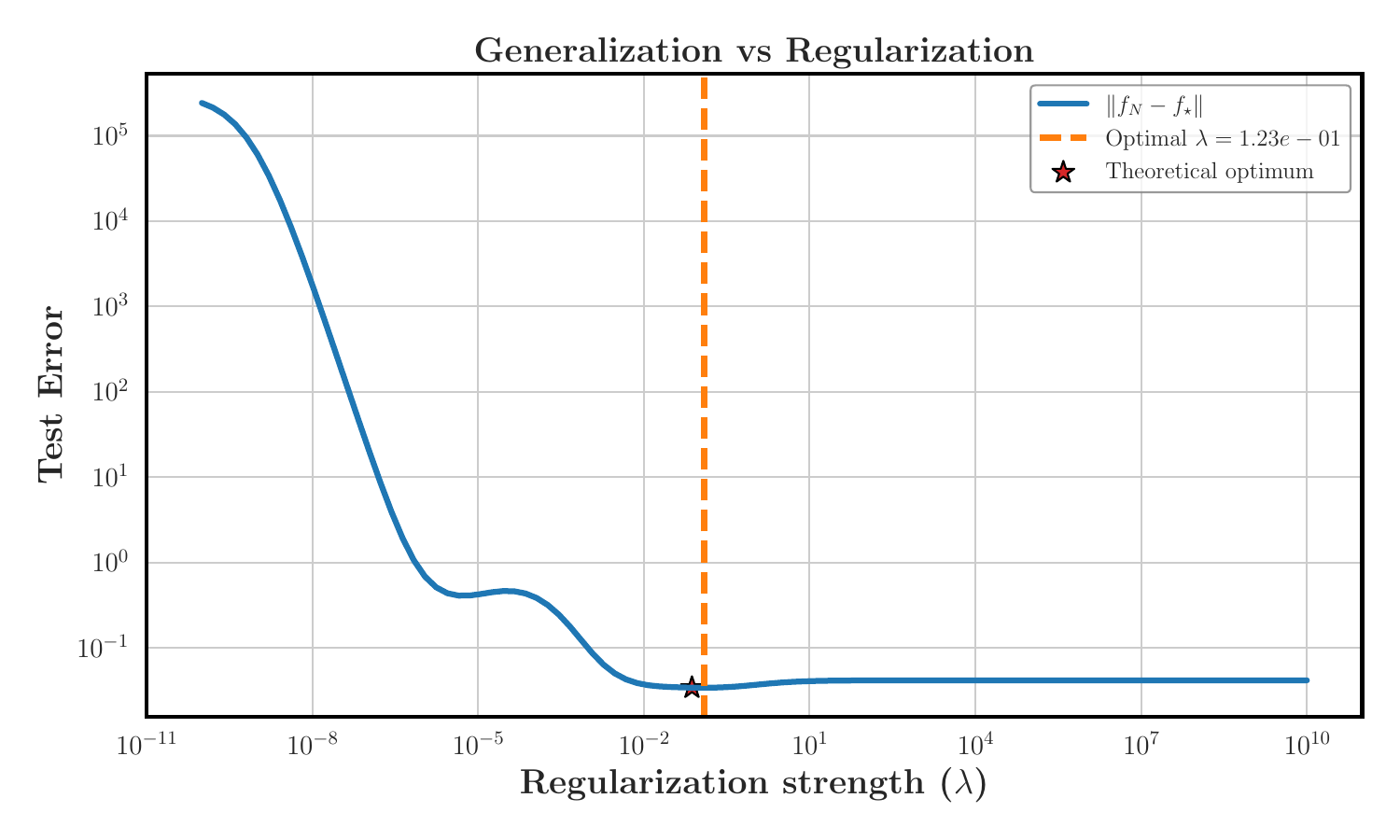}
         \caption{Generalization vs regularization.}
         \label{fig:generalization-vs-regularization}
     \end{subfigure}
    \caption{
        \emph{(a)} Comparison of the true function $f_\star$ (blue), the synthetic generator $g$ (green), and the learned estimator $f_N$ (orange), obtained via \Cref{lem:kernel-reg}, with parameters $r = 2.0$, $s = 0.8$, and $s' = 1.5$. \emph{(b)} Prediction error $|f_N - f_\star|_{L_2}$ as a function of the regularization strength $\lambda$. The U-shaped curve attains its minimum at $\lambda^\star$ (orange dashed line), which closely matches the theoretical optimum (star marker).
    }
    \label{fig:lambda_opt}
\end{figure}

\begin{definition}[Bias-Variance Decomposition] \label{def:bias-variance}
Define the test error $\mathcal{R}_N(\lambda; g)$ to be the population mean squared error between the regressor and the true label averaged over noise:
\vspace{-0.15cm}
\begin{align*}
    \mathcal{R}_N(\lambda; g) = \E_{\rvx, \varepsilon}\left[ (f_\star(\rvx) - f_N(\rvx))^2\right]\,.
\end{align*}
We decompose the test error into a bias $\mathcal{B}$ and variance $\mathcal{V}$, with $\mathcal{R}_N(\lambda; g) = \mathcal{B}^2 + \mathcal{V}$, such that:
\begin{align*}
    &\mathcal{B}^2 = \E_\rvx \left[f_\star(\rvx) - \E_\varepsilon\left[f_N(\rvx)\right]\right]^2\,, \\
    &\mathcal{V} = \E_{\rvx, \varepsilon} \left[(f_N(\rvx) - \E_\varepsilon\left[f_N(\rvx)\right])^2\right]\,.
\end{align*}
\end{definition}
We now present a bias–variance decomposition of \Cref{eq:kernel-reg}, along with a corollary characterizing the optimal number of synthetic samples. Proofs are in \Cref{app:kernel-error-rate,app:kernel-optimal}.
\begin{theorem}[Generalization Error Bound]\label{thm:kernel-error-rate}
    Under \Cref{ass:eigendecay}, for the kernel regression problem defined in \Cref{eq:kernel-reg} and any fixed regularization parameter \( \lambda > 0 \), the test error admits the bound:
    \vspace{-0.15cm}
    \begin{align*}
        \mathcal{R}_N(\lambda; g) = \mathcal{O}\left(\frac{\mathcal{D}(f_\star, g) + \sigma^2}{N\lambda^2} + \lambda^{2 - \frac{1}{4r}} \mathcal{D}(f_\star, g)\right)\,,
    \end{align*}
    where $\lambda = \tilde{\lambda} / (1-\tilde{\lambda})$, and \(\mathcal{D}(f_\star, g)^2 = \sum_{j = 1}^\infty \frac{1}{\mu_j^2}(\theta_j - \omega_j)^2\) denotes the discrepancy between the target function \( f_\star \) and the synthetic generator \( g \).
\end{theorem}
\begin{corollary}[Optimal Regularization and Synthetic Sample Size]\label{cor:kernel-opt-lambda}
    Under the assumptions of \Cref{thm:kernel-error-rate}, the optimal regularization parameter that minimizes the test error is given by
    \vspace{-0.25cm}
    \begin{align*}
        \lambda^\star \asymp \left(\frac{\mathcal{D}(f_\star, g) + \sigma^2}{N \mathcal{D}(f_\star, g)}\right)^{\frac{4r}{16r+1}}\,.
    \end{align*}
    Setting \(\lambda = \frac{M}{N} \), i.e., $\tilde{\lambda} = \frac{M}{N + M}$ (see \Cref{app:ansantz}), the optimal number of synthetic samples satisfies:
    \vspace{-0.15cm}
    \begin{align*}
        M^\star \asymp  \left(1 + \frac{\sigma^2}{\mathcal{D}(f_\star, g)}\right)^{\frac{4r}{16r+1}} N^{\frac{12r + 1}{16r + 1}}\,.
    \end{align*}
\end{corollary}
\looseness=-1 We empirically validate our theory in \Cref{fig:lambda_opt}, observing a U-curve as predicted by \Cref{thm:kernel-error-rate}, with error minimized near the theoretical $\lambda^*$. See \Cref{app:kernel-experiments} for details.

\section{GENERALIZATION ERROR}
\label{sec:generalization}
We begin by introducing the notation and formal setting used throughout the remainder of the paper. We consider learning on a separable complete metric space $(\X, d_\X)$. We define the sample space $\Sc_N = \X^N$ 
and the random training dataset of $N$ \gls{iid} samples from $p_\rx$ over $\X$ is denoted by $\rmS = \{\rx_1, \dots, \rx_N\} \in \Sc_N$, with joint law $p_\rmS$. 
Consider some measurable hypothesis space $\cH$ and a loss function $\ell: \cH \times \X \to \R$ that quantifies the performance of a hypothesis, and we assume $\ell(h, \cdot) \in L^1(p_\rx)$ for each $h \in \cH$. We define the empirical and population risks as,
\vspace{-0.15cm}
\begin{align*}
     \Ell_\rmS(h) = \frac{1}{N} \sum_{i=1}^N \ell(h, \rx_i), \quad
    \Ell_\X(h) = r(h) = \E_{p_\rx}[\ell(h, \rx)].
\end{align*}
For $r \in [1, \infty)$, the Wasserstein $r$-distance between two probability measures $p$ and $q$ on $\X$ with finite $r$-moments is defined as $\wass{r}{p}{q} = \inf_{\gamma \in \Gamma(p, q)}( \E_{(x, y) \sim \gamma} [d(x, y)^r])^{1/r}$, where $\Gamma(p, q)$ is the set of all couplings of $p$ and $q$. See \Cref{app:notation} for more detailed notation.

\subsection{Synthetic Data As A Regularizer}
Following the motivation in \Cref{sec:kernel-ridge}, we consider training on a mixture of real and synthetic data, where the synthetic data acts as a form of regularization. We focus on the following mixed loss:
\begin{align*}
    \gR_{\lambda}(h, \rmS) = (1-\lambda) \Ell_\rmS(h) + \lambda \E_{\rvx \sim p'_{\rvx}}\left[ \ell(h, \rvx)\right],
\end{align*}
where $p'_{\rvx}$ denotes the distribution of synthetic data, which may differ from the real distribution $p_{\rvx}$. We are interested in upper-bounding the generalization error of the algorithm that minimizes the mixed-loss. Our approach leverages a strategy from the learning theory literature known as algorithmic stability.
\begin{definition}[Uniform Stability]\label{def:algo-stable}
    Let $\gA: \gS \mapsto \gH$ denote an algorithm. Algorithm $\gA$ is $\varepsilon$-uniformly stable if for all $\rmS, \rmS' \in \gX^N$ such that $\rmS, \rmS'$ differ in at most one example, the corresponding outputs $\gA(\rmS)$ and $\gA(\rmS')$ satisfy $\sup_{\rvx \in \gX} |\ell(\gA(\rmS); \rvx) - \ell(\gA(\rmS');\rvx)| \leq \varepsilon$.
\end{definition}

This notion of algorithmic stability captures sensitivity of an algorithm on individual changes in the dataset. Under this property, it has been shown that generalization gap bounds in both expectation and high probability can be obtained \citep{DBLP:conf/nips/BousquetE00, DBLP:journals/jmlr/BousquetE02}. In our analysis we consider the general case where \(\cH\) consists of set of functions between \(\X\) and some metric space \(\Y\). We make the assumption that it is a compact subset \(L^\infty(p_\rx)\).
\begin{assumption}\label{ass:diam_H}
The hypothesis class \(\mathcal{H}\) is a set of measurable functions of the form \(\X \to \Y\) that are \(L\)-Lipschitz and, for some \(D > 0\) satisfies \(\|h-h'\|_{L^\infty(p_\rx)}\leq D\), for any \(h, h' \in \cH\).
\end{assumption}

Standard generalization bounds (e.g., \cite{DBLP:journals/tit/RussoZ20,DBLP:conf/itw/LopezJ18,DBLP:conf/colt/ClericoSDD22}) rely on regularity conditions on the loss function $\ell$. We now recall the regularity conditions adopted in this work. We recall that a differentiable function \(\phi: \Y \to \R\) is \(m\)-strongly convex for some constant \(m > 0\) if it satisfies \( \phi(x) \geq \phi(y) + \langle \nabla \phi(y), y - x \rangle + \frac{m}{2}\|x-y\|^2\) and is \(M\)-smooth if it satisfies \( \phi(x) \leq \phi(y) + \langle \nabla \phi(y), y - x \rangle + \frac{M}{2}\|x-y\|^2\).

\begin{assumption}\label{ass:loss-func} 
The loss function takes the form \(\ell(h, \rx) = c(h(\rx), \rx)\) for a function \(c: \Y \times \X \to \R^+\) that is $M$-smooth and, for every \(\rx \in \X\), the function \(c(\cdot, \rx)\) is \(m\)-strongly convex and satisfies \(\inf_{y \in \Y} c(y, \rx) = 0\).
\end{assumption}

This is satisfied by many common learning objectives, including regression with mean squared error and classification with cross-entropy loss. Furthermore, the use of smoothness and strong-convexity is standard within algorithmic stability and generalization (e.g.,~\cite{DBLP:journals/jmlr/BousquetE02,DBLP:conf/nips/BousquetE00,DBLP:conf/icml/CharlesP18,DBLP:journals/corr/abs-1910-07833,DBLP:journals/corr/abs-2307-03357,DBLP:journals/jmlr/Shalev-ShwartzSSS10,DBLP:journals/corr/abs-1902-10710,DBLP:conf/colt/AttiaK22,NEURIPS2021_a4ee59dd}).

\begin{assumption}\label{ass:dim}
Suppose that $\sigma := \E_{\rvx \sim p_\rvx}[\|\rvx\|^2]^{1/2} < \infty$ and $\sigma' := \E_{\rvx \sim p'_\rvx}[\|\rvx\|^2]^{1/2} < \infty$ and suppose that there exists $d^*, C > 0$ such that for all $x \in \operatorname{supp}(p_\rvx)$ and all sufficiently small $\delta > 0$,
\begin{equation*}
    p_{\rx}(B_\delta(x)) \geq C \delta^{d^*}.
\end{equation*}
\end{assumption}

We now state a result showing that the mixed-loss algorithm is uniformly stable and provides a bound on the generalization gap.


\begin{theorem}[Mixed-data Generalization Bound]\label{thm:gen-gap-mixed}
Suppose that Assumptions \ref{ass:diam_H}--\ref{ass:dim} hold, that \(h_\rmS \in \argmin_{h \in \cH} \gR_{\lambda}(h, \rmS)\) and set \(\gR^\star~=~\min_{h \in \cH} r(h)\). Then, there exists \(N_0, \gR^\star_0, \mathcal{W}_0 > 0\) such that whenever \(N \geq N_0, \gR^\star \leq \gR^\star_0\) and \(\wass{2}{p_\rx}{p'_\rx} \leq \mathcal{W}_0\), the algorithm \(\gA(\rmS) = h_\rmS\) is uniformly stable with constant,
\begin{gather*}
    \varepsilon \lesssim \E[\Ell_\rmS(h_\rmS)] + L^{\frac{2d^*}{d^*+2}} \Delta^{\frac{2}{d^* + 2}},\\
    \Delta = \frac{M}{C m^2 \lambda} \gR^\star + \frac{\sqrt{M} D}{C m \lambda  N} + \frac{M \xi}{C m^2} \wass{2}{p_\rx}{p'_\rx},
\end{gather*}
where \(\xi\) is as defined in \eqref{eq:bound-wass_0}. Furthermore, we have the generalization bound,
\begin{align*}
    \E[r(h_\rmS)] \lesssim \gR^\star + \lambda\xi \wass{2}{p_\rvx}{p'_\rvx} + (1-\lambda) L^{\frac{2d^*}{d^*+2}} \Delta^{\frac{2}{d^* + 2}}.
\end{align*}
\end{theorem}

The proof of \Cref{thm:gen-gap-mixed}, along with another result of stability for the mixed loss is discussed in \Cref{app:proof-gen}. The dimension \(d^*\) can be intuitively understood as the intrinsic dimension of the real data manifold. Notably, the generalization bound exhibits a U-shaped dependence on \(\lambda\), similar to \Cref{thm:kernel-error-rate}: for a fixed distributional discrepancy \(\wass{2}{p_\rvx}{p'_\rvx}\), there exists an optimal mixing parameter \(\lambda\). This reflects a trade-off between algorithmic stability (which improves with more synthetic data) and distributional mismatch. In particular, when \(\wass{2}{p_\rvx}{p'_\rvx} = 0\), the optimal \(\lambda\) is 1, suggesting that it is beneficial to generate as much synthetic data as possible. We refer to the ratio \(\frac{\lambda}{1 - \lambda}\) as the \emph{synthetic-to-real} ratio, which approximates \(\frac{M}{N}\) in the finite-sample setting. 


\section{EXPERIMENTS: REAL-WORLD MEDICAL IMAGES} \label{subsec:brain-experiments} 


Multiple sclerosis (MS) is a chronic neurological disease affecting millions worldwide~\citep{tullman2013overview}. T2-hyperintense lesions in MRI reflect neuroinflammatory damage and serve as key biomarkers for diagnosis, monitoring, and prognosis~\citep{mcginley2021diagnosis}.
Accurate segmentation of MS lesions in MRI remains a challenging problem. \gls{ml} methods must contend with substantial variability in image characteristics, lesion appearance, and domain shift, arising from differences in scanners, acquisition protocols, and imaging parameters between training and test sets~\citep{zeng2020review}. Furthermore, publicly available datasets for lesion segmentation remain limited in size and diversity, as acquiring labeled, heterogeneous MRI data is both costly and time-consuming.

\looseness=-1 Our work is motivated by the need to improve lesion segmentation performance under limited and heterogeneous training data and distributional shift. Specifically, we consider the setting where a synthetic data generator is available to address data scarcity and mitigate domain shift between the source (training) and target (test) domains, while potentially introducing additional distributional discrepancies. Following our theoretical result in \Cref{sec:generalization}, we study the in-domain setup in this section and refer to \Cref{sec:domain-shift} for out-of-domain scenario.


\begin{figure}[htb]
    \centering
     \begin{subfigure}[b]{0.5\textwidth}
         \centering
         \includegraphics[width=0.8\textwidth]{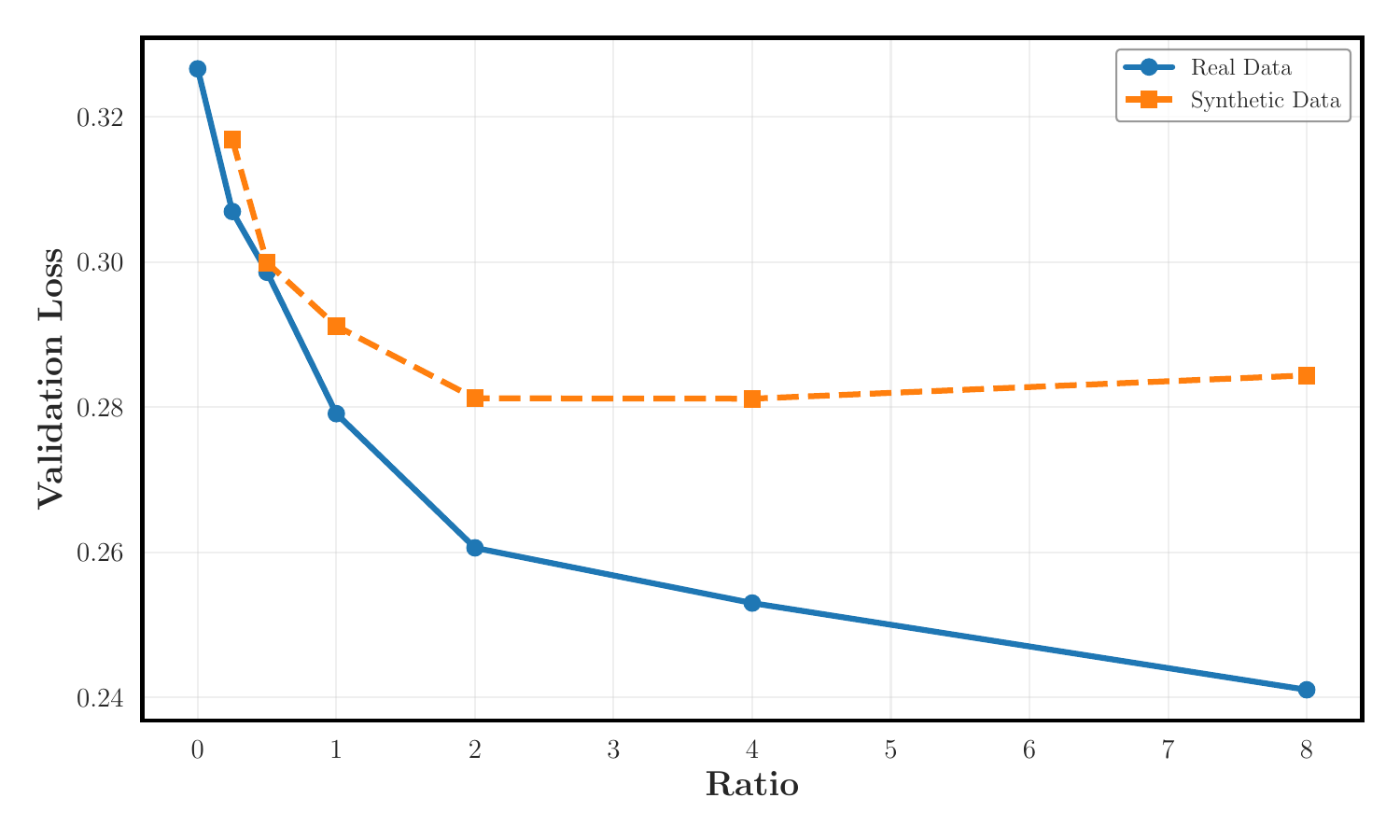}
         \caption{The effect of synthetic data.}
         \label{fig:validloss-vs-ratio}
     \end{subfigure}
     \hfill
     \begin{subfigure}[b]{0.5\textwidth}
         \centering
         \includegraphics[width=0.8\textwidth]{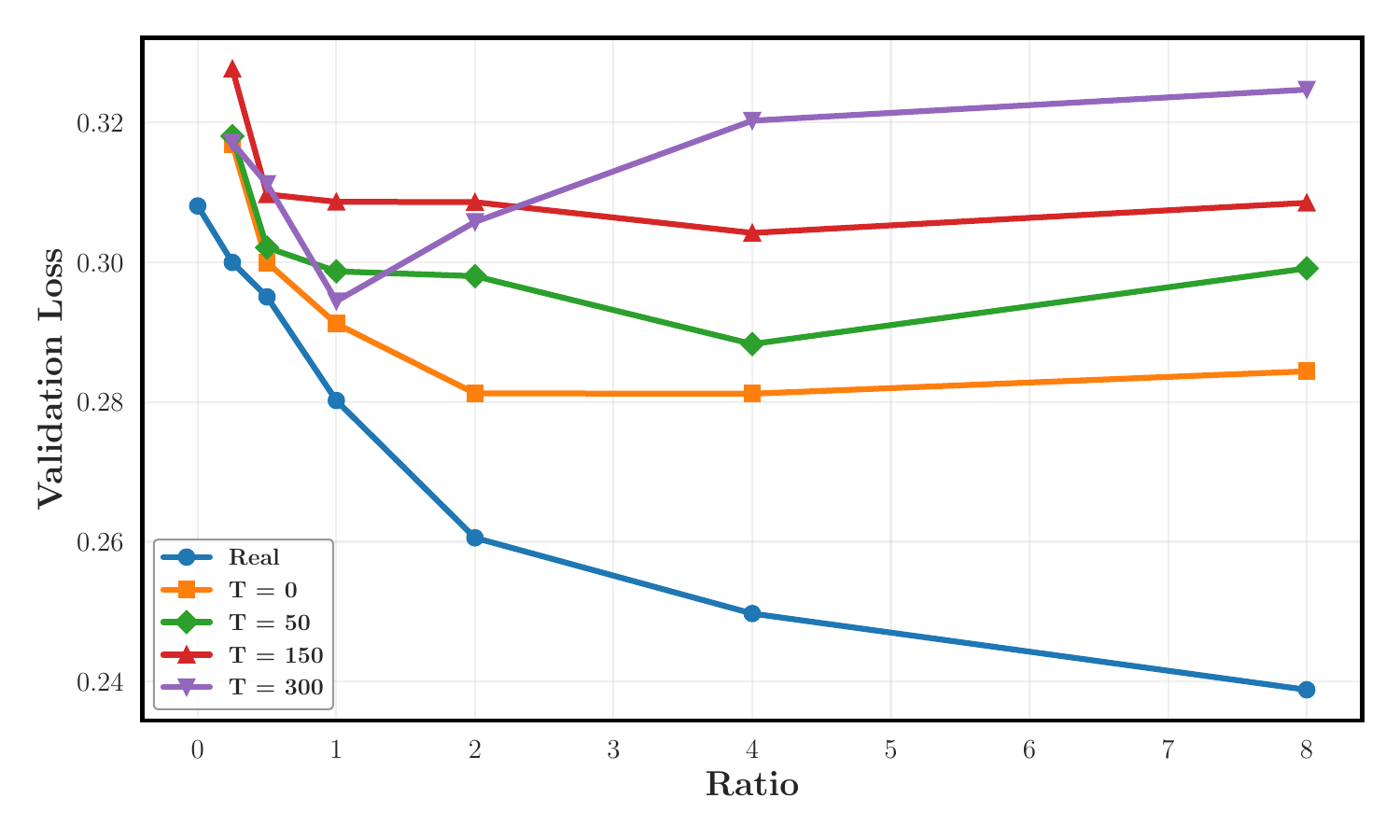}
         \caption{The effect of distributional distance.}
         \label{fig:validloss-vs-timestep}
     \end{subfigure}
    \caption{\emph{(a)} Validation loss decreases consistently as more real data is added (blue line), while increasing synthetic data (orange dashed line) produces a U-shaped curve, indicating an optimal mixing ratio $\lambda$, as predicted by \Cref{thm:gen-gap-mixed}. \emph{(b)} Effect of distributional distance: varying the diffusion model timestep $T \in \{0, 50, 150, 300\}$ controls the noise level of synthetic samples. The U-shaped trend persists across all $T$ but becomes sharper with increased discrepancy between real and synthetic distributions. 
    }
    \label{fig:brain-mri-in-domain}
\end{figure}

We conduct our experiments on the \textit{NO.MS} dataset~\citep{NOMS}, one of the largest and most comprehensive clinical trial datasets for MS. It comprises over 200{,}000 MRI scans from more than 11{,}000 patients. Ground-truth lesion annotations are generated using an automated tool and subsequently refined by expert radiologists. The data originates from two Contract Research Organizations (CROs), NeuroRx and MIAC, introducing inherent domain variability. For the downstream segmentation task, we use a training set of 100 NeuroRx scans ($\sim$4{,}500 slices) and a fixed validation set of 20 NeuroRx scans ($\sim$1{,}000 slices). In addition, we train a conditional diffusion model on the rest of NeuroRx data as our synthetic data generator. To empirically validate the theoretical insights from \Cref{thm:gen-gap-mixed}, we design two experiments:

\vspace{-0.25cm}
\begin{enumerate}[leftmargin=0.75cm]
    \item \textbf{Effect of synthetic data:} We augment the training set by varying the synthetic-to-real ratio from 0.25 to 8, and compare performance against the ground truth, where the real dataset is scaled up. 
    \item \textbf{Distributional distance:} While we do not have direct access to the distance between the true and synthetic distributions, we study the effect of this discrepancy by varying the sampling timestep of the diffusion model ($T =$ 50, 150, 300) out of 600 total denoising steps. We expect that samples from noisier timesteps exhibit greater distributional distance from the real data.
\end{enumerate}
\vspace{-0.25cm}

More details on the segmentation model architecture and hyperparameters are provided in \Cref{app:brain-mri}. \Cref{fig:brain-mri-in-domain} shows that an appropriately chosen synthetic-to-real data ratio improves performance on the downstream segmentation task. \Cref{fig:validloss-vs-ratio} compares validation loss when increasing the amount of synthetic data versus scaling up real data. As expected, adding more real data consistently improves performance. In contrast, synthetic data exhibits a U-shaped effect: moderate amounts enhance generalization, while excessive amounts degrade it, indicating the existence of an optimal interpolation parameter $\lambda$.

\Cref{fig:validloss-vs-timestep} further examines how this behavior depends on the distributional distance between real and synthetic data. By varying the diffusion timestep $T$ which controls the noise level in generated samples, we observe that the U-shape persists but becomes sharper as the synthetic data diverges further from the real distribution. These findings support our results that the generalization gap is influenced by both the mixing ratio and the distributional discrepancy between data sources. The relationship between the optimal synthetic-to-real ratio and distributional distance is further illustrated in \Cref{fig:optimal-ratio-vs-timestep} in \Cref{app:brain-mri}.


\section{DOMAIN ADAPTATION} \label{sec:domain-shift}

In this section, we study the learning problem under \emph{domain shift} \citep{pmlr-v97-zhang19i,stacke2019closer,redko2020survey,SHUI2022109808}: the real training data consist of samples from a source domain \( \gX \) with distribution \( p_\rvx \), while the goal is to evaluate the learned model on a distinct target domain \( \gX^\star \) with distribution \( p_\rvx^\star \), from which no real data are available. To address this distribution mismatch, we assume access to synthetic data generated on the target domain \( \gX^\star \), though drawn from a potentially imperfect distribution \( p_\rvx' \neq p_\rvx^\star \). As in previous sections, this synthetic data is used to regularize the \gls{erm} objective, aiming to improve generalization to the target domain in the absence of real samples from \( p_\rvx^\star \).

We first analyze this setting within the kernel framework (\Cref{sec:kernel-ridge}). Specifically, we consider a dataset of \( N \) real training pairs \( \rvy_n = \Tilde{f}(\rvx_n) + \varepsilon_n \), and a synthetic data generator \( g \) as defined earlier. The test error is measured with respect to a ground truth function \( f_\star \). The main difference from \Cref{sec:kernel-ridge} is that the training function \( \Tilde{f} \) differs from \( f_\star \), capturing the domain shift. Although the empirical estimator remains unchanged (\Cref{eq:kernel-reg}), the generalization behavior is affected by the discrepancy between the training and target domains. Our result shows that stronger regularization can improve performance when the synthetic data more accurately approximates the target domain than the source data, providing a principled guideline for tuning \( \lambda \) under domain shift. 
See \Cref{app:proof-kernel-domain} for the proof.

\begin{figure}[htb]
    \centering
     \begin{subfigure}[htb]{0.5\textwidth}
         \centering
         \includegraphics[width=0.8\textwidth]{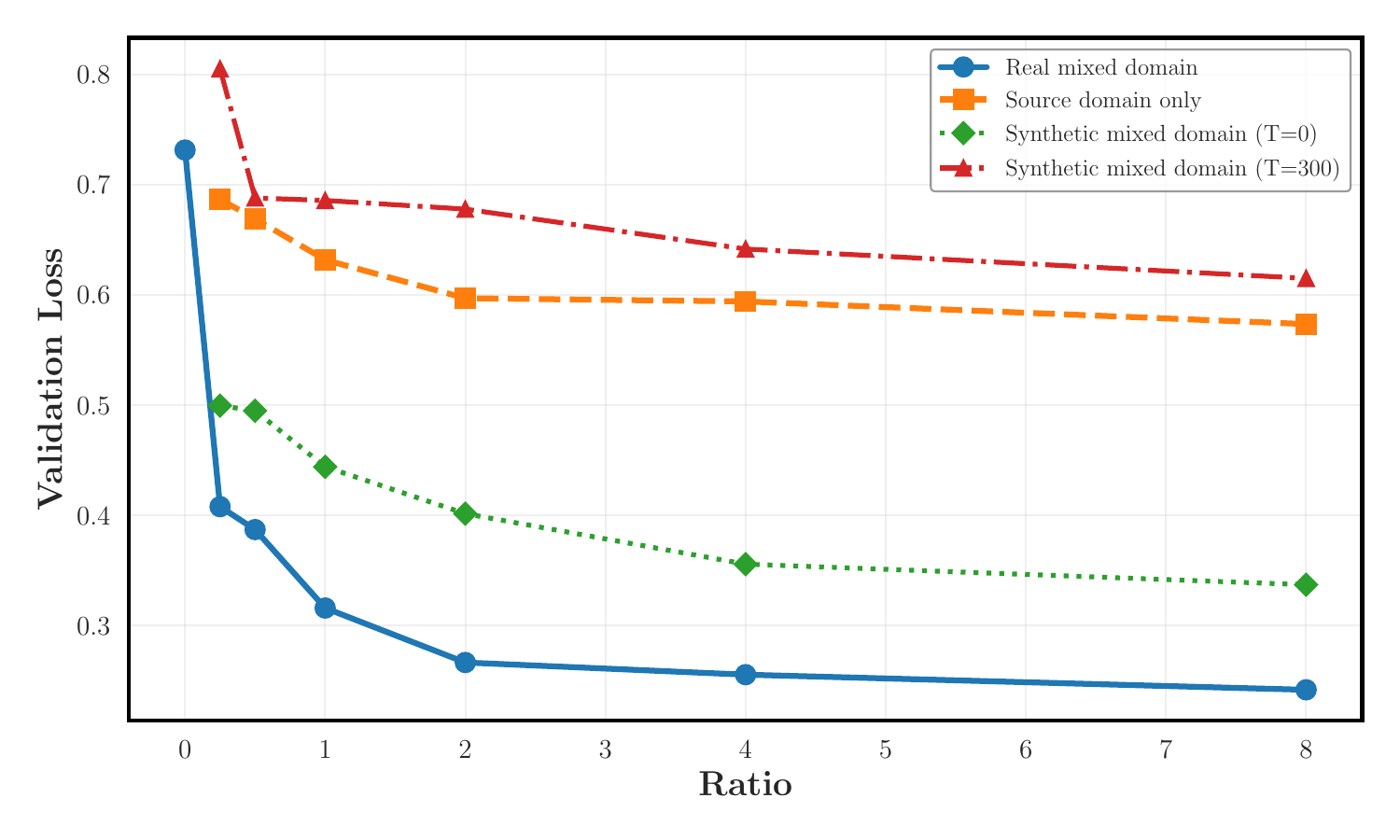}
         \caption{The effect of synthetic data in domain shift.}
         \label{fig:validloss-vs-ratio-out-domain}
     \end{subfigure}
     \hfill
     \begin{subfigure}[htb]{0.5\textwidth}
         \centering
         \includegraphics[width=0.8\textwidth]{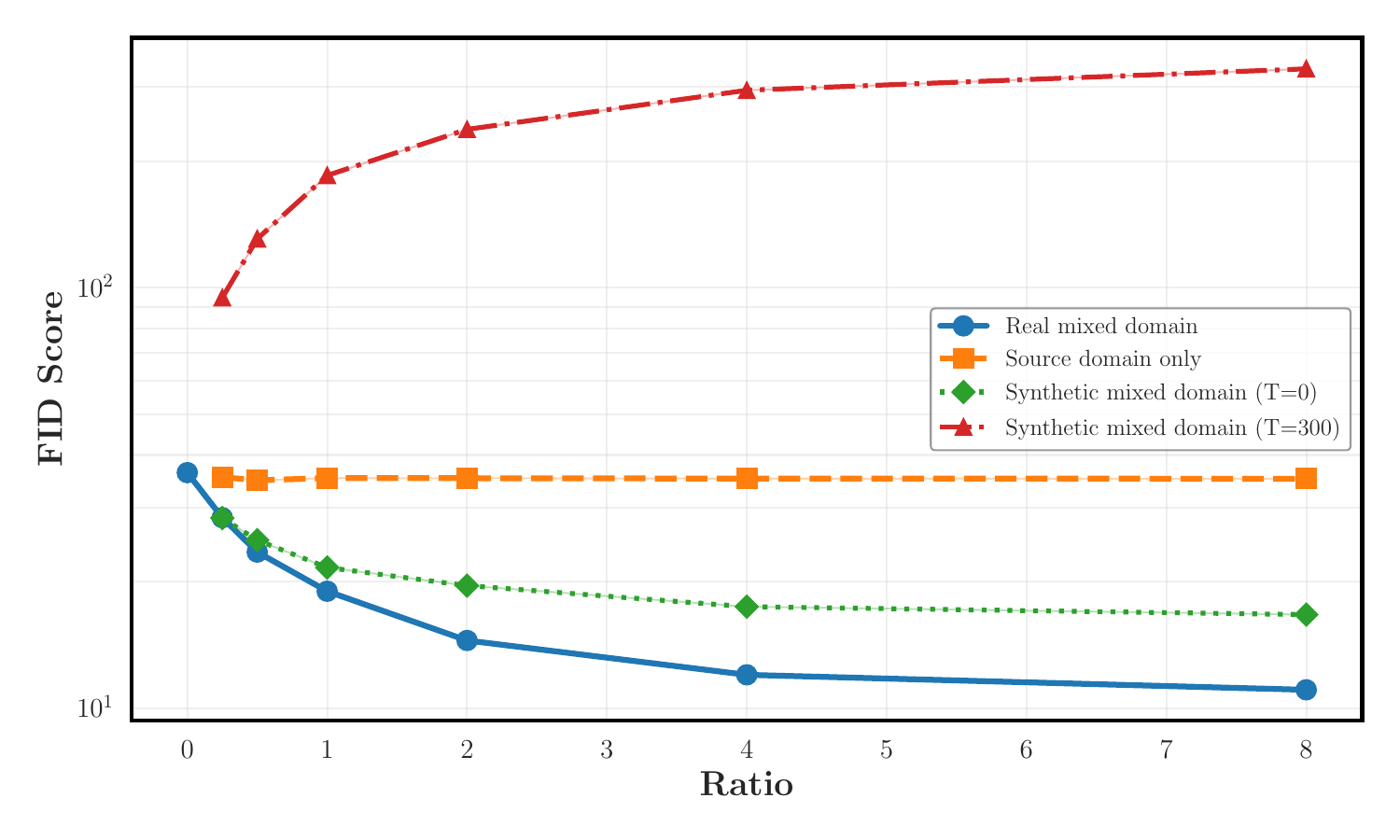}
         \caption{FID score vs. synthetic-to-real data ratio}
         \label{fig:fid-vs-ratio}
     \end{subfigure}
     \caption{\emph{(a)} Effect of synthetic data from distributions close to (green dashed) or far from (red dashed) the target, compared to target (blue) and source (orange) baselines. Results show the trade-off between distributional shift and regularization predicted by \Cref{thm:gen-gap-out-domain}. \emph{(b)} FID as a proxy for distributional shift: $T = 0$ (green) aligns with the target, while noisy (red) and source (orange) data show higher FID and reduced utility.}
\label{fig:brain-mri-out-domain}
\end{figure}

\begin{theorem}[Generalization under Domain Shift]\label{thm:kernel-domain}
    Under \Cref{ass:eigendecay}, for the kernel regression problem defined in \Cref{eq:kernel-reg}, distributional discrepancy \( \gD(\cdot, \cdot) \) as in \Cref{thm:kernel-error-rate}, and any fixed regularization parameter \( \lambda > 0 \), the test error under domain shift satisfies the bound:
    \vspace{-0.15cm}
    \begin{align*}
        \textstyle \mathcal{R}_N(\lambda; g) \leq \frac{\gD(f_\star, \Tilde{f})}{\lambda^2} + \lambda^{2} \mathcal{D}(f_\star, g) + \frac{\sigma^2 + \gD(f_\star, \Tilde{f}) + \gD(f_\star, g)}{N\lambda^2}.
    \end{align*}
\end{theorem}
\vspace{-0.15cm}
We now extend this result to the setup in \Cref{sec:generalization}, where test error is measured with respect to $p^\star_\rvx$. The resulting generalization gap is stated below; see \Cref{app:proof-gen-out-domain} for the proof.
\begin{theorem}[Mixed-data Generalization under Domain Shift]\label{thm:gen-gap-out-domain}
Suppose that Assumptions \ref{ass:diam_H}--\ref{ass:dim} hold and that \(h_\rmS \in \argmin_{h \in \cH} \gR_{\lambda}(h, \rmS)\). Then, whenever \(N \geq N_0, \gR^\star \leq \gR^\star_0\) and \(\wass{2}{p_\rx}{p'_\rx} \leq \mathcal{W}_0\), the generalization gap under the domain shift satisfies
\begin{align*}
    \E[r(h_\rmS)] &\lesssim r^\star + \lambda\xi \wass{2}{p^\star_\rvx}{p'_\rvx} + (1-\lambda)\xi \wass{2}{p^\star_\rvx}{p_\rvx}\\
    & \qquad + (1-\lambda) L^{\frac{2d^*}{d^*+2}} \Delta^{\frac{2}{d^* + 2}},
\end{align*}
where \(r^\star = \min_{h \in \cH} r^\star(h)\) is minimum population risk minimizer of the target domain and \(\Delta, \xi, N_0, \gR^\star_0, \mathcal{W}_0\) are as in Theorem \ref{thm:gen-gap-mixed}.
\end{theorem}
\vspace{-0.15cm}
As expected, compared to \Cref{thm:kernel-error-rate,thm:gen-gap-mixed}, these bounds include an additional term that captures the mismatch between the source and target distributions. Consequently, the optimal choice of \( \lambda \) depends on the relative magnitudes of \( \gD(f_\star, \Tilde{f}) \) and \( \mathcal{D}(f_\star, g) \) or similarly \(\wass{2}{p^\star_\rvx}{p_\rvx}\) and \(\wass{2}{p^\star_\rvx}{p'_\rvx}\). Intuitively, when the synthetic data generator more closely approximates the target domain, it is beneficial to choose a larger regularization parameter \( \lambda \). In the special case where \( f_\star = \Tilde{f} \) or \(\wass{2}{p^\star_\rvx}{p_\rvx} = 0\), the bound reduces to the previous results, albeit with potentially larger constants for the kernel regression case, arising from the more general proof strategy.
\vspace{-0.2cm}
\paragraph{Experimental setup} We follow the experimental setup for medical brain MRI scans described in \Cref{subsec:brain-experiments} to study the effect of domain shift. Since we have two data sources (MIAC and NeuroRx), we can naturally adapt the setup to introduce domain shift: we treat MIAC as the source domain and NeuroRx as the target domain. The synthetic data generator, a conditional diffusion model, is trained on NeuroRx, thus approximating the target distribution. As before, we vary the synthetic-to-real data ratio in the range $0.25$ to $8$, and compare the resulting performance. Results are shown in \Cref{fig:validloss-vs-ratio-out-domain}. We include two baselines in this experiment: (1) access to real data from the target domain for training the downstream segmentation task on NeuroRx (blue line), and (2) no access to either target or synthetic data, with only increased source domain data available (orange line). To examine the impact of distributional discrepancy between synthetic and target data, we adopt the same approach as in \Cref{subsec:brain-experiments}, sampling from the diffusion model at two timesteps, $T \in \{0, 300\}$. We expect $T = 0$ (green dashed line) to closely match the target distribution, while $T = 300$ (red dashed line) reflects a greater distributional distance. As observed, synthetic data can significantly improve performance when the distributional distance between the synthetic and target is small. 
However, when the synthetic generator induces a large distributional shift, using additional source data alone can be more effective, but if the source domain itself is far from the target, neither synthetic nor source data is likely to help.
This observation aligns with our theoretical understanding of the trade-offs in generalization error, where the benefit of additional data depends critically on the distributional closeness to the target domain.

\section{INSIGHTS FOR PRACTITIONERS}\label{sec:practical}

\begin{figure*}[htb]
    \centering
     \begin{subfigure}[bht]{0.99\textwidth}
         \centering
         \includegraphics[width=0.85\textwidth]{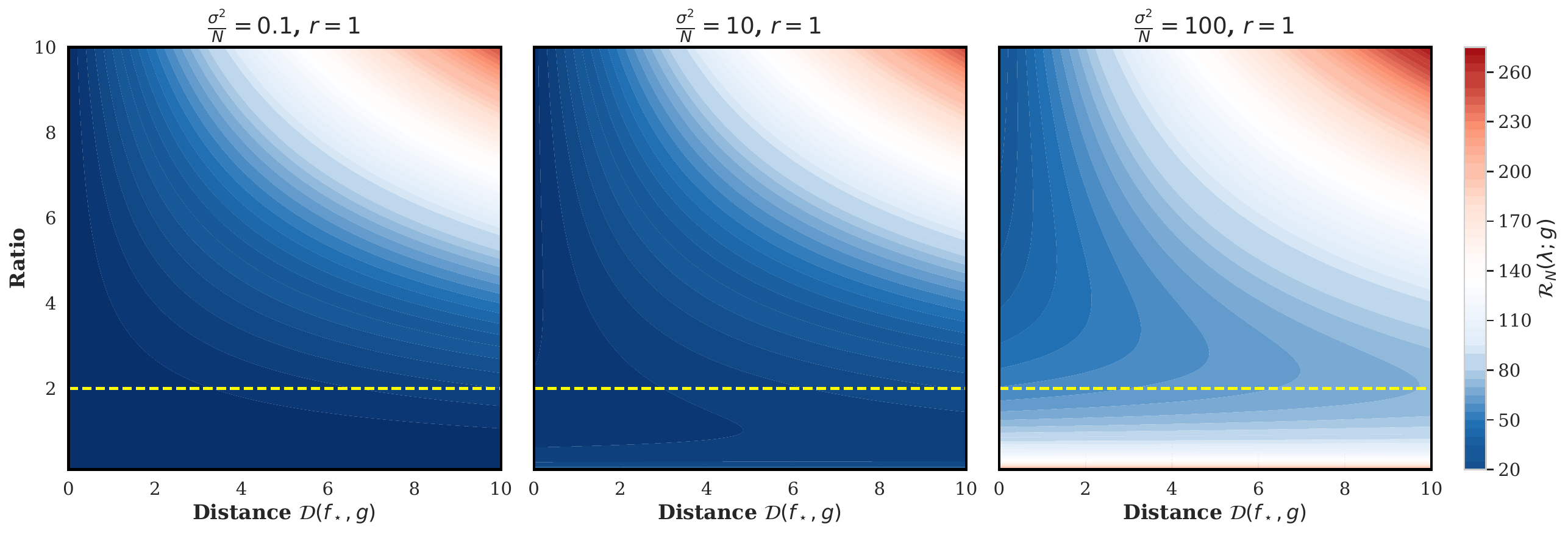}
         \caption{Contour plot illustrating the behavior described in \Cref{thm:kernel-error-rate}.}
         \label{fig:in-domain-heatmap}
     \end{subfigure}
     \vfill
     \begin{subfigure}[bht]{0.99\textwidth}
         \centering
         \includegraphics[width=0.85\textwidth]{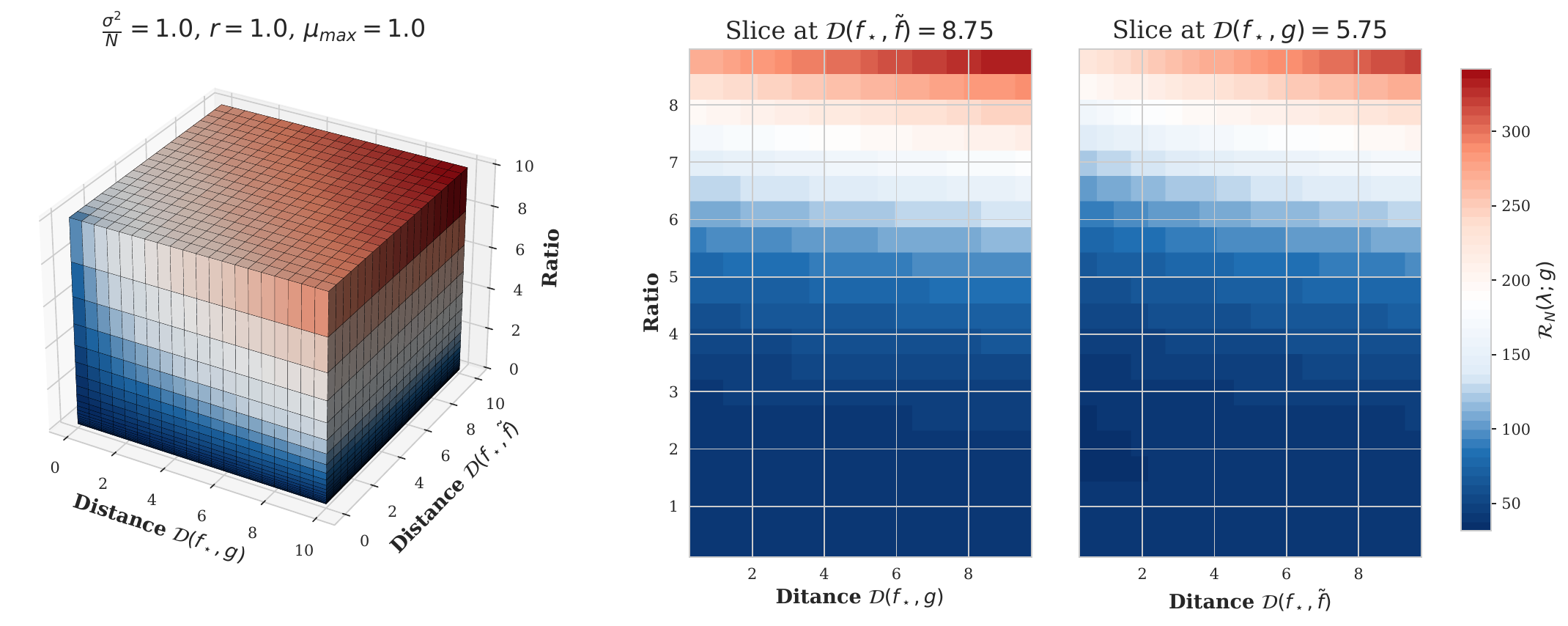}
         \caption{Contour plot illustrating the behavior described in \Cref{thm:kernel-domain}.}
         \label{fig:out-domain-heatmap}
     \end{subfigure}
     \caption{Effect of synthetic-to-real data ratio and distributional distance(s) on the error rate: \emph{(a)} in-domain scenario across various signal-to-noise ratios for the real dataset; \emph{(b)} out-of-domain scenario. In both cases, one should ideally choose $\lambda$ such that it lies within the blue regions, which correspond to lower error rates.}
    \label{fig:strategy-plans}
\end{figure*}

To apply our theoretical results, practitioners must estimate key quantities affecting the generalization bound, such as distributional distances, noise levels, and hypothesis class complexity. This section outlines practical ways to approximate these quantities and offers heuristics based on empirical evidence.

Exact distributional distances between real and synthetic data, or across domains, are rarely available. Applications therefore rely on proxies. In our experiments (\Cref{subsec:brain-experiments,sec:domain-shift}), we first used diffusion timesteps, and here adopt the widely used \gls{fid} metric. While \gls{fid} is not a true distance, it approximates distributional alignment via the Wasserstein distance between Gaussians fitted to Inception embeddings of real and generated images. When synthetic data is well aligned with the target domain, \gls{fid} provides a useful indicator for comparing generators or estimating alignment. As shown in \Cref{fig:fid-vs-ratio}, adding synthetic data from \(T=0\) (green) reduces \gls{fid} similarly to adding real target-domain data (blue), mirroring trends in \Cref{fig:validloss-vs-ratio-out-domain}. In contrast, source-domain or noisy diffusion samples do not improve—and often worsen—\gls{fid}. Depending on context, alternatives such as cross-validation loss or \gls{kld} may also be viable. \looseness=-1

We also propose a more accurate estimate based on \Cref{cor:kernel-opt-lambda}. In image experiments, we exploit frequency-domain features: a 2D Fourier transform preserves information while decomposing energy across frequencies, analogous to the kernel eigenbasis and suitable for analyzing eigendecay.
\vspace{-0.2cm}
\begin{itemize}
    \item \textbf{Distributional distance.}
    Assuming both real and synthetic images are from related distributions (possibly with a shift), we compute the radially averaged power spectral density (RAPSD) for each distribution. We then estimate the distributional distance as the $\ell_2$-distance between the average RAPSD vectors.
    
    \item \textbf{Decay exponent.} The eigendecay exponent $r$ is estimated from the slope of a log-log plot of power spectrum ($Q$) vs.~frequency ($\omega$):
    \[
    Q \asymp \omega^{-2r} 
    \quad \Rightarrow \quad 
    \log Q \approx -2r \log \omega + \text{const}.
    \]
\end{itemize}
\vspace{-0.15cm}
We implement this on a real dataset, where the distance between real and synthetic data at \(T=0\) is $0.85$, which we use for our experiments.

Another critical factor is data variance. This can be tricky, as many datasets are normalized to unit variance. In our setup, the variance in RGB setup yields \(\sigma = 251.4\), giving \(\lambda^\star = 1.958\), close to the observed ratio of 2. While normalization yields $\lambda^\star = 0.12$. This highlights the sensitivity of our result to the scale of variance and is one of the limitations of our method.

\Cref{fig:strategy-plans} illustrates how the bound varies across scenarios, providing heuristics for mixing real and synthetic data. Lower bounds (cooler colors) indicate better generalization; higher ones (red) warn of overfitting. See \Cref{app:practical-insight} for extended results.

\textbf{In-domain.} When heterogeneity is small to moderate and synthetic quality is high (\(D(f_*, g)\) small), augmenting with up to twice as much synthetic as real data is effective (\Cref{fig:in-domain-heatmap}). Beyond this, gains diminish, with higher computational cost but little improvement. In biomedical cases (e.g., brain MRI) with small \(N\) and high \(\sigma^2\), a 1:2 ratio still works well, though further increases only help with exceptionally accurate generators. Otherwise, collecting more real data is preferable.\looseness=-1

\textbf{Out-of-domain.} Under domain shift (\Cref{fig:out-domain-heatmap}), the same 1:2 ratio is robust if the generator is good. Larger ratios harm performance, especially under severe shift (\(\gD(f_\star, \tilde{f}) \approx 8.75\)). For moderate shifts (\(\gD(f_\star, g) \approx 5\)), 1:2 remains a reliable choice. Overall, ratios between 1:1 and 1:2 seem to be effective. Modest augmentation helps even with shift, but excessive synthetic data misaligned with the target distribution degrades performance. Careful tuning of the augmentation ratio is thus crucial in the out-of-domain case.

Finally, our theory assumes Lipschitz continuity, which can be estimated via gradient norms or enforced by clipping. This only scales the bound by a constant and does not affect the optimal ratio order.

\section{CONCLUSION}\label{sec:conclusion}

Synthetic data is vital in domains where real data is scarce, costly, or sensitive, such as healthcare. We develop a principled framework that characterizes the trade-off between real and synthetic data and prove the existence of an optimal synthetic-to-real ratio that minimizes generalization error, first in kernel regression and then more generally via stability. Experiments on benchmarks and real-world datasets validate these predictions, revealing a non-monotonic relationship between performance and synthetic proportion and extending to domain adaptation with distribution shift.  

We show that most traditional, model-agnostic techniques that rely on uniform bounds are often loose and unable to capture the phenomena observed in practice. Our modified \gls{erm} objective relaxes these assumptions by ignoring the variance of synthetic data, and yields tighter guarantees. However, a deeper understanding is still needed; for instance, PAC-Bayes bounds that treat synthetic data as a prior may provide a promising next step.\looseness=-1 


\acknowledgments{We thank Russ Webb and Dieter Haering for their useful feedback.}

\bibliography{bib}

\clearpage
\section*{Checklist}

\begin{enumerate}

  \item For all models and algorithms presented, check if you include:
  \begin{enumerate}
    \item A clear description of the mathematical setting, assumptions, algorithm, and/or model. \textcolor{AppleBlue}{Yes, see \Cref{sec:kernel-ridge,sec:generalization}, and \Cref{app:notation} for the mathematical setup.}
    \item An analysis of the properties and complexity (time, space, sample size) of any algorithm. \textcolor{AppleBlue}{Yes, the main results and analysis are in \Cref{thm:kernel-domain,thm:kernel-error-rate} for the kernel regression case and \Cref{thm:gen-gap-mixed,thm:gen-gap-out-domain} for the general case.}
    \item (Optional) Anonymized source code, with specification of all dependencies, including external libraries. \textcolor{AppleBlue}{No, experiments primarily serve to illustrate our theoretical findings. We are happy to release the code upon request by the reviewers.}
  \end{enumerate}

  \item For any theoretical claim, check if you include:
  \begin{enumerate}
    \item Statements of the full set of assumptions of all theoretical results. \textcolor{AppleBlue}{Yes, see \Cref{ass:eigendecay}, \Cref{ass:diam_H} and \Cref{ass:loss-func}.}
    \item Complete proofs of all theoretical results. \textcolor{AppleBlue}{Yes, proofs are available in \Cref{app:proof-kernel-reg,app:proof-kernel-domain} and \Cref{app:proof-gen-gap-mixed,app:proof-gen-out-domain}.}
    \item Clear explanations of any assumptions. \textcolor{AppleBlue}{Yes, see \Cref{ass:eigendecay}, \Cref{ass:diam_H} and \Cref{ass:loss-func} for references.}
  \end{enumerate}

  \item For all figures and tables that present empirical results, check if you include:
  \begin{enumerate}
    \item The code, data, and instructions needed to reproduce the main experimental results (either in the supplemental material or as a URL). \textcolor{AppleBlue}{No, experiments primarily serve to illustrate our theoretical findings. We are happy to release the code upon request by the reviewers.}
    \item All the training details (e.g., data splits, hyperparameters, how they were chosen). \textcolor{AppleBlue}{Yes, all experimental details are provided in \Cref{subsec:brain-experiments} and \Cref{app:kernel-experiments,app:cifar-experiments}.}
    \item A clear definition of the specific measure or statistics and error bars (e.g., with respect to the random seed after running experiments multiple times). \textcolor{AppleBlue}{Yes, see \Cref{fig:mean-loss-ratio,fig:mean-optimal-choice,fig:mean-domain-shift} for the error bar and averaged runs over three seeds.}
    \item A description of the computing infrastructure used. (e.g., type of GPUs, internal cluster, or cloud provider). \textcolor{AppleBlue}{Yes, the compute budget and infrastructure is explained in \Cref{sec:all-exp}.} 
  \end{enumerate}

  \item If you are using existing assets (e.g., code, data, models) or curating/releasing new assets, check if you include:
  \begin{enumerate}
    \item Citations of the creator If your work uses existing assets. \textcolor{AppleBlue}{Yes, the description of the data along with the related references is available in \Cref{subsec:brain-experiments,app:cifar-experiments}).}
    \item The license information of the assets, if applicable. \textcolor{AppleBlue}{Yes. For NO.MS, the raw data (anonymized) and associated documents (e.g., protocol, reporting and analysis plan, clinical study report) can be requested via \href{https://www.clinicalstudydatarequest.com}{https://www.clinicalstudydatarequest.com} by signing a Data Sharing Agreement.}
    \item New assets either in the supplemental material or as a URL, if applicable. \textcolor{AppleBlue}{Not Applicable.}
    \item Information about consent from data providers/curators. \textcolor{AppleBlue}{Yes. We have signed the Data Sharing Agreement for No.MS as explained before. CIFAR10 is publicly available.}
    \item Discussion of sensible content if applicable, e.g., personally identifiable information or offensive content. \textcolor{AppleBlue}{Not Applicable.}
  \end{enumerate}

  \item If you used crowdsourcing or conducted research with human subjects, check if you include:
  \begin{enumerate}
    \item The full text of instructions given to participants and screenshots. \textcolor{AppleBlue}{Not Applicable.}
    \item Descriptions of potential participant risks, with links to Institutional Review Board (IRB) approvals if applicable. \textcolor{AppleBlue}{Not Applicable.}
    \item The estimated hourly wage paid to participants and the total amount spent on participant compensation. \textcolor{AppleBlue}{Not Applicable.}
  \end{enumerate}

\end{enumerate}

\clearpage
\onecolumn
\hypersetup{linkcolor=black}
\appendixpage
\startcontents[sections]
\printcontents[sections]{l}{1}

\clearpage
\hypersetup{linkcolor=hrefblue}


\clearpage
\appendix
\thispagestyle{empty}
\onecolumn
\section{BROADER IMPACT}\label{sec:impact}

This work shows how synthetic data can be effectively integrated with real data to improve the performance and generalization of downstream tasks in both in-domain and out-of-domain settings.

There are several potential benefits of our work:
\begin{enumerate}[leftmargin=0.75cm]
\item Our framework enables practitioners to use an effective synthetic-to-real data ratio that yields improved performance at a reduced computational cost, therefore reducing carbon footprint.\looseness=-1 
\item Although our experiments focus on lesion segmentation, the underlying theory and insights are broadly applicable. Practitioners in various domains can leverage our framework to address challenges related to low-data regimes and domain shifts by exploiting powerful generative models to synthesize data, issues that are common across many applied fields.
\item We identify key factors necessary for evaluating the impact of synthetic data. This is particularly relevant in the current landscape, where a wide range of generative and foundation models are available to generate synthetic data. Our findings can help the community make more informed decisions about incorporating the generated samples from these models, particularly their quantity and quality.
\item Our results highlight the importance of distributional shift in achieving better performance, which in turn underscores the potential value of incorporating human feedback into the synthetic data generation process.
\item In scenarios involving biased datasets—closely related to our distribution shift setup—our framework offers a principled way to generate an adequate number of synthetic samples to improve model performance. This is particularly useful not only in data-scarce domains such as healthcare but also in datasets lacking diversity.
\end{enumerate}

We also acknowledge potential risks and undesirable consequences associated with our approach. In efforts to maximize downstream task performance, practitioners may be incentivized to collect additional data or train more powerful generative models. This introduces several challenges: 
\begin{enumerate}[leftmargin=0.75cm]
\item Collecting extensive data about a subject raises concerns about responsible data acquisition.
\item Training larger generative models requires increased computational resources, which may have a greater environmental impact.
\end{enumerate}

\section{LIMITATIONS}\label{sec:limitations}

While our work provides theoretical insights and practical guidelines for combining synthetic and real data, several limitations remain:

\begin{enumerate}[leftmargin=0.75cm]
\item Our analysis involves certain approximations to key parameters that affect the generalization bound and the optimal synthetic-to-real data ratio. The sensitivity of the results to our approach, and studying other ways of approximating them needs further investigation.
\item Although our theory aligns with empirical trends observed in lesion segmentation, we have not validated the proposed bounds across a broader range of applications. Extending the empirical evaluation to diverse domains would help assess the generality of our framework.
\item We focus on providing theoretical and practical insights and do not present a concrete algorithm that integrates a specific real dataset with a synthetic data generator. Developing such an algorithm would facilitate adoption in real-world settings.
\item Our experiments are restricted to the image modality. Investigating how the framework extends to other data types, such as text, audio, or multimodal settings, remains an open and promising direction for future work.
\end{enumerate}

\section{OTHER NOTATIONS}\label{app:notation}

We denote scalar or vector-valued \gls{rv} by $\rvx$, and collections of \gls{rv}s by $\rmX$, with corresponding probability densities $p_{\rvx}$ and $p_{\rmX}$. Realizations of these variables are denoted by $\vx$ and $\mX$, respectively, with $\vx$ taking values in a measurable space $\mathcal{X}$. The conditional distribution of a random variable $\rvy$ given $\rvx = \vx$ is denoted by $p_{\rvy|\rvx = \vx}$. The expectation of a measurable function $f: \mathcal{X} \to \R$ is written as $\E[f(\rvx)] = \E_{\vx \sim p_{\rvx}}[f(\vx)]$. For integers $a \leq c \leq b$, we denote by $\rmX_{a:b} = \{\rvx_a, \rvx_{a+1}, \ldots, \rvx_b\}$ a finite collection of \gls{rv}s, and by $\rmX_{a:b}^{(\neq c)} = \rmX_{a:b} \setminus \{\rvx_c\}$ the subset excluding $\rvx_c$. The density $p_{\rmX_{a:b}}$ denotes the joint density of the variables in $\rmX_{a:b}$.

We use $[K]$ to denote the index set $\{1, \ldots, K\}$, and reserve Latin letters for samples and Greek letters for parameters or distributions.

\begin{definition}[Lipschitz Continuity]
A function $f: \gZ \to \R^q$, for $(\gZ, d_\gZ)$ a metric space, is $\xi$-Lipschitz if for all $z, z' \in \gZ$, $\|f(z) - f(z')\| \leq \xi \, d_\gZ(z, z')$.
\end{definition}

Let $\gR_\lambda(h) = \E_\rmS[\gR_{\lambda}(h, \rmS)]$ denote the expected mixed loss, and $r_\lambda(h)$ the hybrid population risk:
\begin{align}\label{eq:r-mixed}
    r_\lambda(h_\rmS) &= (1-\lambda) r(h) + \lambda \E_{\rvx \sim p'_{\rvx}}\left[ \ell(h, \rvx)\right].
\end{align}

\begin{definition}[Generalization Gap]\label{def:generalization-gap}
    The \emph{generalization error} of a hypothesis $h$ is defined as the absolute difference between its population and empirical risks:
     \begin{align*}
          g_\rmS(h) = |\Ell_\X(h) - \Ell_\rmS(h)|\,.
     \end{align*}
     The \emph{generalization gap} of a learning algorithm is the expected generalization error:
     \begin{align*}
     \G = \E_{p_{\rh, \rmS}}[g_\rmS(\rh)] = \E_{p_{\rh, \rmS}} \left[|\Ell_\X(\rh) - \Ell_\rmS(\rh)|\right]\,.
     \end{align*}
 \end{definition}
 We can now define the generalization gap in the mixed-data setting as:
 \begin{align*}\label{eq:gap-mixed}
     \G = \E_{p_{\rh, \rmS}}[g_\rmS(\rh)] = \E_{p_{\rh, \rmS}}\left[|r(\rh) - \gR_\lambda(\rh, \rmS)|\right].
 \end{align*}

\section{TRADITIONAL BOUNDS FAIL TO CAPTURE THE TRADE-OFF}\label{app:traditional-bound}

We use the traditional bounds in learning theory to study the mixture of real and synthetic data. We assume a bounded loss $\ell \in [0,1]$ and hypothesis class $\mathcal H$. The goal is to bound the distance between $\hat \gR_{p_\rvx}(h) = \Ell_{\rmS}(h)$ and $\gR_{p_\rvx}(h) = r(h)$ as expressed in \Cref{sec:generalization}.

\paragraph{Bias via Mixture Mismatch} For any $f \in \gH$, and $\alpha\in[0,1]$, let $q_\alpha := (1-\alpha) p_\rvx + \alpha p'_\rvx$. We have
\begin{align}
\gR_{p_\rvx}(f)
&=R_{q_\alpha}(f)+\alpha \big(\gR_{p_\rvx}(f)-\gR_{p'_\rvx}(f)\big) \\
&\le \gR_{q_\alpha}(f)+\alpha\,\mathrm{IPM}_{\ell\circ\mathcal H}(p_\rvx,p'_\rvx), \label{eq:tradition-bias}
\end{align}
where \(
\mathrm{IPM}_{\ell\circ\mathcal H}(p_\rvx,p'_\rvx)
:=\sup_{f\in\mathcal H}\left|\mathbb E_{p_\rvx}[\ell(f)]-\mathbb E_{p'_\rvx}[\ell(f)]\right|\).

\paragraph{Estimation for $\gR_{q_\alpha}(f)$}
Let $S_N\sim p_\rvx^{\otimes N}$ and $S'_M\sim p'^{\otimes M}_\rvx$ 
be independent samples of size $N, M$, from $p_\rvx$ and $p'_\rvx$, respectively.
Suppose we train on a mixture of these samples so the empirical risk becomes
\[
\hat \gR_{q_\alpha}(f)
=(1-\alpha)\,\hat \gR_{S_N}(f)+\alpha\,\hat \gR_{S'_M}(f),
\]
and the population risk is $\gR_{q_\alpha}(f)$. By standard uniform convergence bounds \citep{DBLP:journals/jmlr/Shalev-ShwartzSSS10}, with probability at least $1-\delta$, simultaneously for all $f\in\mathcal H$,
\begin{align*}
\gR_{p_\rvx}(f) &\le \hat \gR_{S_N}(f)+\varepsilon_N,\\
\gR_{p'_\rvx}(f)&\le \hat \gR_{S'_M}(f)+\varepsilon_M,
\end{align*}
where, for example,
\begin{align*}
\varepsilon_N &= 2\,\mathfrak R_N^{(p_\rvx)}(\ell\circ\mathcal H)
+\sqrt{\tfrac{\log(4/\delta)}{2N}},\\
\varepsilon_M &= 2\,\mathfrak R_M^{(p'_\rvx)}(\ell\circ\mathcal H)
+\sqrt{\tfrac{\log(4/\delta)}{2M}}.
\end{align*}
Here $\mathfrak R_n^{(P)}$ is the Rademacher complexity under distribution $P$. Now, multiplying the two bounds by $(1-\alpha)$ and $\alpha$, respectively and adding, we have
\begin{align}\label{eq:tradition-pop}
    \gR_{q_\alpha}(f)\;\le\;\hat \gR_{q_\alpha}(f)+(1-\alpha)\varepsilon_N+\alpha\varepsilon_M.
\end{align}

\paragraph{Combine with bias} Let $\hat f_\alpha\in\arg\min_{f\in\mathcal H}\hat \gR_{q_\alpha}(f)$. Then, with probability at least $1-\delta$, by combining \Cref{eq:tradition-bias,eq:tradition-pop}, we have:
\begin{align}\label{eq:tradition}
    \gR_{p_\rvx}(\hat f_\alpha)
\;\le\;
\hat \gR_{q_\alpha}(\hat f_\alpha)
+(1-\alpha)\varepsilon_N+\alpha\varepsilon_M
+\alpha\,\mathrm{IPM}_{\ell\circ\mathcal H}(p_\rvx,p'_\rvx)\,.
\end{align}

Suppose a favourable regime where the Rademacher complexity shrinks at the canonical rate $1/\sqrt{n}$ with a distribution and sample size independent constant. Note that kernel methods with bounded kernels (\Cref{sec:kernel-ridge}) and Lipschitz losses (\Cref{sec:generalization}) are examples of such regimes. Therefore, we can bound the Rademacher complexities as
\[\mathfrak R_n^{(P)}(\ell\circ\mathcal H)\le C_{\ell\circ\mathcal H}/\sqrt{n}\]
for some constant $C_{\ell\circ\mathcal H}$ independent of $P$ and $n$. Then we can simplify the above to
\[\varepsilon_N
\le 2C_{\ell\circ\mathcal H}/\sqrt{N}+\sqrt{\log(4/\delta)/(2N)},\qquad
\varepsilon_M
\le 2C_{\ell\circ\mathcal H}/\sqrt{M}+\sqrt{\log(4/\delta)/(2M)}.
\]
Therefore, there exists a constant $C_\delta$ such that the following holds:
\[\varepsilon_N \leq \frac{C_\delta}{\sqrt{N}}, \qquad \varepsilon_M\leq \frac{C_\delta}{\sqrt{M}}\]

\subsection{Choice of Optimum Regularization in the Traditional Setting}
In our approach, we use an empirical risk of the form
\begin{align}\label{eq:tradition-erm}
\tilde \gR(f)
=\frac1{N+M}\sum_{i=1}^{N}\ell(f(X_i),Y_i)
+ \frac1{N+M}\sum_{j=1}^{M}\ell(f(X'_i),Y'_i)\,,    
\end{align}
where $S_N = \left((X_i,Y_i); i \in [N]\right)$ and $S'_M = \left((X'_j,Y'_j); j \in [M]\right)$ are samples from $p_\rvx$ and $p'_\rvx$, respectively.

To connect to the previous analysis, we can choose
\(
\alpha_N = \frac{M}{N+M}
\).
With this choice of $\alpha$, and by \Cref{eq:tradition}, we have 
\[
\gR_D(\hat f_{\alpha_N})
\;\le\;
\hat \gR_{q_{\alpha_N}}(\hat f_{\alpha_N})
+\frac{N}{M+N}\frac{C_\delta}{\sqrt{N}}+\frac{M}{M+N}\frac{C_\delta}{\sqrt{M}}
+\frac{M}{N+M}\,\mathrm{IPM}_{\ell\circ\mathcal H}(p_\rvx,p'_\rvx).
\]
Assume that $M = \lambda N$ for some $\lambda > 0$. 
Then, we have
\[
\gR_D(\hat f_{\alpha_N})
\;\le\;
\hat \gR_{q_{\alpha_N}}(\hat f_{\alpha_N})
+C_\delta \frac{1+\sqrt{\lambda}}{(1+\lambda)\sqrt{N}}
+\frac{\lambda}{1+\lambda}\,\mathrm{IPM}_{\ell\circ\mathcal H}(p_\rvx,p'_\rvx).
\]

Note that the second term,
\(C_\delta \frac{1+\sqrt{\lambda}}{(1+\lambda)\sqrt{N}}
\),
which arises from the variance component, attains a non-trivial maximum. When $\lambda = 0$, the bound reduces to the error using only real data (i.e., without synthetic augmentation). As $\lambda$ increases, the variance term eventually decreases, while the bias term increases but only up to the discrepancy between the two distributions. Consequently, the optimal $\lambda$ is either zero or infinity, depending on whether the discrepancy is smaller than the variance term at $\lambda = 0$. Under a fixed computational budget, however, this yields a non-trivial optimal choice of $\lambda$.

\paragraph{Optimal Choice of $\alpha$ (Effective Sample View)} Let us bound the estimation error by
\[
(1-\alpha)\varepsilon_N+\alpha\varepsilon_M
\;\lesssim\;\frac{c}{\sqrt{N_\alpha}},\qquad 
N_\alpha=(1-\alpha)N+\alpha M,
\]
with $c\approx \tilde 2C_{\ell\circ\mathcal H} + \sqrt{\log(1/\delta)}$. Thus, we need to minimize the following bound:
\[
\rho(\alpha)=\frac{c}{\sqrt{(1-\alpha)N+\alpha M}}+\alpha \mathrm{IPM}_{\ell\circ\mathcal H}(p_\rvx,p'_\rvx).
\]

Let $\Delta=M-N$ and
\(
n^\star=\left(\frac{\,c\,\Delta}{2\mathrm{IPM}_{\ell\circ\mathcal H}(p_\rvx,p'_\rvx)}\right)^{2/3}
\). Then, the optimal mixing weight is
\[
\;
\alpha^\star=\mathrm{clip}_{[0,1]}\!\left(\frac{n^\star-N}{M-N}\right).
\;
\]

\noindent
This provides us with the following decision rule:
\[
\alpha^\star=
\begin{cases}
0, & \mathrm{IPM}_{\ell\circ\mathcal H}(p_\rvx,p'_\rvx)\ge \dfrac{c}{\sqrt N},\\[2mm]
1, & \mathrm{IPM}_{\ell\circ\mathcal H}(p_\rvx,p'_\rvx)\le \dfrac{c}{\sqrt M},\\[2mm]
\frac{n^\star-N}{M-N}, & \text{otherwise}.
\end{cases}
\]

\paragraph{Optimal $M$ with Fixed $N$ (Uniform Concatenation)}
If training is done on the mixture dataset, following \Cref{eq:tradition-erm} with each point weighted equally, i.e.~,
\[
\alpha=\frac{M}{N+M}=\frac{\lambda}{1+\lambda},\qquad M=\lambda N,
\]
then, we need to optimize with respect to the following bound:
\[
\rho(\lambda)=\frac{c}{\sqrt{N(1+\lambda)}}+\frac{\lambda}{1+\lambda}\mathrm{IPM}_{\ell\circ\mathcal H}(p_\rvx,p'_\rvx),\qquad \lambda\ge 0.
\]

\noindent
This formalization gives us the following threshold behavior:
\[
M^\star=
\begin{cases}
0, & \mathrm{IPM}_{\ell\circ\mathcal H}(p_\rvx,p'_\rvx)\ \ge\ \dfrac{c}{\sqrt N},\\[2mm]
\text{``as large as allowed,''} & \mathrm{IPM}_{\ell\circ\mathcal H}(p_\rvx,p'_\rvx)\ <\ \dfrac{c}{\sqrt N}.
\end{cases}
\]

\noindent
A finite \emph{balance point} is obtained by equating variance and bias:
\[\;M_{\text{bal}}=\Bigg(\Big(\frac{2c}{\mathrm{IPM}_{\ell\circ\mathcal H}(p_\rvx,p'_\rvx)\sqrt N}\Big)^{\!2}-1\Bigg)_+\;N.\;
\]

Therefore, although traditional bounds offer an intuitive understanding of the \gls{erm}’s behavior, they do not capture its complete behavior and can be overly loose. 

To solve this problem and bridge the gap between the generalization bounds and what we observe in practice, we propose an alternative option for the optimization problem that we introduce in the next section.

\subsection{Alternative Formalization: Analysis of Ansatz}\label{app:ansantz}

In practice, we have the following \gls{erm}:
\begin{align}\label{george:ERM}
    f_N = \argmin_{f \in \mathcal{H}_k}  \sum_{n = 1}^N \left(\ry_n - f(\rx_n)\right)^2 + \sum_{m=1}^M \left(\tilde{\ry}_m - f(\tilde{\rx}_m)\right)^2.
\end{align}

Notice that the objective is equivalent to
\begin{align*}
    &\sum_{n = 1}^N \left(\ry_n - f(\rx_n)\right)^2 + \sum_{m=1}^M \left(\tilde{\ry}_m - f(\tilde{\rx}_m)\right)^2\\
    &\stackrel{\circ}{=}
    \frac{1}{N}\sum_{n = 1}^N \left(\ry_n - f(\rx_n)\right)^2 + \frac{1}{N}\sum_{m=1}^M \left(\tilde{\ry}_m - f(\tilde{\rx}_m)\right)^2\\
    &\stackrel{\circ}{=}
    \frac{1}{N}\sum_{n = 1}^N \left(\ry_n - f(\rx_n)\right)^2 + \frac{M}{N} \frac{1}{M}\sum_{m=1}^M \left(\tilde{\ry}_m - f(\tilde{\rx}_m)\right)^2\,.
    \intertext{We can now approximate the empirical loss over the synthetic samples with the corresponding population loss:}
    &\approx 
    \frac{1}{N}\sum_{n = 1}^N \left(\ry_n - f(\rx_n)\right)^2 + \frac{M}{N} \|f - g\|^2\,.
    \intertext{Letting $\lambda = M/N$:}
    &=
    \frac{1}{N}\sum_{n = 1}^N \left(\ry_n - f(\rx_n)\right)^2 + \lambda \|f - g\|^2\,.
    \intertext{Using $\lambda  = \tilde{\lambda}/(1-\tilde \lambda)$, we have}
    &=\frac{1}{N}\sum_{n = 1}^N \left(\ry_n - f(\rx_n)\right)^2 + \frac{\tilde \lambda}{1-\tilde \lambda} \|f - g\|^2
    \\
    &\stackrel{\circ}{=} 
    \frac{1-\tilde \lambda}{N}\sum_{n = 1}^N \left(\ry_n - f(\rx_n)\right)^2 + \tilde \lambda \|f - g\|^2.
\end{align*}

In this formalization, we deliberately omit the variance term arising from the synthetic samples, as it is expected to play only a secondary role and may obscure the regularizing effect introduced by the synthetic data. As we see in \Cref{sec:kernel-ridge,sec:generalization}, this formalization helps us to improve the bound for the U-shape observed in practice, as opposed to the classical case, where methods fail to capture this trade-off.


\section{ASYMPTOTIC EFFECT OF SYNTHETIC DATA IN KERNEL REGRESSION} \label{sec:infinite-data-kernel}

Suppose we have \(M\) synthetic samples \(\{(\tilde x_m,\tilde y_m)\}_{m=1}^M\), where \(\tilde x_m \sim p(x)\) i.i.d., and \(\tilde y_m = g(\tilde x_m)\). We assume these synthetic samples are noiseless, reflecting access to the exact synthetic data generator. Then the \gls{erm} objective
\begin{align*}
    f_N = \argmin_{f \in \gH_k}\; \frac{1}{N} \sum_{n=1}^N (y_n - f(x_n))^2
    + \lambda\, \frac{1}{M} \sum_{m=1}^M \bigl(f(\tilde x_m) - g(\tilde x_m)\bigr)^2
\end{align*}
satisfies, by the (strong) law of large numbers,
\begin{align*}
    \lim_{M \to \infty} \frac{1}{M} \sum_{m=1}^M \bigl(f(\tilde x_m) - g(\tilde x_m)\bigr)^2
    = \E_{\tilde x \sim p} \bigl[(f(\tilde x) - g(\tilde x))^2\bigr]
    = \bigl\langle f - g,\, T_K(f - g) \bigr\rangle_{\gH_k},
\end{align*}
where the kernel integral operator \(T_K : \gH_k \to \gH_k\) is defined by
\begin{align*}
    (T_K\,h)(\cdot)
    = \int K(\cdot, x)\, h(x)\, p(x)\, \mathrm{d}x.
\end{align*}
Note that while the synthetic covariates \(\tilde x_m\) are drawn i.i.d.\ from the same marginal distribution as the real data, the synthetic labels \(\tilde y_m\) follow a potentially different mapping \(g\), as determined by the data generator.

\paragraph{Equivalence of \(\boldsymbol{L^2(p)}\) and \gls{rkhs} norms}
{By Mercer’s theorem \citep{Mercer:1909dea}, the operator \(T_K\) admits the spectral decomposition
\[
K(x,y) = \sum_{i=1}^\infty \mu_i\, \phi_i(x)\, \phi_i(y),
\]
where \(\{\phi_i\}\) form an orthonormal basis in \(L^2(p)\) and \(\mu_i > 0\) are the eigenvalues. Any function \(h \in \gH_k\) can be written as \(h = \sum_i a_i \sqrt{\mu_i}\, \phi_i\), yielding
\[
\|h\|_{L^2(p)}^2 = \sum_{i=1}^\infty \mu_i\, a_i^2,
\qquad
\|h\|_{\gH_k}^2 = \sum_{i=1}^\infty a_i^2.
\]
If the nonzero eigenvalues satisfy \(0 < \mu_{\min} \le \mu_i \le \mu_{\max} < \infty\), then
\[
\mu_{\min} \|h\|_{\gH_k}^2 \le \|h\|_{L^2(p)}^2 \le \mu_{\max} \|h\|_{\gH_k}^2,
\]
so the norms are equivalent up to constants:
\[
\|h\|_{L^2(p)}^2 \asymp \|h\|_{\gH_k}^2.
\]
Under this spectral assumption, the \(L^2(p)\) term \(\E_{\tilde x \sim p}[(f(\tilde x) - g(\tilde x))^2]\) appearing in the infinite-\(M\) limit is thus proportional to \(\|f - g\|_{\gH_k}^2\).}

Note that we use this equivalence to motivate our setup, we study the effect of limited synthetic data in \Cref{sec:generalization} more precisely.

\section{TECHNICAL PROOFS OF MODIFIED KERNEL REGRESSION}

\subsection{Proof of \texorpdfstring{\Cref{lem:kernel-reg}}{lemma 2.1}}\label{app:proof-kernel-reg}
\begin{manuallemma}{\ref{lem:kernel-reg}}
Let \( K_N \in \R^{N \times N} \) be the empirical kernel matrix with entries \( (K_N)_{ij} = K(\rx_i, \rx_j) \). Define the integral operator \( T_K : L^2(p_x) \to L^2(p_x) \) by \((T_K f)(\rx) = \int K(\rx, x') f(x')\, dp_x(x') = \E_{\rx'}\left[K(\rx, \rx') f(\rx')\right]\).
    Let \( \lambda_N = N \lambda \). Then the solution to \Cref{eq:kernel-reg} has the closed-form representation:
    \begin{align*}
        \textstyle \boldsymbol{\alpha} = \left(K_N + \lambda_N I\right)^{-1} \left(K_N \boldsymbol{\alpha}_\star + \lambda_N \boldsymbol{\beta} + \boldsymbol{\varepsilon}\right)\,,
    \end{align*}
    where \( \boldsymbol{\alpha}_\star \), \( \boldsymbol{\beta} \), and \( \boldsymbol{\varepsilon} \) are the coefficients of \( f_\star \), \( g \), and the noise vector in the training basis. 
\end{manuallemma}

\begin{proof}
Let us define $\lambda = \frac{\tilde{\lambda}}{1- \tilde{\lambda}}$. We can rewrite the \gls{erm} using the Representer theorem as following:
\begin{align*}
    \boldsymbol{\alpha} = \argmin_{\hat{\boldsymbol{\alpha}}} \frac{1}{N} \|\rvy - K_N \hat{\boldsymbol{\alpha}}\|^2 + \lambda \|\hat{\boldsymbol{\alpha}} - \boldsymbol{\beta}\|^2_{\mathcal{H}_k} + \lambda \|g_\perp\|^2_{\mathcal{H}_k}\,.
\end{align*}

\medskip
\noindent\textbf{Finite‐sample solution.} Taking the derivation with respect to $\boldsymbol{\alpha}$, we have
    \begin{align}\label{eq:kernel-derivation}
        K_N\left[(K_N + \lambda_N I) \boldsymbol{\alpha} - \rvy - \lambda_N \boldsymbol{\beta}\right] = 0
    \end{align}
    Solving the optimization, similarly to the standard regularized kernel regression, we achieve:
    \begin{align*}
        \boldsymbol{\alpha} = \left(K_N + \lambda_N I\right)^{-1}(\rvy + \lambda_N \boldsymbol{\beta})\,,
    \end{align*}
    where we conclude the proof by noting that $\rvy = K_N \boldsymbol{\alpha}_\star + \boldsymbol{\varepsilon}$. This also results in the fact that $f_N(\rx) = K_\rx \left(K_N + \lambda_N I\right)^{-1}(\rvy + \lambda_N \boldsymbol{\beta})$, where $K_\rx = (K(\rx, \rx_1), \ldots, K(\rx, \rx_N))$. 

We now study the behavior of this closed-form solution in the population limit, which becomes useful later.

\medskip
\noindent\textbf{Population limit and expectation.} We use \Cref{eq:kernel-derivation}, so we have:
    \begin{align*}
        K_N\left[(K_N + \lambda_N I) (\boldsymbol{\alpha} -\boldsymbol{\beta})- \rvy + K_N \boldsymbol{\beta}\right] &= 0\\
        \boldsymbol{\alpha} - \boldsymbol{\beta} &= \left(K_N + \lambda_N I\right)^{-1}(\rvy - K_N\boldsymbol{\beta})\\
        f_N(\rx) - g(\rx) &= K_\rx \left(K_N + \lambda_N I\right)^{-1}(\rvy - K_N\boldsymbol{\beta})\,.
    \end{align*}
    Now, consider the population limit where the sample size $N \to \infty$. The empirical kernel matrix $K_N$ converges to the integral operator $T_K$, which is a classic approach in kernel ridge regression \citep{singh2023kernel}. Therefore, $T_K \approx \frac{1}{N} \sum_{n = 1}^N K(\cdot, \rx_n) \otimes K(\cdot, \rx_n)$, which means $K_N \approx N T_K$. So, we have
    \begin{align*}
        f_N - g &= \left(T_K +\lambda I\right)^{-1}\left( T_K f_\star - T_K g + \frac{1}{N}\boldsymbol{\varepsilon}\right)\\
        \E_\varepsilon\left[f_N - g\right] &= \left(T_K + \lambda I\right)^{-1} T_K (f_\star - g)\,.
    \end{align*}
    Noting that $\E_\varepsilon[g] = g$, we get the following result in the population limit:
    \begin{align}
        \E_{\varepsilon}\left[f_N\right] = g + \left(T_K + \lambda I\right)^{-1} T_K (f_\star - g)\,.
    \end{align}
\end{proof}

\subsection{Proof of \texorpdfstring{\Cref{thm:kernel-error-rate}}{theorem 2.2}}\label{app:kernel-error-rate}

\begin{manualthm}{\ref{thm:kernel-error-rate}}
    Under \Cref{ass:eigendecay}, for the kernel regression problem defined in \Cref{eq:kernel-reg} and any fixed regularization parameter \( \lambda > 0 \), the test error admits the bound:
    \begin{align*}
        \mathcal{R}_N(\lambda; g) = \mathcal{O}\left(\frac{\mathcal{D}(f_\star, g) + \sigma^2}{N\lambda^2} + \lambda^{2 - \frac{1}{4r}} \mathcal{D}(f_\star, g)
        \right)\,,
    \end{align*}
    where \(\mathcal{D}(f_\star, g)^2 = \sum_{j = 1}^\infty \frac{1}{\mu_j^2}(\theta_j - \omega_j)^2\) denotes the discrepancy between the target function \( f_\star \) and the synthetic generator \( g \).
\end{manualthm}

\begin{proof}
To bound the test error, we use the bias-variance decomposition in \Cref{def:bias-variance}. We start with the variance term. We follow the approach of \cite{DBLP:journals/corr/abs-2403-08938}. So, we have:
\begin{align}
    \mathcal{V} &= \E_{\rvx, \varepsilon} \left[(f_N(\rvx) - \E_\varepsilon\left[f_N(\rvx)\right])^2\right]\\
    &= \E_{\rvx, \varepsilon} \left[\left(\frac{1}{N}K_\rvx \left(T_K + \lambda I\right)^{-1}\boldsymbol{\varepsilon}\right)\right]\\
    &= \frac{\sigma^2}{N^2} \tr\left(N T_K^2 \left(T_K+ \lambda I\right)^{-2}\right)\\
    &= \frac{\sigma^2}{N} \sum_{j = 1}^\infty \frac{\mu_j^2}{(\mu_j+\lambda)^2}\,.
\end{align}
Note that the above discussion assumes the population limit. An analogous behaviour holds in the finite-sample setting. Define $\kappa = \sup_{\vx \in \gX} \sqrt{K(\vx, \vx)}$. Then
\begin{align}
    \mathcal{V} &= \E_{\rvx, \varepsilon} \!\left[(f_N(\rvx) - \E_\varepsilon[f_N(\rvx)])^2\right] \\
    &= \E_{\rvx, \varepsilon} \!\left[\left(\tfrac{1}{N}K_\rvx (K_N + \lambda I)^{-1}\boldsymbol{\varepsilon}\right)^2\right] \\
    &\leq \frac{\kappa^2 \sigma^2}{N \lambda^2}\,, \label{eq:var-rate}
\end{align}
where the inequality follows from $\|(K_N + \lambda I)^{-1}\|\leq \frac{1}{\lambda}$ and the bound $\|K_\rvx\|^2\leq N \kappa^2$. Now, let us bound the bias term. We have:
\begin{align*}
    \mathcal{B}^2 &= \E_\rvx \left[(f_\star - \E_\varepsilon\left[f_N\right])(\rx)\right]^2\\
    &\leq \E_\rmS\left[\|\E_\varepsilon\left[f_N\right] - f_\lambda\|_{\gH_K}^2\right] + \|f_\star - f_\lambda\|_{p_\rvx}^2\,,
\end{align*}
where the last line is resulted from Jensen's inequality and triangle inequality. Moreover, note that $\|f_\star - f_\lambda\|_{p_\rvx}^2 \leq \kappa^2 \|f_\star - f_\lambda\|_{\gH}^2$. We define $f_\lambda$ as the population limit of $f_N$:
\begin{align}\label{eq:pop_limit-f}
    f_\lambda = g + (T_K + \lambda I)^{-1}T_K (f_\star - g)\,.
\end{align}
To bound the bias term, we also define an additional auxiliary function $f_{N, \lambda}$:
\begin{align*}
    f_{N, \lambda} = g + (K_N + \lambda I)^{-1}T_K (f_\star - g)\,.
\end{align*}
This function helps us compute the first term of bias.

\medskip
\noindent\textbf{Population and sampling bias $\E_\rmS\left[\|\E_\varepsilon\left[f_N\right] - f_\lambda\|_{\gH_K}^2\right]$.} We rewrite this term as follows:
\begin{align*}
    \E_\rmS\left[\|\E_\varepsilon\left[f_N\right] - f_\lambda\|_{\gH_K}^2\right] &\leq \E_\rmS\left[\|\E_\varepsilon\left[f_N\right] - f_{N, \lambda}\|_{\gH_K}^2\right] + \E_\rmS\left[\|f_{N, \lambda} - f_\lambda\|_{\gH_K}^2\right]\\
    &\leq \E_\rmS\left[\|(K_N+\lambda I)^{-1}(K_N - T_K)(f_\star - g)\|_{\gH_K}^2\right] + \E_\rmS\left[\|f_{N, \lambda} - f_\lambda\|_{\gH_K}^2\right]\\
    &\leq \frac{1}{\lambda^2} E_\rmS\left[\|(K_N - T_K)(f_\star - g)\|_{\gH_K}^2\right] + \E_\rmS\left[\|f_{N, \lambda} - f_\lambda\|_{\gH_K}^2\right]\\
    &\leq \frac{\kappa^2 \|f_\star - g\|_{p_\rvx}^2}{N\lambda^2} + \E_\rmS\left[\|f_{N, \lambda} - f_\lambda\|_{\gH_K}^2\right]\,,
\end{align*}
where we used the fact that $\|(K_N+\lambda I)^{-1}\| \leq \frac{1}{\lambda}$, and the last inequality is proved in \cite{SMALE2005285}[Theorem $3$]. We continue the bound by first noticing that \Cref{eq:pop_limit-f} gives us $(T_K + \lambda I)(f_\lambda - g) = T_K(f_\star - g)$:
\begin{align*}
    \E_\rmS\left[\|\E_\varepsilon\left[f_N\right] - f_\lambda\|_{\gH_K}^2\right] &\leq \frac{\kappa^2 \|f_\star - g\|_{p_\rvx}^2}{N\lambda^2} + \E_\rmS\left[\|f_{N, \lambda} - f_\lambda\|_{\gH_K}^2\right]\\
    &\leq \frac{\kappa^2 \|f_\star - g\|_{p_\rvx}^2}{N\lambda^2} + \E_\rmS\left[\|(K_N + \lambda I)^{-1}(T_K + \lambda I)(f_\lambda - g) - (f_\lambda - g)\|_{\gH_K}^2\right]\\
    &\leq \frac{\kappa^2 \|f_\star - g\|_{p_\rvx}^2}{N\lambda^2} + \E_\rmS\left[\|(K_N + \lambda I)^{-1}(T_K - K_N)(f_\lambda - g)\|_{\gH_K}^2\right]\\
    &\leq \frac{\kappa^2 \|f_\star - g\|_{p_\rvx}^2}{N\lambda^2} + \frac{1}{\lambda^2} \E_\rmS\left[\|(T_K - K_N) (f_\lambda - g)\|_{\gH_K}^2\right]\\
     &\leq \frac{\kappa^2 \|f_\star - g\|_{p_\rvx}^2}{N\lambda^2} + \frac{\kappa^2 \|f_\lambda - g\|_{p_\rvx}^2}{N\lambda^2}\,,
\end{align*}
where we have used \cite{SMALE2005285}[Theorem $3$] once more. Moreover, we note that \Cref{eq:pop_limit-f} is also the solution to the following Kernel optimization:
\begin{align*}
    f_\lambda = g + \argmin_{f \in \gH_K}\left\{\|f - (f_\star - g)\|_{p_\rvx}^2 + \lambda \|f\|_{\gH_K}^2\right\}\,.
\end{align*}
Therefore, setting $f$ to zero, we have $\|f_\lambda - (f_\star - g)\|_{p_\rvx}^2 + \lambda \|f_\lambda\|_{\gH_K}^2 \leq \|f_\star - g\|_{p_\rvx}^2$, from which we can conclude that $\|f_\lambda - g\|_{p_\rvx}^2 \leq 2 \|f_\star - g\|_{p_\rvx}^2$. Putting all these results together and the fact that $\|f_\star - g\|_{p_\rvx}^2 \leq \sup_{\vx} K(x, x) \|f_\star - g\|_{\gH_K}^2$, we have:
\begin{align}\label{eq:bias-sample}
    \E_\rmS\left[\|\E_\varepsilon\left[f_N\right] - f_\lambda\|_{\gH_K}^2\right] &\leq \frac{3\kappa^4 \|f_\star - g\|_{\gH}^2}{N\lambda^2}\,.
\end{align}

\medskip
\noindent\textbf{Population bias $\|f_\star - f_\lambda\|_{\gH}^2$.} To bound this term, we substitute the Mercer decomposition of $f_\star$ and $g$, and the fact that the eigenvalues of $(T_K + \lambda I)^{-1}$ are $1 / (\lambda + \mu_j)$ as following:
\begin{align*}
    f_\lambda &= g + \left(T_K + \lambda I\right)^{-1} T_K (f_\star - g)\\
    &= \sum_{j}^\infty \left(\frac{\mu_j}{\mu_j + \lambda}\theta_j + \frac{\lambda}{\mu_j + \lambda}\omega_j\right)\phi_j\,.
\end{align*}
Therefore, we have
\begin{align*}
    \|f_\star - f_\lambda\|_{\gH}^2 = \|\sum_{j = 1}^\infty \frac{\lambda}{\mu_j + \lambda}(\theta_j - \omega_j) \phi_j\|^2\,,
\end{align*}
We can bound this bias term as follows by noting that $\{\phi_j\}_j$ consist the orthonormal basis:
\begin{align}\label{eq:bias-rate}
    \|f_\star - f_\lambda\|_{\gH}^2 = \sum_{j = 1}^\infty \frac{\lambda^2}{(\mu_j + \lambda)^2} (\theta_j - \omega_j)^2\,.
\end{align}

Now, combining the results of \Cref{eq:bias-sample,eq:bias-rate} gives us an upper-bound for $\mathcal{B}^2$, and combining them with \Cref{eq:var-rate} gets a bound for the test error. We have:
\begin{align*}
    \mathcal{R}_N(\lambda; g) &= \frac{3\kappa^4 \|f_\star - g\|_{\gH}^2}{N\lambda^2} + \sum_{j = 1}^\infty \frac{\lambda^2}{(\mu_j + \lambda)^2} (\theta_j - \omega_j)^2 + \frac{\kappa^2\sigma^2}{N\lambda^2}\\
    &\leq \frac{3\kappa^4 \|f_\star - g\|_{\gH}^2}{N\lambda^2} + \lambda^2 \sqrt{\left(\sum_{j = 1}^\infty \frac{\mu_j^2}{(\mu_j + \lambda)^2} \right)\left(\sum_{j = 1}^\infty \frac{1}{\mu_j^2}(\theta_j - \omega_j)^2\right)} + \frac{\kappa^2 \sigma^2}{N \lambda^2}\,,
\end{align*}
where the inequality is due to the Cauchy-Schwarz inequality. Since $\sum_{j = 1}^\infty \frac{1}{\mu_j^2}(\theta_j - \omega_j)^2 = \mathcal{D}(f_\star, g)^2$, we can now write:
\begin{align*}
    \mathcal{R}_N(\lambda; g) &= \mathcal{O}\left(\frac{3\kappa^4 \|f_\star - g\|_{\gH}^2}{N\lambda^2} + \lambda^2 \sqrt{\mathcal{D}(f_\star, g)^2\left(\sum_{j = 1}^\infty \frac{\mu_j^2}{(\mu_j + \lambda)^2}\right)} + \frac{\kappa^2 \sigma^2}{N\lambda^2} 
    \right)\\
    &= \mathcal{O}\left(\frac{3\kappa^4 \|f_\star - g\|_{\gH}^2}{N\lambda^2} + \lambda^2 \mathcal{D}(f_\star, g) \sqrt{\int_{0}^\infty \left(\frac{x^{-2r}}{x^{-2r} + \lambda}\right)^2\,dx} + \frac{\kappa^2\sigma^2}{N\lambda^2} 
    \right)\\
    &= \mathcal{O}\left(\frac{3\kappa^4 \|f_\star - g\|_{\gH}^2}{N\lambda^2} + \lambda^{2 - \frac{1}{4r}} \mathcal{D}(f_\star, g) C_r + \frac{\kappa^2\sigma^2}{N\lambda^2} 
    \right)\,, \label{eq:final-rate-kernel}
\end{align*}
where 
\begin{align*}
    C_r = \left(\int_{0}^\infty \left(\frac{x^{-2r}}{x^{-2r} + \lambda}\right)^2\,dx\right)^{1/2} = \left(\frac{1}{2r}\int_{0}^\infty \frac{v^{1-1/2r}}{(v+1)^2}\, dv\right)^{1/2} = \sqrt{\frac{B(1/2r, 2 - 1/2r)}{2r}}\,,
\end{align*}
with $B(z_1, z_2)$ denoting the beta function. We conclude by noting that $\mathcal{D}(f_\star, g) \geq \|f_\star - g\|_{\gH}^2$.
\end{proof}

\subsection{Proof of \texorpdfstring{\Cref{cor:kernel-opt-lambda}}{corollary}} \label{app:kernel-optimal}

\begin{manualcor}{\ref{cor:kernel-opt-lambda}}
    Under the assumptions of \Cref{thm:kernel-error-rate}, the optimal regularization parameter that minimizes the test error is given by
    \begin{align*}
        \lambda^\star \asymp \left(\frac{\mathcal{D}(f_\star, g) + \sigma^2}{N \mathcal{D}(f_\star, g)}\right)^{\frac{4r}{16r+1}}.
    \end{align*}
    Setting \( \lambda = \frac{M}{N} \), the optimal number of synthetic samples satisfies:
    \begin{align*}
        M^\star \asymp \left(1 + \frac{\sigma^2}{\mathcal{D}(f_\star, g)}\right)^{\frac{4r}{16r+1}} N^{\frac{12r + 1}{16r + 1}}.
    \end{align*}
\end{manualcor}

\begin{proof}
    The result follows by minimizing the bound in \Cref{thm:kernel-error-rate}:
    \[
        \mathcal{R}_N(\lambda; g) = \mathcal{O}\left(\frac{\mathcal{D}(f_\star, g) + \sigma^2}{N\lambda^2} + \lambda^{2 - \frac{1}{4r}} \mathcal{D}(f_\star, g) \right)\,,
    \]
    We differentiate the right-hand side with respect to \( \lambda \) and set the derivative to zero:
    \begin{align*}
        \frac{\partial \mathcal{R}_N}{\partial \lambda} 
        &= \left(2 - \frac{1}{4r} \right) \lambda^{1 - \frac{1}{4r}} \mathcal{D}(f_\star, g)  - 2\frac{\mathcal{D}(f_\star, g) + \sigma^2}{N} \lambda^{-3} = 0\,.
    \end{align*}
    Solving for \( \lambda \) gives
    \begin{align*}
        \lambda^\star 
        = \left(\frac{8r \left(\mathcal{D}(f_\star, g) + \sigma^2\right)}{(8r - 1)N \mathcal{D}(f_\star, g)}\right)^{\frac{4r}{16r+1}} 
        \asymp \left(\frac{\mathcal{D}(f_\star, g) + \sigma^2}{N \mathcal{D}(f_\star, g)}\right)^{\frac{4r}{16r+1}}\,. 
    \end{align*}
    Substituting \( \lambda^\star = \frac{M^\star}{N} \) yields
    \begin{align*}
        M^\star = N \lambda^\star \asymp \left(1 + \frac{\sigma^2}{\mathcal{D}(f_\star, g)}\right)^{\frac{4r}{16r+1}} N^{\frac{12r + 1}{16r + 1}}\,,
    \end{align*}
    completing the proof.
\end{proof}

\section{GENERALIZATION BOUND WITH MIXED REAL AND SYNTHETIC}\label{app:proof-gen}

\subsection{Lemmata}

\begin{definition}
The upper packing dimension of a measure \(\nu\) is the quantity \(d^*\) defined by:
\begin{equation*}
    d^* := \operatorname{ess\,sup}(\Phi^*), \quad \Phi^*(x) := \limsup_{\delta \to 0} \frac{\log p_\delta(x)}{\log{\delta}}.
\end{equation*}
\end{definition}


\begin{definition}[$\gD$-Regularity {\cite{DBLP:conf/colt/ClericoSDD22}}]\label{def:Dreg}
Let $\gD$ be a measurable map $\gP \times \gP \to [0, +\infty]$. Fix $\mu \in \gP$ and $\xi \geq 0$. We say that a function $f: \gZ \to \R$ is $R_\gD(\xi)$-regular with respect to $\mu$ if $f \in L^1(\mu)$ and for every $\nu \in \gP$ such that $\Supp(\nu) \subseteq \Supp(\mu)$ and $f \in L^1(\nu)$,
\[
|\E_\mu[f(Z)] - \E_\nu[f(Z)]| \leq \xi\, \gD(\mu, \nu).
\]
\end{definition}


\begin{lemma}\label{lem:stability}
Consider a mapping \(A: \Sc_N \to \cH\) and define the random variable \(\tilde{\rx} \sim p_\rx\) such that \(\tilde{\rx} \indep \rmS\). Suppose there exists \(\varepsilon \geq 0\) such that for any \(i \in \{1, ..., N\}\), it holds that
\begin{equation}\label{eq:stability_1}
    \sup_i \E[\ell(h^i, \rx_i) - \ell(h, \rx_i)] \leq \varepsilon,
\end{equation}
where \(h = A(\rmS), h^i = A(\rmS^i)\) and \(\rmS^i = \{\rx_1, ..., \rx_{i-1}, \tilde{\rx}, \rx_{i+1}, ..., \rx_N\}\). Then it holds that
\begin{equation*}
    \E[\Ell_\X(A(\rmS)) - \Ell_\rmS(A(\rmS))] \leq \varepsilon.
\end{equation*}
\end{lemma}
\begin{proof}
Follows from Lemma 7 of \cite{DBLP:journals/jmlr/BousquetE02}.
\end{proof}

Now we obtain Wasserstein continuity bounds using a similar technique to \cite{DBLP:journals/corr/Polyanskiy015,DBLP:journals/corr/RaginskyRT17}.
\begin{lemma}\label{lem:bound-wass}
Suppose that Assumptions~\ref{ass:diam_H}--\ref{ass:dim} hold, then for any $h \in \cH$,
\begin{equation}\label{eq:bound-wass_0}
    \big | \E_{p_\rx}[\ell(h, \rx)] - \E_{p'_\rx}[\ell(h, \rx)] \big | \leq \xi \, \gW_2(p_\rx, p'_\rx), \qquad \xi := 3\sqrt{1+L^2}\left(\|\nabla c(0,0)\| + M\tau\right),
\end{equation}
where $\tau^2 = 2B_\cH^2 + (1+2L^2)(\sigma^2 + {\sigma'}^2)$, and $B_\cH = \sup_{h \in \cH} \|h(0)\|$.
\end{lemma}
\begin{proof}
Define the map $\phi_h(\rx) = (h(\rx), \rx)$. Since $h$ is $L$-Lipschitz,
\begin{equation*}
    \|\phi_h(\rx) - \phi_h(\rx')\|^2 = \|h(\rx) - h(\rx')\|^2 + \|\rx - \rx'\|^2 \leq (1+L^2)\|\rx - \rx'\|^2,
\end{equation*}
so $\phi_h$ is $\sqrt{1+L^2}$-Lipschitz. It follows that
\begin{equation}\label{eq:pushforward-w2}
    \gW_2(\phi_h \# p_\rx,\, \phi_h \# p'_\rx) \leq \sqrt{1+L^2}\, \gW_2(p_\rx, p'_\rx).
\end{equation}
Let $(X, Y)$ be an optimal $\gW_2$-coupling of $p_\rx$ and $p'_\rx$, and set $Z = \phi_h(X)$ and $W = \phi_h(Y)$. By $M$-smoothness of $c$,
\begin{equation*}
    c(Z) - c(W) \leq \langle \nabla c(W),\, Z - W \rangle + \frac{M}{2}\|Z - W\|^2.
\end{equation*}
Taking expectations and applying Cauchy--Schwarz to the first term gives
\begin{equation}\label{eq:smooth-split}
    \E[c(Z)] - \E[c(W)] \leq \sqrt{\E[\|\nabla c(W)\|^2]}\; \gW_2(Z, W) + \frac{M}{2}\,\gW_2(Z, W)^2.
\end{equation}
We bound each term separately. For the quadratic term, the triangle inequality for $\gW_2$ gives $\gW_2(Z, W) \leq \sigma_Z + \sigma_W$ where $\sigma_Z = \sqrt{\E[\|Z\|^2]}$, so that
\begin{equation}\label{eq:quadratic-bound}
    \frac{M}{2}\gW_2(Z,W)^2 \leq \frac{M}{2}(\sigma_Z + \sigma_W)\,\gW_2(Z,W).
\end{equation}
Since $h$ is $L$-Lipschitz, $\|h(y)\| \leq \|h(0)\| + L\|y\| \leq B_\cH + L\|y\|$, and therefore
\begin{equation*}
    \sigma_W^2 = \E[\|h(Y)\|^2 + \|Y\|^2] \leq 2B_\cH^2 + (1+2L^2){\sigma'}^2 \leq \tau^2,
\end{equation*}
and similarly $\sigma_Z^2 \leq 2B_\cH^2 + (1+2L^2)\sigma^2 \leq \tau^2$. For the gradient term, since $\nabla c$ is $M$-Lipschitz we have $\|\nabla c(w)\| \leq \|\nabla c(0,0)\| + M\|w\|$, giving
\begin{align*}
    \E[\|\nabla c(W)\|^2] &\leq 2\|\nabla c(0,0)\|^2 + 2M^2 \E[\|W\|^2] \leq 2\big(\|\nabla c(0,0)\|^2 + M^2 \tau^2\big).
\end{align*}
Substituting into \eqref{eq:smooth-split} using \eqref{eq:quadratic-bound},
\begin{align*}
    \E[c(Z)] - \E[c(W)] &\leq \bigg[\sqrt{\E[\|\nabla c(W)\|^2]} + \frac{M}{2}(\sigma_Z + \sigma_W)\bigg]\gW_2(Z, W)\\
    &\leq \Big[\sqrt{2}\big(\|\nabla c(0,0)\| + M\tau\big) + M\tau\Big]\gW_2(Z, W)\\
    &\leq 3\big(\|\nabla c(0,0)\| + M\tau\big)\,\gW_2(Z, W).
\end{align*}
Combining with \eqref{eq:pushforward-w2} and exchanging the roles of $p_\rx$ and $p'_\rx$ completes the proof.
\end{proof}

\subsection{Stability of the Mixed Risk Minimizer} \label{sec:stability}

Denote by \(\gA\), the algorithm that minimizes the mixed empirical risk:
\begin{equation*}
    \gA(\rmS) := \operatorname{argmin}_{h \in \cH} \gR_{\lambda}(h, \rmS) = h_\rmS.
\end{equation*}

We develop stability bounds utilizing the loss on synthetic data as a source of regularization. The methodology modifies an approach used in the proof of Proposition 6 of \citep{farghly2026implicit}, using function space convexity to obtain stability bounds in \(L^2(p'_\rx)\) (Lemma \ref{lem:global_stability}) followed by a change of measure argument (Lemma \ref{lem:lipschitz-stability}).

\begin{lemma}\label{lem:smoothnessbound}
Suppose that the function \(f: \Y \to \R\) is \(M\)-smooth, then for any \(y \in \Y\),
\begin{equation*}
    f(y) - f^* \geq \frac{1}{2M} \| \nabla f(y) \|^2,
\end{equation*}
where \(f^* := \inf_{y \in \Y} f(y)\).
\end{lemma}
\begin{proof}
Let \(\langle \cdot, \cdot \rangle\) denote the inner product associated with the space \(\Y\). From smoothness, it follows that \(f\) is differentiable. Setting \(z = y - \frac{1}{M} \nabla f(y)\), it further follows from smoothness that,
\begin{align*}
    f(z) - f(y) &\leq \langle \nabla f(y), z -y \rangle + \frac{M}{2} \|z - y\|^2\\
    &\leq - \frac{1}{M} \langle \nabla f(y), \nabla f(y) \rangle + \frac{1}{2M} \|\nabla f(y)\|^2\\
    &\leq - \frac{1}{2M} \|\nabla f(y)\|^2.
\end{align*}
Rearranging and using the fact that \(f(z) \geq f^*\) leads to the bound in the statement.
\end{proof}

\begin{lemma}\label{lem:global_stability}
Suppose Assumption \ref{ass:loss-func} holds, then for any \(i \in \{1, ..., N\}\) it holds that,
\begin{equation*}
    \E \bigg [ \int \| h(x) - h^i(x) \|^2 p'_\rx(dx) \bigg ] \leq \frac{8M}{m^2 \lambda} \E[\gR_{\lambda}(h, \rmS)] + \frac{4\sqrt{2 M} D (1-\lambda)}{mN\lambda} \Big ( \E[\Ell_\rmS(h)]^{1/2} + \E[\Ell_\X(h)]^{1/2} \Big ),
\end{equation*}
where \(h = \gA(\rmS), h^i = \gA(\rmS^i)\), \(\rmS^i\) is as defined in Lemma \ref{lem:stability}.
\end{lemma}
\begin{proof}
Define the measures,
\begin{gather*}
    \hat{q}(d\rx) := \frac{1-\lambda}{N} \sum_{x_j \in \rmS} \delta_{\rx_j}(d\rx) + \lambda p'_\rx(d\rx), \qquad \tilde{q}(d\rx) := \frac{1-\lambda}{N} \sum_{\rx_j \in \rmS^i} \delta_{\rx_j}(d\rx) + \lambda p'_\rx(d\rx).
\end{gather*}
Using the strong convexity of \(c\) we obtain,
\begin{equation*}
    \langle h^i(\rx) - h(\rx), \nabla_1 c(h^i(\rx), \rx) \rangle \geq c(h^i(\rx), \rx) - c(h(\rx), \rx) + \frac{m}{2} \|h^i(\rx) - h(\rx)\|^2.
\end{equation*}
which when integrated with respect to \(\hat{q}\), leads to
\begin{equation*}
    \int \langle h^i(\rx) - h(\rx), \nabla_1 c(h^i(\rx), \rx) \rangle \, \hat{q}(d\rx) \geq \gR_{\lambda}(h^i, \rmS) - \gR_{\lambda}(h, \rmS) + \frac{m}{2} \int \|h^i(\rx) - h(\rx)\|^2 \hat{q}(d\rx).
\end{equation*}
The right-hand side is lower bounded further using \(\gR_{\lambda}(h^i, \rmS) \geq \gR_{\lambda}(h, \rmS)\) and the left-hand side is upper bounded using,
\begin{align*}
    &\int \langle h^i(\rx) - h(\rx), \nabla_1 c(h^i(\rx), \rx) \rangle \, \hat{q}(d\rx)\\
    & \qquad =\int \langle h^i(\rx) - h(\rx), \nabla_1 c(h^i(\rx), \rx) \rangle \, \tilde{q}(d\rx)\\
    & \qquad \qquad + \frac{1-\lambda}{N} \bigg ( \langle h^i(\rx_i) - h(\rx_i), \nabla_1 c(h^i(\rx_i), \rx_i) \rangle - \langle h^i(\tilde{\rx}) - h(\tilde{\rx}), \nabla_1 c(h^i(\tilde{\rx}), \tilde{\rx}) \rangle \bigg )\\
    & \qquad \leq \bigg (\int \| h^i(\rx) - h(\rx) \|^2\, \tilde{q}(d\rx) \bigg )^{1/2}  \bigg (\int \| \nabla_1 c(h^i(\rx), \rx) \|^2\, \tilde{q}(d\rx) \bigg )^{1/2}\\
    & \qquad \qquad + \frac{D(1-\lambda)}{N} \bigg ( \|\nabla_1 c(h^i(\rx_i), \rx_i)\| + \|\nabla_1 c(h^i(\tilde{\rx}), \tilde{\rx})\| \bigg )\\
    & \qquad \leq \sqrt{2 M} \bigg (\int \| h^i(\rx) - h(\rx) \|^2\, \tilde{q}(d\rx) \bigg )^{1/2} \gR_{\lambda}(h^i, \rmS^i)^{1/2} + \frac{\sqrt{2 M} D (1-\lambda)}{N} \Big ( c(h^i(\rx_i), \rx_i)^{1/2} + c(h^i(\tilde{\rx}), \tilde{\rx})^{1/2} \Big ).
\end{align*}
The first inequality above follows from the Cauchy--Schwarz inequality and the second follows from Lemma~\ref{lem:smoothnessbound}. 

This results in the bound,
\begin{align*}
    \int \|h^i(\rx) - h(\rx)\|^2 \hat{q}(d\rx) &\leq \frac{2 \sqrt{2M}}{m} \bigg (\int \| h^i(\rx) - h(\rx) \|^2\, \tilde{q}(d\rx) \bigg )^{1/2} \gR_{\lambda}(h^i, \rmS^i)^{1/2}\\
    & \qquad + \frac{2 \sqrt{2 M} D (1-\lambda)}{mN} \Big ( c(h^i(\rx_i), \rx_i)^{1/2} + c(h^i(\tilde{\rx}), \tilde{\rx})^{1/2} \Big ).
\end{align*}
Taking the expectation, we use the fact that \((h, h^i, \rmS^i)\) shares the same law as \((h^i, h, \rmS)\) and thus can be exchanged, as well as the symmetry of the algorithm \(\gA\) under permutations in the dataset, to obtain,
\begin{align*}
    \E \bigg [ \int \|h^i(\rx) - h(\rx)\|^2 \hat{q}(d\rx) \bigg ] &\leq \frac{2\sqrt{2M}}{m} \bigg ( \E \bigg [ \int \| h^i(\rx) - h(\rx) \|^2\, \hat{q}(d\rx) \bigg ] \bigg )^{1/2} \E[\gR_{\lambda}(h, \rmS)]^{1/2}\\
    & \qquad + \frac{2\sqrt{2 M} D (1-\lambda)}{mN} \Big ( \E[\Ell_\rmS(h)]^{1/2} + \E[\Ell_\X(h)]^{1/2} \Big ).
\end{align*}
By solving the quadratic equation, we deduce that this implies,
\begin{align*}
    \E \bigg [ \int \|h^i(x) - h(x)\|^2 \hat{q}(dx) \bigg ]^{1/2} &\leq \frac{\sqrt{2M}}{m} \E[\gR_{\lambda}(h, \rmS)]^{1/2}\\
    & \qquad + \sqrt{\frac{2M}{m^2} \E[\gR_{\lambda}(h, \rmS)] + \frac{2 \sqrt{2 M} D (1-\lambda)}{mN} \Big ( \E[\Ell_\rmS(h)]^{1/2} + \E[\Ell_\X(h)]^{1/2} \Big )}.
\end{align*}
This leads to the bound,
\begin{align*}
    \E \bigg [ \int \| h^i(x) - h(x) \|^2 p'_\rx(dx) \bigg ] &\leq \frac{1}{\lambda} \E \bigg [ \int \|h^i(x) - h(x)\|^2 \hat{q}(dx) \bigg ]\\
    &\leq \frac{8M}{m^2 \lambda} \E[\gR_{\lambda}(h, \rmS)] + \frac{4\sqrt{2 M} D (1-\lambda)}{mN\lambda} \Big ( \E[\Ell_\rmS(h)]^{1/2} + \E[\Ell_\X(h)]^{1/2} \Big ).
\end{align*}
\end{proof}

\begin{lemma}\label{lem:lipschitz-stability}
Suppose that \(\cH\) consists of \(L\)-Lipschitz functions, then for any \(c > 1\) and \(\delta > 0\) sufficiently small, the assumption in \Cref{eq:stability_1} is satisfied with
\begin{equation*}
    \varepsilon \leq \E[\Ell_\rmS(\gA(\rmS))] + 2 M C^{-1} \delta^{-d^*} \sup_i \E \bigg [ \int \| h(x) - h^i(x) \|^2 p_\rx(dx) \bigg ] + 4 M L^2\delta^2.
\end{equation*}
\end{lemma}
\begin{proof}
Define \(\varepsilon = \E[c(h^i(\rx_i), \rx_i) - c(h(\rx_i), \rx_i)]\), then because of the smoothness of \(c\), we obtain
\begin{align*}
    \varepsilon &\leq \E[\langle h^i(\rx_i) - h(\rx_i), \nabla_1 c(h(\rx_i), \rx_i) \rangle] + \frac{M}{2} \E[\|h^i(\rx_i) - h(\rx_i)\|^2]\\
    &\leq \sqrt{2M} \E[\|h^i(\rx_i) - h(\rx_i)\|^2]^{1/2} \E[c(h(\rx_i), \rx_i)]^{1/2} + \frac{M}{2} \E[\|h^i(\rx_i) - h(\rx_i)\|^2]\\
    &\leq \sqrt{2M} \E[\|h^i(\rx_i) - h(\rx_i)\|^2]^{1/2} \E[\Ell_\rmS(h)]^{1/2} + \frac{M}{2} \E[\|h^i(\rx_i) - h(\rx_i)\|^2]\\
    &\leq \E[\Ell_\rmS(h)] + M \E[\|h^i(\rx_i) - h(\rx_i)\|^2].
\end{align*}
Define the measure,
\begin{equation*}
    p^{\tilde{\rx}, \delta}_\rx(dx) := \mathbb{1}_{B_\delta(\tilde{\rx})}(x) \, p_\rx(B_\delta(\tilde{\rx}))^{-1} \, p_\rx(dx).
\end{equation*}

Then we can relate function evaluations to the integral over \(p'_\rx\) as follows:
\begin{align*}
    \|h(\tilde{\rx}) - h^i(\tilde{\rx})\| &\leq \bigg ( \int \|h(x) - h^i(x)\|^2 \, p^{\tilde{\rx}, \delta}_\rx(dx) \bigg )^{1/2} + \bigg ( \int \|h(x) - h(\tilde{\rx})\|^2  \, p^{\tilde{\rx}, \delta}_\rx(dx) \bigg )^{1/2}\\
    & \qquad + \bigg ( \int \|h^i(x) - h^i(\tilde{\rx})\|^2 \, p^{\tilde{\rx}, \delta}_\rx(dx) \bigg )^{1/2}\\
    &\leq p_\rx(B_\delta(\tilde{\rx}))^{-1/2} \bigg ( \int \|h(x) - h^i(x)\|^2 \, p_\rx(dx) \bigg )^{1/2} + 2 L \delta.
\end{align*}
Taking the expectation gives,
\begin{align*}
    \E[\|h(\rx_i) - h^i(\rx_i)\|^2]^{1/2} &\leq \E_{\tilde{\rx} \sim p_\rx} \Big [ p_\rx(B_\delta(\tilde{\rx}))^{-1} \Big ]^{1/2} \E \bigg [ \int \|h(x) - h^i(x)\|^2 \, p_\rx(dx) \bigg ]^{1/2} +  2 L \delta.
\end{align*}
Using Assumption \ref{ass:dim}, we obtain that for $\delta$ sufficiently small,
\begin{align*}
    \E[\|h(\rx_i) - h^i(\rx_i)\|^2]^{1/2} &\leq C^{-1/2} \delta^{-d^*/2} \E \bigg [ \int \|h(x) - h^i(x)\|^2 \, p_\rx(dx) \bigg ]^{1/2} +  2 L \delta.
\end{align*}
\end{proof}

\subsection{Proof of \texorpdfstring{\Cref{thm:gen-gap-mixed}}{theorem 3.1}} \label{app:proof-gen-gap-mixed}

\begin{manualthm}{\ref{thm:gen-gap-mixed}}
Suppose Assumptions \ref{ass:diam_H}, \ref{ass:loss-func} and \ref{ass:dim} hold, suppose that \(h_\rmS \in \argmin_{h \in \cH} \gR_{\lambda}(h, \rmS)\) and let \(\gR^\star = \min_{h \in \cH} r(h)\). Then, there exists \(N_0, \gR^\star_0, \mathcal{W}_0 > 0\) such that whenever \(N \geq N_0, \gR^\star \leq \gR^\star_0\) and \(\wass{2}{p_\rx}{p'_\rx} \leq \mathcal{W}_0\), the algorithm \(\gA(\rmS) = h_\rmS\) is uniformly stable with stability constant,
\begin{equation}\label{eq:gen-gap-mixed_0}
    \varepsilon \lesssim \E[\Ell_\rmS(h_\rmS)] + L^{\frac{2 d^*}{d^*+2}} \Delta^{\frac{2}{d^* + 2}}, \qquad \Delta = \frac{M}{C m^2 \lambda} \gR^\star + \frac{\sqrt{M} D}{C m \lambda N} + \frac{M \xi}{C m^2} \wass{2}{p_\rx}{p'_\rx}.
\end{equation}
Furthermore, we have the generalization bound,
\begin{align*}
    \E[r(h_\rmS)] \lesssim \gR^\star + \lambda\xi \wass{2}{p_\rvx}{p'_\rvx} + (1-\lambda) L^{\frac{2 d^*}{d^*+2}} \Delta^{\frac{2}{d^* + 2}}.
\end{align*}
\end{manualthm}

\begin{proof}
To obtain the stability bound in the first part of the theorem, we combine lemmas \ref{lem:global_stability} and \ref{lem:lipschitz-stability}. We optimize $\delta$ in the bound of Lemma \ref{lem:lipschitz-stability}, such that
\begin{equation*}
    \delta^{d^*+2} = \frac{d^* K}{2 L^2}, \qquad K = C^{-1} \sup_i \E \bigg [ \int \| h(x) - h^i(x) \|^2 p_\rx(dx) \bigg ],
\end{equation*}
so that for $K$ sufficiently small, we obtain stability with constant,
\begin{equation}\label{eq:gen-gap_1}
    \varepsilon \lesssim \E[\Ell_\rmS(\gA(\rmS))] + K^{\frac{2}{d^* + 2}} L^{\frac{2 d^*}{d^* + 2}}.
\end{equation}
To control $K$ we use that $h$ and $h^i$ are Lipschitz and thus,
\begin{align*}
    K &\leq C^{-1} \sup_i \bigg \{ \E \bigg [ \int \|h(\tilde{x}) - h^i(\tilde{x})\|^2 p'_\rx(d\tilde{x}) \bigg ]^{\frac{1}{2}} + 2\E \bigg [ \int \|h^i(x) - h^i(\tilde{x})\|^2 \pi(dx, d\tilde{x}) \bigg ]^{\frac{1}{2}}  \bigg \}^2\\
    &\leq 2 C^{-1} \sup_i  \E \bigg [ \int \| h(x) - h^i(x) \|^2 p'_\rx(dx) \bigg ] + 8 C^{-1} L^2 \wass{2}{p_\rx}{p'_\rx}^2
\end{align*}
which when substituting the bound from Lemma \ref{lem:global_stability}, produces,
\begin{align*}
    K &\leq \frac{16M}{C m^2 \lambda} \E[\gR_{\lambda}(h, \rmS)] + \frac{8\sqrt{2 M} D (1-\lambda)}{C m \lambda N} \Big ( \E[\Ell_\rmS(h)]^{1/2} + \E[\Ell_\X(h)]^{1/2} \Big ) + \frac{8 L^2}{C} \wass{2}{p_\rx}{p'_\rx}^2.
\end{align*}
Since $\ell$ is smooth and $\mathcal{H}$ has bounded diameter, we have that $\E[\Ell_\rmS(h)]$ and $\E[\Ell_\X(h)]$ are uniformly bounded over $h \in \mathcal{H}$. Therefore, it follows that $K$ can be taken sufficiently small as soon as $N^{-1}$, $\E[\gR_{\lambda}(h, \rmS)]$ and $ \wass{2}{p_\rx}{p'_\rx}$ are all taken sufficiently small. From Lemma \ref{lem:stability}, it follows that $\E[\Ell_\X(h)]^{1/2} \leq \E[\Ell_\rmS(h)]^{1/2} + \varepsilon^{1/2}$. Therefore, substituting back into \eqref{eq:gen-gap_1}, we obtain,
\begin{align*}
    \varepsilon &\lesssim \E[\Ell_\rmS(\gA(\rmS))] + \bigg ( \frac{M L^{d^*}}{C m^2 \lambda} \E[\gR_{\lambda}(h, \rmS)] + \frac{\sqrt{M} D L^{d^*} (1-\lambda)}{C m \lambda N} \Big ( \E[\Ell_\rmS(h)]^{1/2} + \varepsilon^{1/2} \Big ) + \frac{L^{2+d^*}}{C} \wass{2}{p_\rx}{p'_\rx}^2 \bigg )^{\frac{2}{d^* + 2}}\\
    &\lesssim \E[\Ell_\rmS(\gA(\rmS))] + \bigg ( \frac{M L^{d^*}}{C m^2 \lambda} \E[\gR_{\lambda}(h, \rmS)] + \frac{\sqrt{M} D L^{d^*} (1-\lambda)}{C m \lambda N} \E[\Ell_\rmS(h)]^{1/2} + \frac{L^{2+d^*}}{C} \wass{2}{p_\rx}{p'_\rx}^2 \bigg )^{\frac{2}{d^* + 2}}\\
    & \qquad + \bigg ( \frac{\sqrt{M} D L^{d^*} (1-\lambda)}{C m \lambda N}\bigg )^{\frac{2}{d^*+2}} \varepsilon^{\frac{1}{d^*+2}}.
\end{align*}
It then follows from an application of Young's inequality,
\begin{align*}
    \varepsilon &\lesssim \E[\Ell_\rmS(\gA(\rmS))] + \bigg ( \frac{M L^{d^*}}{C m^2 \lambda} \E[\gR_{\lambda}(h, \rmS)] + \frac{\sqrt{M} D L^{d^*} (1-\lambda)}{Cm\lambda N} \E[\Ell_\rmS(h)]^{1/2} + \frac{L^{2+d^*}}{C} \wass{2}{p_\rx}{p'_\rx}^2 \bigg )^{\frac{2}{d^* + 2}}\\
    & \qquad + \bigg ( \frac{\sqrt{M} D L^{d^*} (1-\lambda)}{C m \lambda N}\bigg )^{\frac{2}{d^*+1}}\\
    &\lesssim \E[\Ell_\rmS(\gA(\rmS))] + \bigg ( \frac{M L^{d^*}}{C m^2 \lambda} \E[\gR_{\lambda}(h, \rmS)] + \frac{\sqrt{M} D L^{d^*}}{C m \lambda N} + \frac{L^{2+d^*}}{C} \wass{2}{p_\rx}{p'_\rx}^2 \bigg )^{\frac{2}{d^* + 2}},
\end{align*}
where in the last line, we use $N \geq \frac{\sqrt{M} D L^{d^*}}{C m \lambda}$ and $(1 - \lambda) \E[\Ell_\rmS(h)] \leq \E[\gR_{\lambda}(h, \rmS)] \leq 1$, by assumption.

To simplify the bound, we also control $\gR_\lambda(h_\rmS, \rmS)$. For any $h_\star \in \gH$, by the optimality of $h_\rmS$, we have
\begin{align*}
    \gR_\lambda(h_\rmS, \rmS) &\leq \gR_\lambda(h_\star, \rmS), \\
    \E_\rmS[\gR_\lambda(h_\rmS, \rmS)] &\leq \E_\rmS[\gR_\lambda(h_\star, \rmS)], \\
    \gR_\lambda(h_\rmS, \rmS) &\leq r_\lambda(h_\star).
\end{align*}
From the definition of $r_\lambda(h_\star)$ (see \Cref{eq:r-mixed}), we can write:
\begin{align}
    \gR_\lambda(h_\rmS, \rmS) 
    &\leq (1 - \lambda) \E_{p_\rvx}[\ell(h_\star, \rvx)] + \lambda \E_{p'_\rvx}[\ell(h_\star, \rvx)] \nonumber\\
    &= r(h_\star) + \lambda \left( \E_{p'_\rvx}[\ell(h_\star, \rvx)] - \E_{p_\rvx}[\ell(h_\star, \rvx)] \right). \label{eq:before-regularity}
\end{align}
Since $\ell(h, \rvx)$ is $\xi$-Lipschitz in $\rvx$, it follows from \Cref{lem:bound-wass} that,
\begin{align*}
    \E_{p'_\rvx}[\ell(h_\star, \rvx)] - \E_{p_\rvx}[\ell(h_\star, \rvx)] \leq \xi \wass{2}{p_\rvx}{p'_\rvx},
\end{align*}
and thus,
\begin{align*}
    \gR_\lambda(h_\rmS, \rmS) \leq r(h_\star) + \xi \lambda \wass{2}{p_\rvx}{p'_\rvx}.
\end{align*}
Finally, we can choose $h_\star = \argmin_{h \in \gH} r(h)$ to tighten the bound. With this, we conclude the proof of the bound in \eqref{eq:gen-gap-mixed_0}, which holds once $N^{-1}, \gR^\star, \wass{2}{p_\rx}{p'_\rx}$ are all taken sufficiently small.

For the second bound of the theorem, we use the following decomposition to upper bound the generalization error:
\begin{align*}
    r(h) &= \left[r(h) - r_\lambda(h)\right] + \left[r_\lambda(h) - \gR_\lambda(h, \rmS) \right] + \gR_\lambda(h, \rmS).
\end{align*}
We proceed by bounding $r(h) - r_\lambda(h)$. We compute
\begin{align*}
    r(h) - r_\lambda(h) 
    &= \E_{p_\rvx}[\ell(h, \rvx)] - (1 - \lambda)\E_{p_\rvx}[\ell(h, \rvx)] - \lambda \E_{p'_\rvx}[\ell(h, \rvx)] \\
    &= \lambda \left( \E_{p_\rvx}[\ell(h, \rvx)] - \E_{p'_\rvx}[\ell(h, \rvx)] \right) \\
    &\leq \lambda \xi\, \mathcal{W}_2(p_\rvx, p'_\rvx),
\end{align*}
where the inequality follows from Lemma~\ref{lem:bound-wass}.

To bound $r_\lambda(h) - \gR_\lambda(h, \rmS)$, we start by substituting the definition of each term and simplifying them. We have:
\begin{align*}
    r_\lambda(h) - \gR_\lambda(h, \rmS))
    &= (1-\lambda) \left(\E_{\rmS, \rvx'}\left[\ell(h_\rmS, \rvx')\right] - \E_{\rmS, i}\left[\ell(h_\rmS, \rvx_i)\right]\right)\\
    &= (1-\lambda) \left(\E_{\rmS, \rvx'}\left[\ell(h_{\rmS'}, \rvx_i)\right] - \E_{\rmS, i}\left[\ell(h_\rmS, \rvx_i)\right]\right)\,,
\end{align*}
where the first equality is due to the fact that $\lambda \E_{p'_\rvx}[\ell(h, \rvx)]$ is common in both terms. The second equality is by the definition of each term, and the fact that $\rvx' \indep \rmS$. Note that $i \sim \text{Unif}([N])$. The final line results from defining $\rmS' = \rmS \cup \{\rvx'\} \setminus \{\rvx_i\}$, which is a neighboring set to $\rmS$. Now, assuming that we have $\varepsilon$-uniformly stable algorithm $\gA$, then we can write
\begin{align*}
    r_\lambda(h) - \gR_\lambda(h, \rmS) &= (1-\lambda) \E_{\rmS, \rvx', i}\left[\ell(h_{\rmS'}, \rvx_i) - \ell(h_{\rmS}, \rvx_i)\right]\\
    &\leq (1-\lambda)\varepsilon\,.
\end{align*}
These combine to give the bound,
\begin{align*}
    \E[r(h) - r(h_\star)] \leq 2 \lambda \xi\, \mathcal{W}_2(p_\rvx, p'_\rvx) + (1 - \lambda) \varepsilon.
\end{align*}
Substituting \eqref{eq:gen-gap-mixed_0} into the bound above, we obtain,
\begin{align*}
    \E[r(h_\rmS) - r(h_\star)] &\lesssim 2\lambda\xi\,\mathcal{W}_2(p_\rvx, p'_\rvx) 
    + (1-\lambda)\E[\Ell_\rmS(\gA(\rmS))] 
    + (1-\lambda) L^2 \Delta^{\frac{2}{d^*+2}}.
\end{align*}
Since $(1-\lambda)\E[\Ell_\rmS(h)] \leq \E[\gR_\lambda(h, \rmS)]$, and using $\E[\gR_\lambda(h_\rmS, \rmS)] \leq \gR^\star + \xi\lambda\wass{2}{p_\rvx}{p'_\rvx}$ from above, we obtain,
\begin{align*}
    (1-\lambda)\E[\Ell_\rmS(\gA(\rmS))] \leq \gR^\star + \xi\lambda\wass{2}{p_\rvx}{p'_\rvx}.
\end{align*}
\end{proof}

\section{THEORETICAL RESULTS AND DISCUSSIONS OF DOMAIN SHIFT} \label{sec:app-domain-shift}

\subsection{Proof of \texorpdfstring{\Cref{thm:kernel-domain}}{theorem 5.1}}\label{app:proof-kernel-domain}

\begin{manualthm}{\ref{thm:kernel-domain}}
    Under \Cref{ass:eigendecay}, for the kernel regression problem defined in \Cref{eq:kernel-reg} and any fixed regularization parameter \( \lambda > 0 \), the test error under domain shift satisfies the bound:
    \begin{align*}
        \mathcal{R}_N(\lambda; g) = \mathcal{O}\left(\lambda^{-2} \gD(f_\star, \Tilde{f}) + \lambda^{2 - \frac{1}{4r}} \mathcal{D}(f_\star, g) + \frac{\sigma^2 + \gD(f_\star, \Tilde{f}) + \gD(f_\star, g)}{N\lambda^2}\right),
    \end{align*}
    where \( \gD(\cdot, \cdot) \) denotes the distributional discrepancy, as in \Cref{thm:kernel-error-rate}.
\end{manualthm}

\begin{proof}
    The proof follows the proof of \Cref{thm:kernel-error-rate} in \Cref{app:kernel-error-rate}, using the bias-variance decomposition. We note that the variance term remains the same as it only depends on the noise of the data, while the bias term will have the dependency on all three terms of $f_\star, \Tilde{f}$ and $g$. Let us start with the formal definition of our bias term, similar to \Cref{app:kernel-error-rate}:
    \begin{align*}
        \mathcal{B}^2 &= \E_\rvx \left[(f_\star - \E_\varepsilon\left[f_N\right])(\rx)\right]^2\\
        &\leq \E_\rmS\left[\|\E_\varepsilon\left[f_N\right] - f_\lambda\|_{\gH_K}^2\right] + \|f_\star - f_\lambda\|_{p_\rvx}^2 \\
        &\leq \E_\rmS\left[\|\E_\varepsilon\left[f_N\right] - \tilde{f}_\lambda\|_{\gH_K}^2\right] + \|\tilde{f}_\lambda - f_\lambda\|_{p_\rvx}^2 + \|f_\star - f_\lambda\|_{p_\rvx}^2\,,
    \end{align*}
    where $\tilde{f}_\lambda$, and $f_\lambda$ are the population limits of $f_N$ with repsect to $\tilde{f}$ and $f_\star$, i.e. $\tilde{f}_\lambda = g + (T_K + \lambda I)^{-1}T_K (\Tilde{f} - g)$, and $f_\lambda = g + (T_K + \lambda I)^{-1}T_K (f_\star - g)$. We have previously studied the first term in \Cref{eq:bias-sample}. Therefore, we have:
    \begin{align*}
        \E_\rmS\left[\|\E_\varepsilon\left[f_N\right] - f_\lambda\|_{\gH_K}^2\right] \leq \frac{3\kappa^4 \|\Tilde{f} - g\|_{\gH}^2}{N\lambda^2} \leq \frac{3\kappa^4}{N\lambda^2}\left(\|f_\star - \Tilde{f}\|_{\gH}^2 + \|\Tilde{f} - g\|_{\gH}^2\right)\leq \frac{3\kappa^4 }{N\lambda^2} \left(\gD(f_\star, \Tilde{f}) + \gD(f_\star, g)\right) \,.
    \end{align*}
    The last term follows from \Cref{eq:bias-rate}, and the computation after that. Therefore, we have:
    \begin{align*}
        \|f_\star - f_\lambda\|_{p_\rvx}^2 &\leq \kappa^2 \|f_\star - f_\lambda\|_{\gH}^2\\
        &\leq \kappa^2 \sum_{j = 1}^\infty \frac{\lambda^2}{(\mu_j + \lambda)^2} (\theta_j - \omega_j)^2\\
        &\leq \kappa^2\lambda^2 \sqrt{\left(\sum_{j = 1}^\infty \frac{\mu_j^2}{(\mu_j + \lambda)^2} \right)\left(\sum_{j = 1}^\infty \frac{1}{\mu_j^2}(\theta_j - \omega_j)^2\right)}\\
        &= \gO\left(\kappa^2 \lambda^{2 - \frac{1}{4r}} \mathcal{D}(f_\star, g) C_r\right)\,.
    \end{align*}
    Now, it's only enough to provide a bound for the second term. We have:
    \begin{align*}
        \|\tilde{f}_\lambda - f_\lambda\|_{p_\rvx}^2 &\leq \kappa^2 \|\tilde{f}_\lambda - f_\lambda\|_{\gH}^2\\
        &= \kappa^2 \|(T_K + \lambda I)^{-1}T_K (f_\star - \tilde{f})\|_{\gH}^2\\
        &\leq \kappa^2 \frac{\kappa^2}{\lambda^2}\|f_\star - \tilde{f}\|_{\gH}^2\\
        &\leq \frac{\kappa^4}{\lambda^2}\gD(f_\star, \tilde{f})\,.
    \end{align*}
    Combining all the terms completes the proof.
\end{proof}

\subsection{Proof of \texorpdfstring{\Cref{thm:gen-gap-out-domain}}{theorem 5.2}}
\label{app:proof-gen-out-domain}

\begin{manualthm}{\ref{thm:gen-gap-out-domain}}
Suppose that Assumptions \ref{ass:diam_H}--\ref{ass:dim} hold and that \(h_\rmS \in \argmin_{h \in \cH} \gR_{\lambda}(h, \rmS)\). Then, whenever \(N \geq N_0, \gR^\star \leq \gR^\star_0\) and \(\wass{2}{p_\rx}{p'_\rx} \leq \mathcal{W}_0\), the generalization gap under the domain shift satisfies
\begin{align*}
    \E[r(h_\rmS)] &\lesssim r^\star + \lambda\xi \wass{2}{p^\star_\rvx}{p'_\rvx} + (1-\lambda)\xi \wass{2}{p^\star_\rvx}{p_\rvx} + (1-\lambda) L^{\frac{2d^*}{d^*+2}} \Delta^{\frac{2}{d^* + 2}},
\end{align*}
where \(r^\star = \min_{h \in \cH} r^\star(h)\) is minimum population risk minimizer of the target domain and \(\Delta, \xi, N_0, \gR^\star_0, \mathcal{W}_0\) are as in Theorem \ref{thm:gen-gap-mixed}.
\end{manualthm}

\begin{proof}
    For any $h \in \gH$, we show $r^\star(h)$, and $r^\star_\lambda(h)$ as
    \begin{align}\label{eq:out-dec}
        r^\star(h) = \E_{p^\star_\rvx}\left[\ell(h, \rvx)\right], \qquad r^\star_\lambda(h) = (1-\lambda)r^\star(h) +  \E_{p'_\rvx}\left[\ell(h, \rvx)\right]\,.
    \end{align}
    We now have the following decomposition for the generalization error:
    \begin{align*}
        r^\star(h) = (r^\star(h) - r^\star_\lambda(h)) + (r^\star_\lambda(h) - r_\lambda(h)) + (r_\lambda(h) - \gR_\lambda(h)) + \gR_\lambda(h)\,.
    \end{align*}
    Similar to \Cref{app:proof-gen-gap-mixed}, we bound each of the terms separately. We have:
    \paragraph{Bounding $r^\star(h) - r^\star_\lambda(h)$:} Let us expand the term by the definition of each component:
    \begin{align*}
        r^\star(h) - r^\star_\lambda(h) &= \E_{p^\star_\rvx}\left[\ell(h, \rvx)\right] - \left((1-\lambda)r^\star(h) +  \E_{p'_\rvx}\left[\ell(h, \rvx)\right]\right)\\
        &= \lambda \left( \E_{p^\star_\rvx}\left[\ell(h, \rvx)\right] - \E_{p'_\rvx}\left[\ell(h, \rvx)\right]\right)\\
        &\leq \lambda \xi \wass{2}{p^\star_\rvx}{p'_\rvx}\,,
    \end{align*}
    where the last inequality is by \Cref{lem:bound-wass}.
    \paragraph{Bounding $r^\star_\lambda(h) - r_\lambda(h)$:} We again use the definitions:
    \begin{align}
    r^\star_\lambda(h) - r_\lambda(h) &= \left((1-\lambda)r^\star(h) +  \E_{p'_\rvx}\left[\ell(h, \rvx)\right]\right) - \left( (1-\lambda)r(h) +  \E_{p'_\rvx}\left[\ell(h, \rvx)\right] \right)\nonumber\\
    &= (1-\lambda) (r^\star(h) - r(h))\nonumber\\
    &= (1-\lambda) \left(\E_{p^\star_\rvx}\left[\ell(h, \rvx)\right] - \E_{p_\rvx}\left[\ell(h, \rvx)\right]\right) \nonumber\\
    &\leq (1-\lambda) \xi \wass{2}{p^\star_\rvx}{p_\rvx}\,,\label{eq:out-dist}
    \end{align}
    where we have once again used \Cref{lem:bound-wass}.
    \paragraph{Bounding $r_\lambda(h) - \gR_\lambda(h)$:} Similar to \Cref{app:proof-gen-gap-mixed}, we refer to this term as the stability term. Note that all the conditions for \Cref{lem:stability} hold here, therefore, this term is the same as \Cref{app:proof-gen-gap-mixed} since the stability is uniform. Thus, $r_\lambda(h) - \gR_\lambda(h) \leq (1-\lambda)\varepsilon$ for:
    \begin{align}\label{eq:eps}
        \varepsilon \lesssim \frac{1}{1-\lambda}\E[\gR_\lambda(h_\rmS)] + L^{\frac{2d^*}{d^*+2}} \Delta^{\frac{2}{d^* + 2}}.
    \end{align}
    We now only need to bound $\gR_\lambda(h_\rmS)$, for both \Cref{eq:out-dec,eq:eps}.
    \paragraph{Bounding $\gR_\lambda(h_\rmS)$:} Since $h_\rmS = \argmin_{h \in \gH} \gR_\lambda(h, \rmS)$ is the empirical minimizer, for any $h' \in \gH$, by optimality of $h_\rmS$, we have
\begin{align*}
    \gR_\lambda(h_\rmS, \rmS) &\leq \gR_\lambda(h', \rmS), \\
    \E_\rmS[\gR_\lambda(h_\rmS, \rmS)] &\leq \E_\rmS[\gR_\lambda(h', \rmS)], \\
    \gR_\lambda(h_\rmS, \rmS) &\leq r_\lambda(h')\,.
\end{align*}
Now, let $h' = \argmin_{h \in \gH} r_\lambda(h)$. Then, for any $h_\star \in \gH$, we have
\begin{align*}
    \gR_\lambda(h_\rmS, \rmS) &\leq r_\lambda(h')\\
    &\leq r^\star_\lambda(h_\star) + (1-\lambda) \xi \wass{2}{p^\star_\rvx}{p_\rvx}\\
    &\leq r^\star(h_\star) + \lambda \left(\E_{p^\star_\rvx}[\ell(h_\star, \rvx)] - \E_{p'_\rvx}[\ell(h_\star, \rvx)]\right) + (1-\lambda) \xi \wass{2}{p^\star_\rvx}{p_\rvx}\\
    &\leq r^\star(h_\star) + \lambda \xi \wass{2}{p^\star_\rvx}{p'_\rvx} + (1-\lambda)\xi \wass{2}{p^\star_\rvx}{p_\rvx}\,,
\end{align*}
where the first inequality is by \Cref{eq:out-dist}, and the second inequality is from the definition and the last one is by \Cref{lem:bound-wass}. Now, let $h_\star = \argmin_{h \in \gH} r^\star(h)$. Combining all bounds together completes the proof.
\end{proof}

\section{EXPERIMENTAL SETUP}\label{sec:all-exp}
\subsection{Optimal Regularization in Kernel Ridge Regression}
\label{app:kernel-experiments}

We study a nonparametric regression problem wherein the ground truth function \( f_\star \) and an auxiliary function \( g \) are both defined as truncated series expansions in an orthonormal sine basis, with polynomially decaying coefficients to encode varying degrees of smoothness. The target function is given by \( f_\star(x) = \sum_{j=1}^{T_f} (j+1)^{-rs} \sin(\pi(j+1)x) \), while \( g \) is constructed analogously using a decay rate \( s'\) over the first $T_g$ terms. Training data consists of \( N = 15 \) \gls{iid} samples \( \{x_i\}_{i=1}^n \) drawn uniformly from \([0, 3]\), with noisy observations \( y_i = f_\star(x_i) + \varepsilon_i \), where \( \varepsilon_i \sim \mathcal{N}(0, 0.1) \), alongside noiseless evaluations of \( g(x_i) \). We employ our modified kernel ridge regression (\Cref{lem:kernel-reg}) method using a Mercer kernel with eigenvalue decay \( \mu_j \asymp j^{-2r} \), incorporating \( g \) as a regularization term to enhances the standard kernel estimator. The predictive performance is evaluated on a dense test grid (test set of 500 points) by computing the empirical \( L_2 \)-distance between the learned function \( f_N \) and the true function \( f_\star \). This procedure is repeated across a logarithmically spaced range of regularization parameters \( \lambda \in [10^{-10}, 10^{10}] \). In addition, we compute the theoretically optimal regularization parameter by minimizing an upper bound derived from the distance \(\gD(f_\star, g)\), which depends explicitly on the eigendecay and coefficient mismatch between \( f_\star \) and \( g \), based on \Cref{thm:kernel-error-rate}.

\begin{figure}[ht]
    \centering
     \begin{subfigure}[b]{0.49\textwidth}
         \centering
         \includegraphics[width=1.\textwidth]{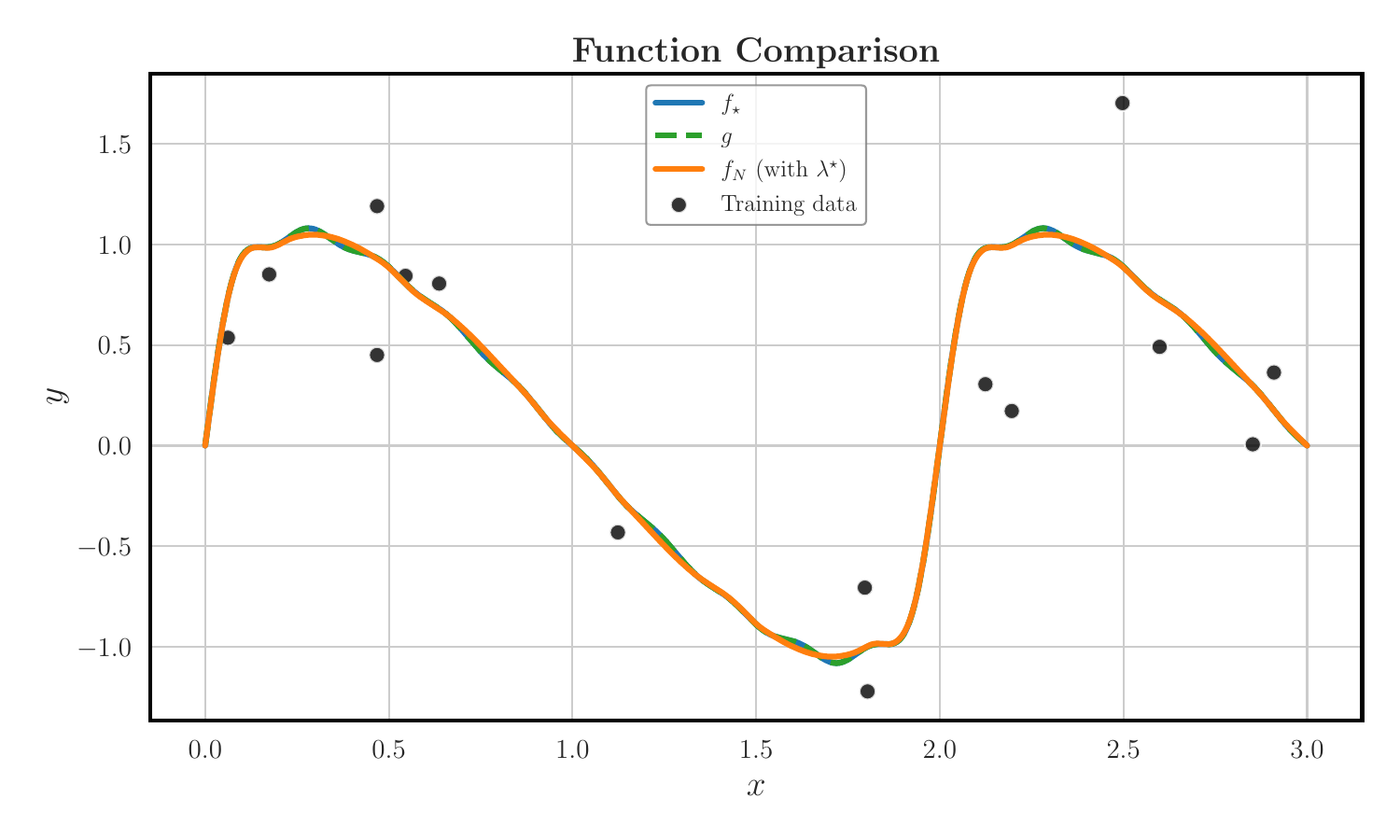}
         \caption{Function comparisons.}
         \label{fig:function-comparison-zero}
     \end{subfigure}
     \hfill
     \begin{subfigure}[b]{0.49\textwidth}
         \centering
         \includegraphics[width=1.\textwidth]{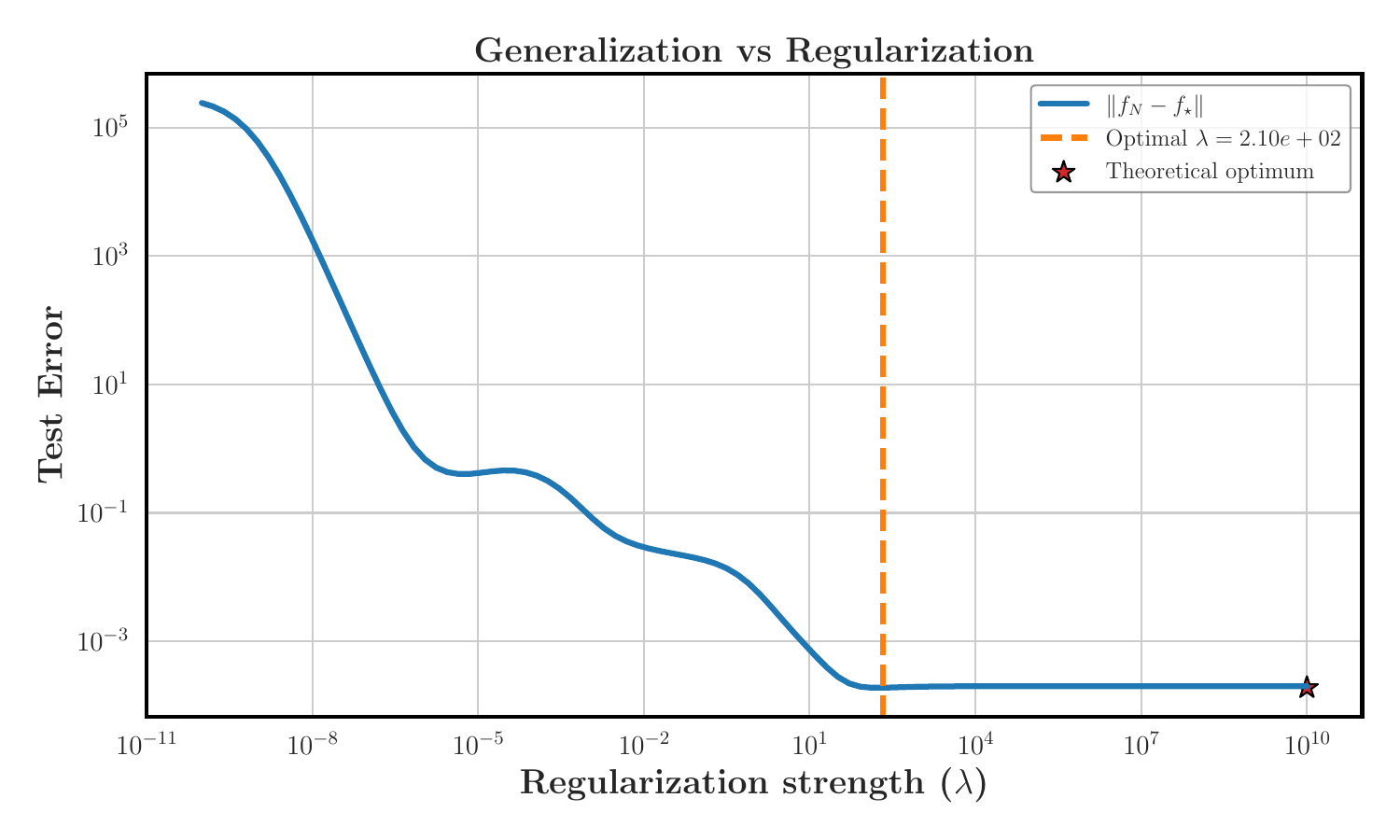}
         \caption{Generalization vs Regularization.}
         \label{fig:generalization-vs-regularization-zero}
     \end{subfigure}
    \caption{
        \emph{(a)} Comparison of the true function $f_\star$ (blue), the synthetic generator $g$ (green), and the estimated function $f_N$ (orange), obtained via \Cref{lem:kernel-reg}, with parameters $r = 2.0$, $s = 0.8$, and $s' = 0.8$. Since $g = f_\star$ in this setting, the RKHS distance is zero and all curves coincide. \emph{(b)} Prediction error $\|f_N - f_\star\|_{L_2}$ as a function of the regularization strength $\lambda$. As expected, there is no U-shaped behaviour since the generator fully matches the true distribution. The theoretical optimum selects a large $\lambda$ (star marker), while the empirical optimum (dashed orange line) selects a smaller value due to numerical precision limits.
    }
    \label{fig:kernel_no_dist}
\end{figure}

\begin{figure}[ht]
    \centering
     \begin{subfigure}[b]{0.49\textwidth}
         \centering
         \includegraphics[width=1.\textwidth]{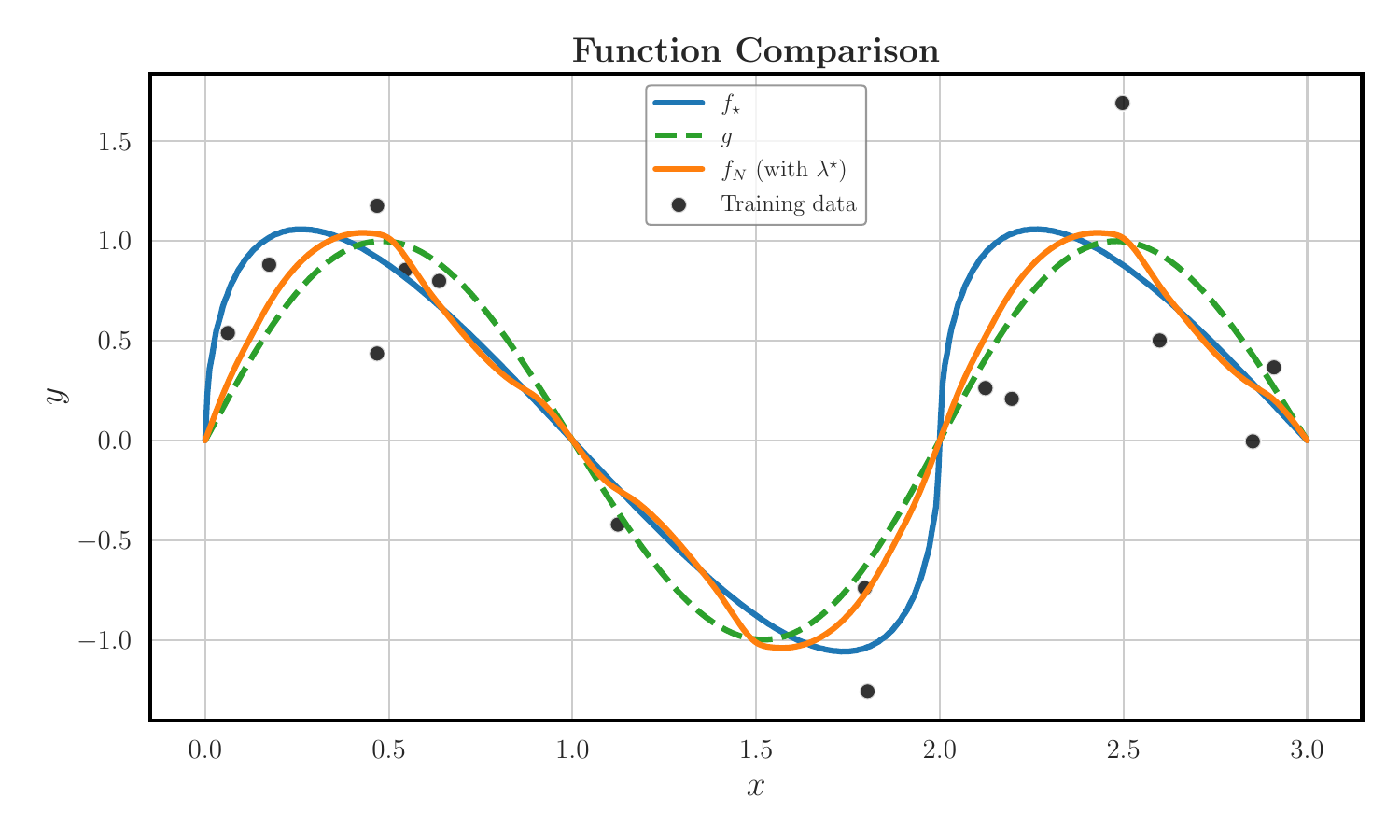}
         \caption{Function comparisons.}
         \label{fig:function-comparison-big}
     \end{subfigure}
     \hfill
     \begin{subfigure}[b]{0.49\textwidth}
         \centering
         \includegraphics[width=1.\textwidth]{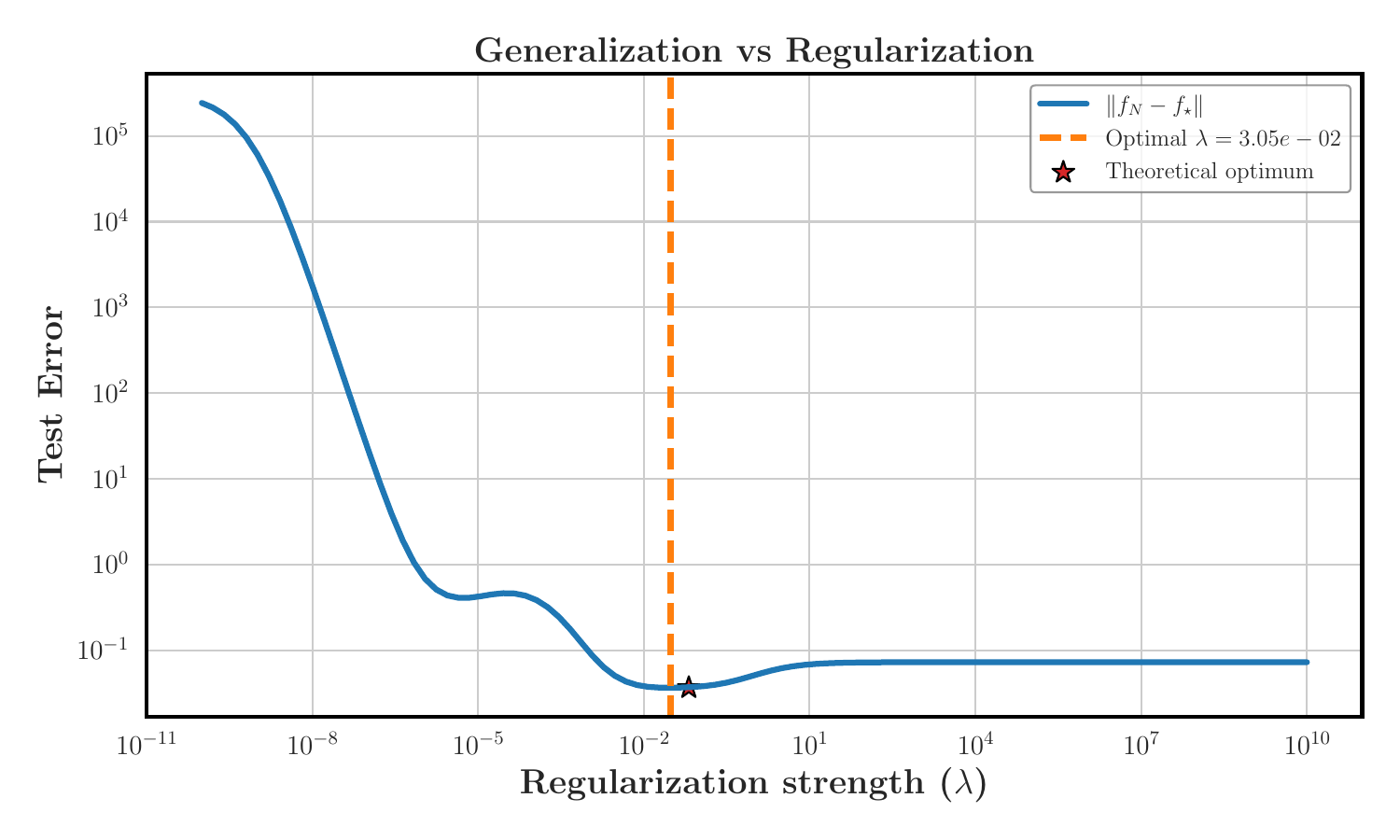}
         \caption{Generalization vs Regularization.}
         \label{fig:generalization-vs-regularization-big}
     \end{subfigure}
    \caption{
        \emph{(a)} Comparison of the true function $f_\star$ (blue), the synthetic generator $g$ (green), and the estimated function $f_N$ (orange), obtained via \Cref{lem:kernel-reg}, with parameters $r = 2.0$, $s = 0.8$, and $s' = 2.5$. The distance between the functions is larger compared to \Cref{fig:lambda_opt}. \emph{(b)} Prediction error $\|f_N - f_\star\|_{L_2}$ as a function of the regularization strength $\lambda$. A clear U-shaped curve is observed, and while the theoretically optimal $\lambda$ (star marker) slightly overestimates the empirical optimum (dashed orange line), the difference is negligible.
    }
    \label{fig:kernel_high_dist}
\end{figure}

Our implementation uses SciPy for numerical integration and optimization, with special care given to numerical stability through pseudo-inverses and adaptive regularization. To capture the effect of the difference between $f_\star$ and $g$, we run the experiment with various values as depicted in \Cref{table:kernel-setting}. \Cref{fig:lambda_opt,fig:kernel_no_dist,fig:kernel_high_dist} illustrate the impact of distributional alignment between the true function $f_\star$ and the synthetic generator $g$ on the behaviour of the estimated function $f_N$ and the choice of regularization strength $\lambda$. In \Cref{fig:kernel_no_dist}, the synthetic generator perfectly matches the true distribution ($s = s'$, and $T_f = T_g$), resulting in no discrepancy between $f_\star$, $g$, and $f_N$. Consequently, the prediction error $\|f_N - f_\star\|_{L_2}$ is minimized for the largest possible regularization strength, and our algorithm successfully selects this value. In contrast, \Cref{fig:kernel_high_dist} considers a case with distribution mismatch (large difference between $s, s'$, and $T_f, T_g$), leading to larger discrepancies between the functions. This results in a characteristic U-shaped prediction error curve, as shown in \Cref{fig:generalization-vs-regularization-big}. While the theoretically chosen regularization strength (star marker) slightly overestimates the empirical optimum (dashed orange line), the difference remains negligible, demonstrating the robustness of our theoretical bound under mismatch. The experimental details are shown in \Cref{table:kernel-setting}.

\begin{table}[ht]
\caption{Effective parameters for the modified kernel regression.}
\label{table:kernel-setting}
\centering
\begin{tabular}{ccccccc}
\toprule
$r$ & $s$ & $s'$ & $T_f$ & $T_g$ & $D(f_\star, g)$ & $\text{Figure}$ \\
\midrule
2.0 & 0.8 & 0.8 & 100 & 100 & 0.0000 & \Cref{fig:kernel_no_dist}\\
2.0 & 0.8 & 1.5 & 100 & 10 & 737.65 & \Cref{fig:lambda_opt}\\
2.0 & 0.8 & 2.5 & 100 & 10 & 15509.16 & \Cref{fig:kernel_high_dist}\\
\bottomrule
\end{tabular}
\end{table}

\FloatBarrier

\subsection{Natural Images on CIFAR-10}\label{app:cifar-experiments}
We investigate the effect of synthetic data on classification performance using a conditional diffusion model. Specifically, we train a diffusion model on CIFAR-10 to generate class-conditional synthetic images, which are then used to augment the real training set. We compare two classifiers: one trained solely on real data, and another trained on a mixture of real and synthetic samples. Performance is evaluated across varying synthetic-to-real data ratios, and validation accuracy is reported for each configuration. The real dataset used to train the diffusion model is disjoint from the one used for training and validating the classifier, allowing us to isolate the effect of synthetic data augmentation. Detailed experimental settings are provided in \Cref{subsec:cifar-experiments}.

\begin{figure}[ht]
    \centering
     \begin{subfigure}[b]{0.49\textwidth}
         \centering
         \includegraphics[width=1.\textwidth]{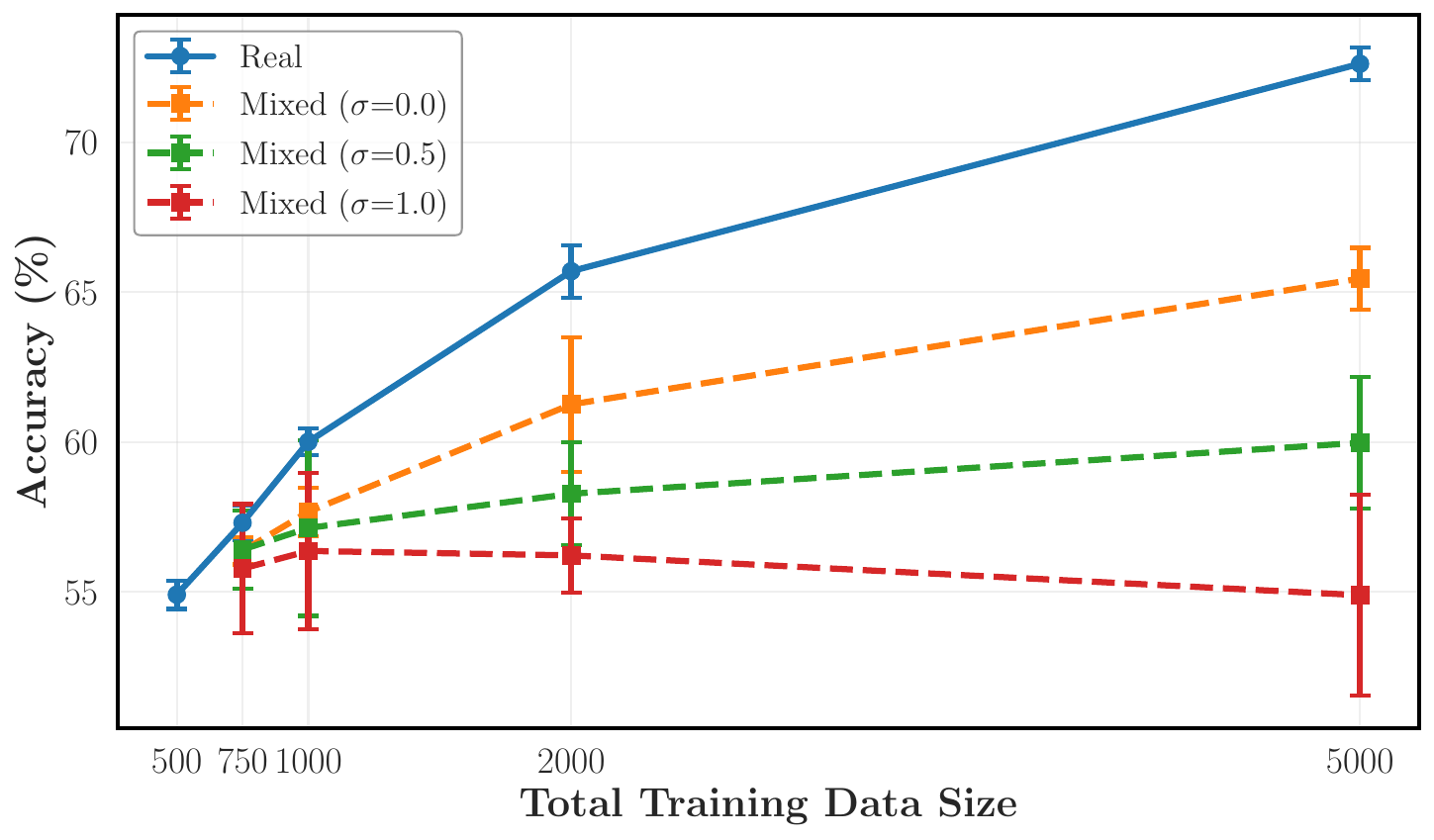}
         \label{fig:overall-accuracy}
     \end{subfigure}
     \hfill
     \begin{subfigure}[b]{0.7\textwidth}
         \centering
         \includegraphics[width=0.8\textwidth]{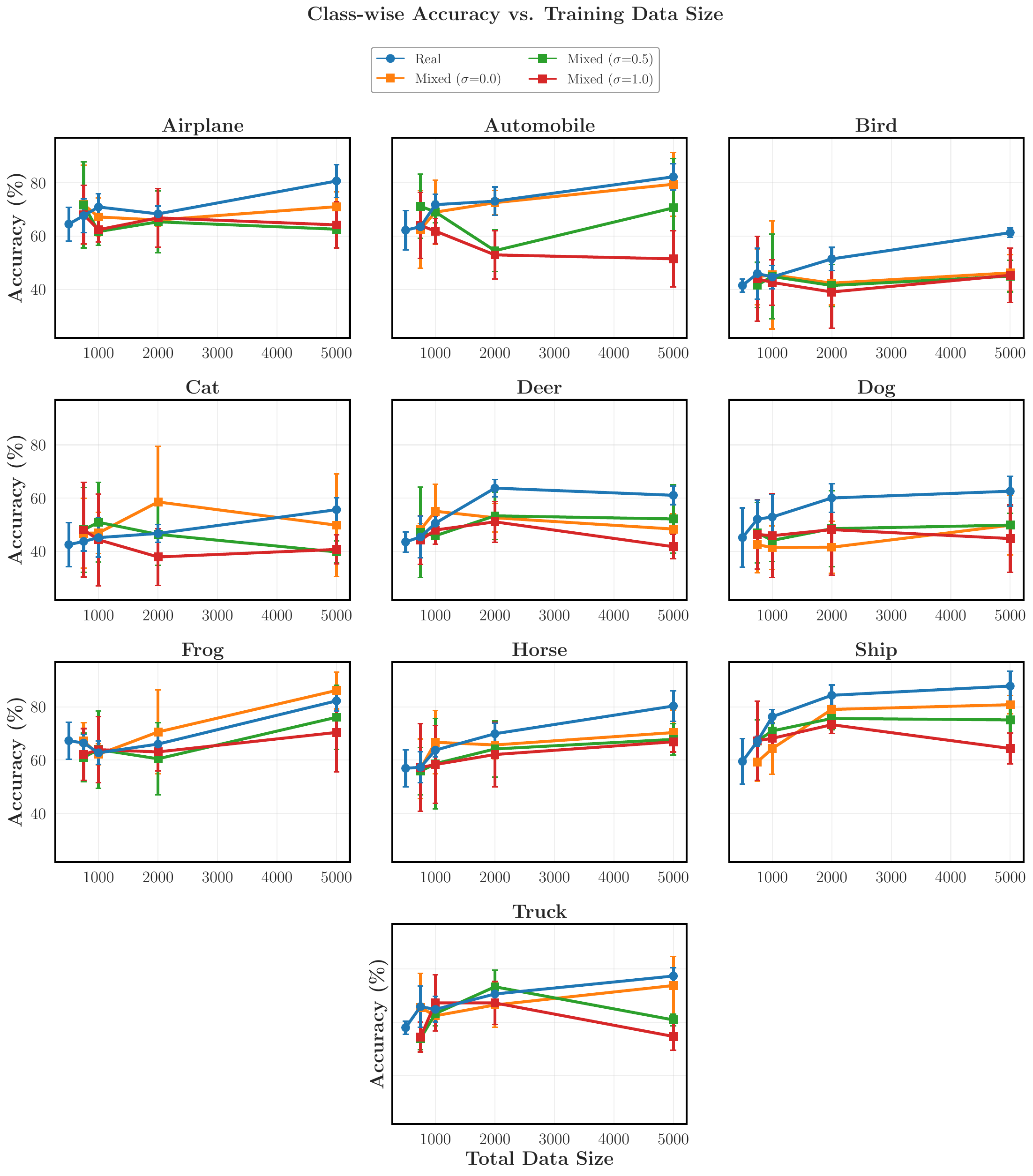}
         \label{fig:class-wise-accuracy}
     \end{subfigure}
    \caption{
        (a) Average accuracy vs. training data size. Increasing the amount of real data (blue line) consistently increases the accuracy of the classification, while for the mixed-data it depends on the quality of the generated samples. (b) Accuracy of each class vs. training data size. We observe a similar pattern, however noisier.
    }
    \label{fig:cifar-plots}
\end{figure}

In \Cref{fig:cifar-plots}, we observe that classification accuracy improves with increasing amounts of mixed training data when the distributional distance between the synthetic generator and real data is small (orange line in \Cref{fig:overall-accuracy}). In contrast, for generators with moderate to high distributional distance (i.e., lower quality - see green and red lines), we observe diminishing returns or even performance degradation. This follows our insights from \Cref{sec:generalization}. Similar trends are observed at the class level, although the results are noisier due to the reduced amount of data per class—approximately one-tenth of the total. These results indicate that the trained diffusion model captures different classes with varying fidelity, which in turn affects per-class generalization. This highlights an important practical consideration: when class-wise generalization is a priority, it is crucial to ensure that the synthetic data generator performs well not only in aggregate but also across different classes or groups.

\subsubsection{Experimental Details for CIFAR-10} \label{subsec:cifar-experiments}

\paragraph{Dataset and preprocessing} We conduct experiments on CIFAR-10 \cite{krizhevsky2009learning}, a dataset of 60,000 colour images (32$\times$32 pixels) across 10 object categories, with 50,000 training and 10,000 test samples. For each run, we stratify the training set to construct three disjoint subsets: a labelled training set $\mathcal{D}_{\text{train}}$ containing $N$ examples, a validation set $\mathcal{D}_{\text{val}}$ of 5,000 examples, and a separate set $\mathcal{D}_{\text{diff}}$ of 50,000 examples used for training the diffusion model. Stratified sampling ensures class balance across all subsets. All images are linearly rescaled to the $[-1, 1]$ range. During classifier training, we apply random horizontal flips as the only form of data augmentation. No augmentations are used during diffusion model training or validation.

\paragraph{Conditional diffusion model} Our synthetic data generator is a class-conditional diffusion model trained on $\mathcal{D}_{\text{diff}}$. The architecture is a UNet2D with six downsampling and upsampling blocks, using channel sizes [128, 128, 256, 256, 512, 512]. Self-attention layers are included at the 16$\times$16 spatial resolution. Class conditioning is achieved via a learnable embedding table of dimension 512. We train the model using a linear noise schedule over $T = 1000$ diffusion steps. Optimization is performed with AdamW using a learning rate of $10^{-4}$ and $(\beta_1, \beta_2) = (0.9, 0.999)$. We apply cosine learning rate decay with 500 warmup steps, use mixed-precision training (FP16), and set the batch size to 64. Each model is trained for 100 epochs.

\paragraph{Classification task} For the downstream task, we use a compact convolutional neural network. It consists of two convolutional layers with 3$\times$3 kernels and output channels 32 and 64, respectively, each followed by ReLU activation and max pooling. The output is flattened and passed through a fully connected layer with 512 units, followed by ReLU, a dropout layer with rate 0.25, and a final fully connected layer with 10 outputs. We train this classifier using the Adam optimizer with a learning rate of $10^{-3}$, a batch size of 64, and up to 20 epochs with early stopping based on validation performance. Cross-entropy loss is used for optimization.

\paragraph{Experimental protocol} For each configuration $(N, M)$, where $M$ denotes the number of synthetic samples to generate, we first train the conditional diffusion model on $\mathcal{D}_{\text{diff}}$. We then sample $M$ class-conditional synthetic images to form $\mathcal{D}_{\text{synth}}$. Two classifiers are trained: $f_{\text{real}}$ on $\mathcal{D}_{\text{train}}$ alone, and $f_{\text{aug}}$ on the augmented dataset $\mathcal{D}_{\text{train}} \cup \mathcal{D}_{\text{synth}}$. Both classifiers are evaluated on the same validation set $\mathcal{D}_{\text{val}}$ using the classification accuracy:
\begin{equation*}
    \text{Acc} = \frac{1}{|\mathcal{D}_{\text{val}}|} \sum_{(x,y)\in\mathcal{D}_{\text{val}}} \mathbb{I}\left(f(x) = y\right).
\end{equation*}

\paragraph{Hyperparameter configurations} We explore several synthetic-to-real data ratios $M/N \in \{0.25, 0.5, 0.75, 1.0, 1.25, 1.5, 2.0, 5.0\}$, with each configuration repeated across multiple random seeds. A summary of the hyperparameters is provided in \Cref{tab:hyperparams-cifar}. All experiments are implemented with the HuggingFace Diffusers library and executed on NVIDIA A100 GPUs with 40GB memory.

\begin{table}[ht]
\caption{Hyperparameter configurations for CIFAR-10 experiments}
\label{tab:hyperparams-cifar}
    \centering
    \begin{tabular}{lc}
        \toprule
        Parameter & Values \\
        \midrule
        Real data size ($N$) & 500 \\
        Synthetic-to-real ratio ($M/N$) & 0.5, 1.0, 3, 9 \\
        Diffusion steps ($T$) & 1000 \\
        Total noise variances & 0.0, 0.5, 1.0 \\
        \bottomrule
    \end{tabular}
\end{table}

\FloatBarrier

\subsection{Real-World Medical Imaging} \label{app:brain-mri}

In this section, we provide additional details for the experimental setup of \Cref{subsec:brain-experiments}.

\paragraph{Diffusion Model Training}
Our conditional diffusion model synthesises MRI slices conditioned on anatomical tissue masks. The architecture begins with a \(1 \times 1\) convolutional layer for initial feature projection, followed by a sinusoidal positional embedding to encode timestep information. The model includes four down-sampling stages, each consisting of two ResNet blocks, a linear attention layer with residual connection, and a \(3 \times 3\) convolutional down-sampling layer. This is followed by a bottleneck module comprising two additional ResNet blocks and another linear attention layer. The up-sampling path mirrors the down-sampling structure, replacing down-sampling layers with convolutional up-sampling layers of the same kernel size. Finally, a \(1 \times 1\) convolutional layer projects the features to the desired output channels.

The model follows a hierarchical channel structure, starting with 64 channels, doubling at each down-sampling stage (64 $\rightarrow$ 128 $\rightarrow$ 256 $\rightarrow$ 512 at the bottleneck), then halving symmetrically during up-sampling back to 64. Conditioning is achieved by concatenating a four-channel binary mask (GM, WM, CSF, lesion) with the timestep embedding and spatial inputs at the input layer. The model is trained on slices from the NeuroRx dataset using mean squared error loss over 600 denoising timesteps. All modalities (T1, T2, PD) are trained independently.

\paragraph{Segmentation Model and Task}
A vanilla U-Net is used for lesion segmentation. The network comprises three downsampling and upsampling layers with skip connections, ReLU activations, and max pooling. Feature channels double in the downsampling path (64 $\rightarrow$ 128 $\rightarrow$ 256) and halve in the upsampling path symmetrically. For all experiments, the models train for up to 800 epochs or until plateau, validated on a fixed NeuroRx set.

\paragraph{Hyperparameter Configuration} 
We perform a targeted grid search for hyperparameters considering our hardware constraints. The search space and the final configuration is shown in \Cref{tab:hyperparams-mri}. All experiments are executed on NVIDIA A100 GPUs with 40GB memory, with no gradient clipping or additional augmentations.

\paragraph{Computational Cost Comparison} The full sweep (used to simulate the behaviour of cross-validation) involves training across 8 different data compositions, with total training time across all runs adding up to approximately 222 GPU-hours. In contrast, following our theory requires estimating certain statistical quantities and evaluating the bound once. The cost of these operations is negligible compared to training, and only a single model needs to be trained at the predicted optimal ratio. Hence, our method can save up to 90\% of the training cost (1 model instead of 8), while still achieving near-optimal performance. The computational cost (training time) of running our experiments in shown in \Cref{tab:runtime}.

\begin{table}[ht]
\caption{Hyperparameter configurations for medical imaging experiments}
\label{tab:hyperparams-mri}
\centering
\begin{tabular}{ll}
\toprule
{Parameter} & {Values / Search Space} \\
\midrule
Batch size & 16, 32, 64, 128 (selected: 128) \\
Learning rate & \{1e-4, 5e-4, 1e-3\} (selected: 1e-4) \\
Epochs & Up to 800 or until loss plateau \\
Optimizer & Adam \\
LR Scheduler & Exponential decay ($\gamma = 0.99$) \\
Weight Init & Kaiming Uniform \\
Loss Function & Focal + Tversky loss (equal weight) \\
Focal Loss Params & $\delta = 0.25$, $\lambda = 2$ \\
Tversky Loss Params & $\alpha = 0.7$, $\beta = 0.3$ \\
Gradient Clipping & None \\
\bottomrule
\end{tabular}
\end{table}

\begin{table}[ht]
\centering
\caption{Computation time as a function of $N+M$.}
\label{tab:runtime}
\begin{tabular}{c c}
\toprule
\textbf{$N+M$} & \textbf{Time (hours)} \\
\midrule
4500  & $\sim$12 \\
5625  & $\sim$15 \\
6750  & $\sim$15 \\
9000  & $\sim$24 \\
13500 & $\sim$36 \\
22500 & $\sim$50 \\
40500 & $\sim$70 \\
\bottomrule
\end{tabular}
\end{table}

\subsubsection{Additional Results}\label{app:additional-experiments}
\begin{figure}[ht]
    \centering
    \includegraphics[width=.6\textwidth]{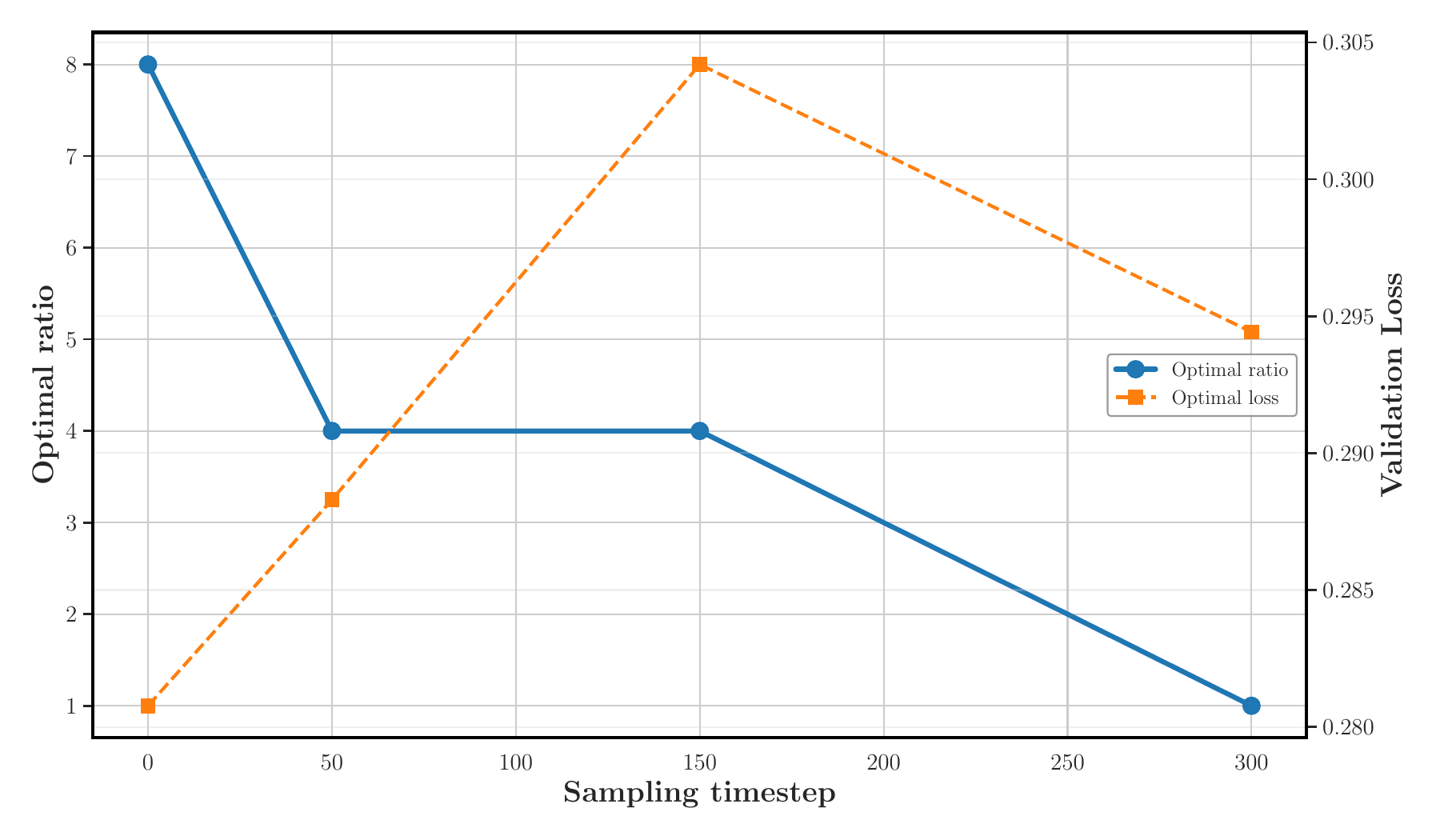}
    \caption{Optimal synthetic-to-real ratio (blue line) and the optimal validation loss (orange dashed line) as the distributional distance or equivalently the diffusion sampling timestep grows.}
    \label{fig:optimal-ratio-vs-timestep}
\end{figure}

From \Cref{thm:gen-gap-mixed}, we expect that as the quality of synthetic samples deteriorates, i.e., as the distributional distance between the synthetic data generator and the real data increases, the optimal synthetic-to-real ratio should decrease, placing greater emphasis on the real data. Consequently, we anticipate an increase in the validation loss. \Cref{fig:optimal-ratio-vs-timestep} empirically supports this expectation. In our setting, this distributional distance is modulated by the sampling timestep of the diffusion model: higher timesteps correspond to noisier, and thus less realistic, synthetic samples.

\begin{figure}[ht]
    \centering
     \begin{subfigure}[b]{0.49\textwidth}
         \centering
         \includegraphics[width=1.\textwidth]{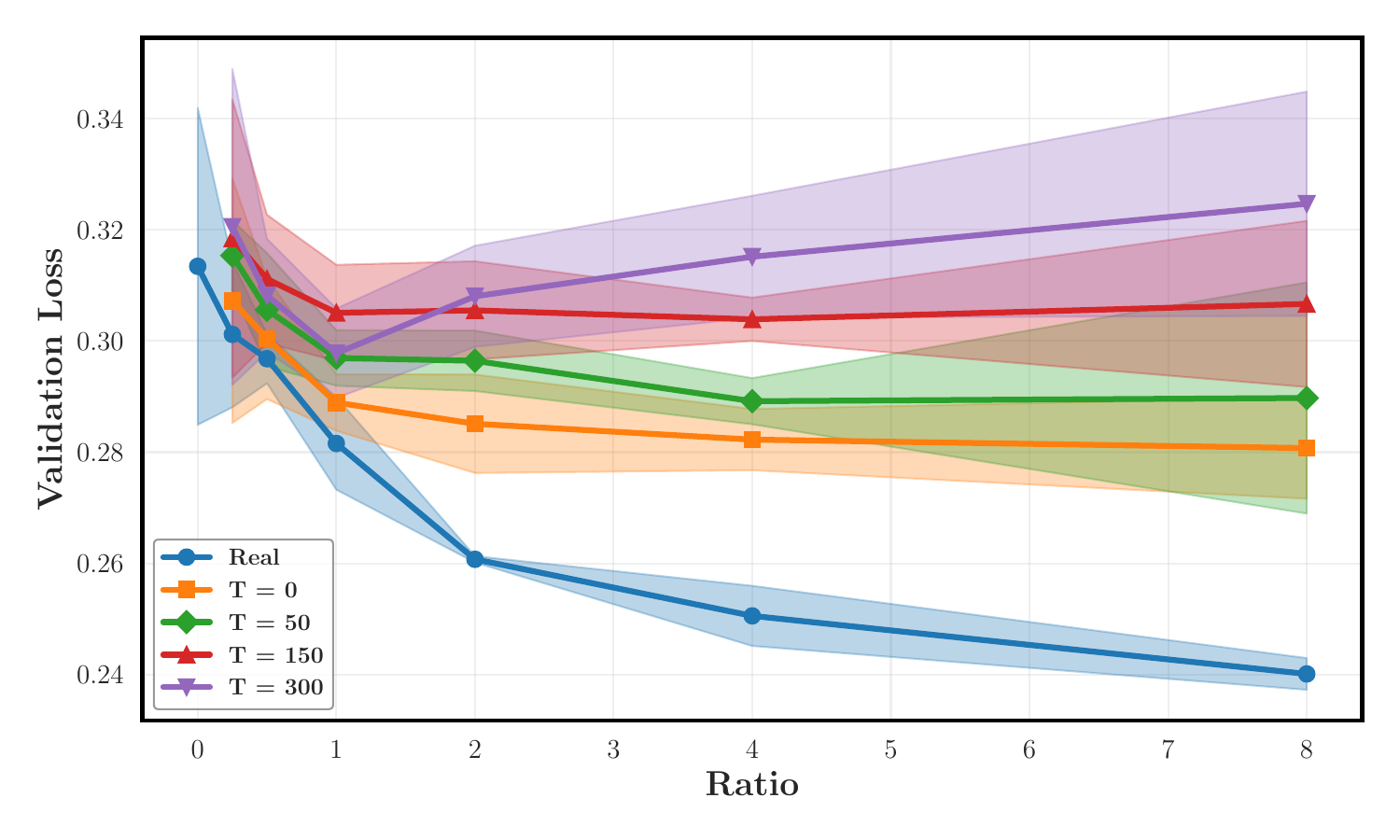}
         \caption{Validation loss vs. synthetic-to-real data ratio.}
         \label{fig:mean-loss-ratio}
     \end{subfigure}
     \hfill
     \begin{subfigure}[b]{0.49\textwidth}
         \centering
         \includegraphics[width=1.\textwidth]{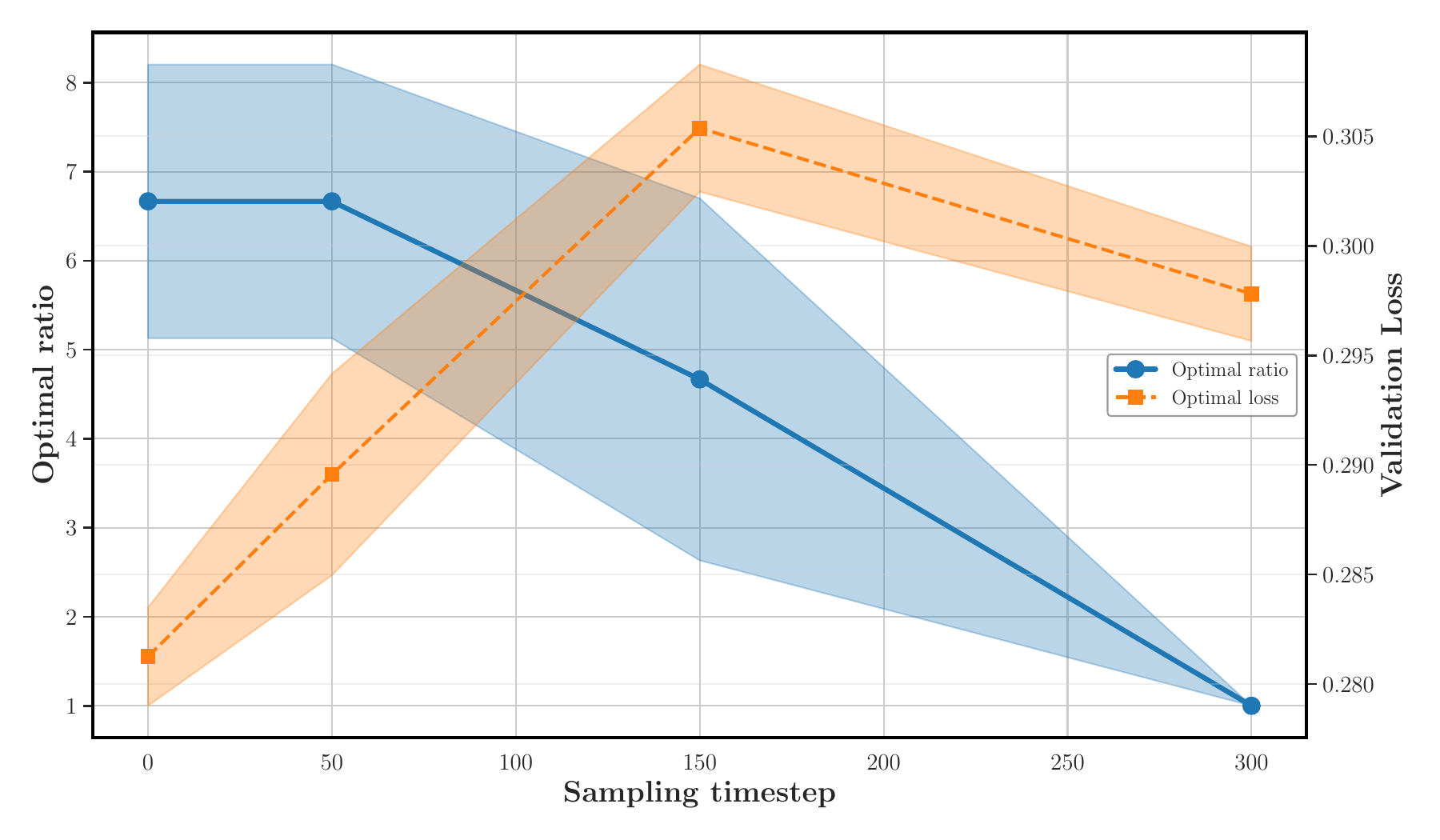}
         \caption{Optimal choice of ratio and final validation loss.}
         \label{fig:mean-optimal-choice}
     \end{subfigure}
     \vfill
     \begin{subfigure}[b]{0.49\textwidth}
         \centering
         \includegraphics[width=1.\textwidth]{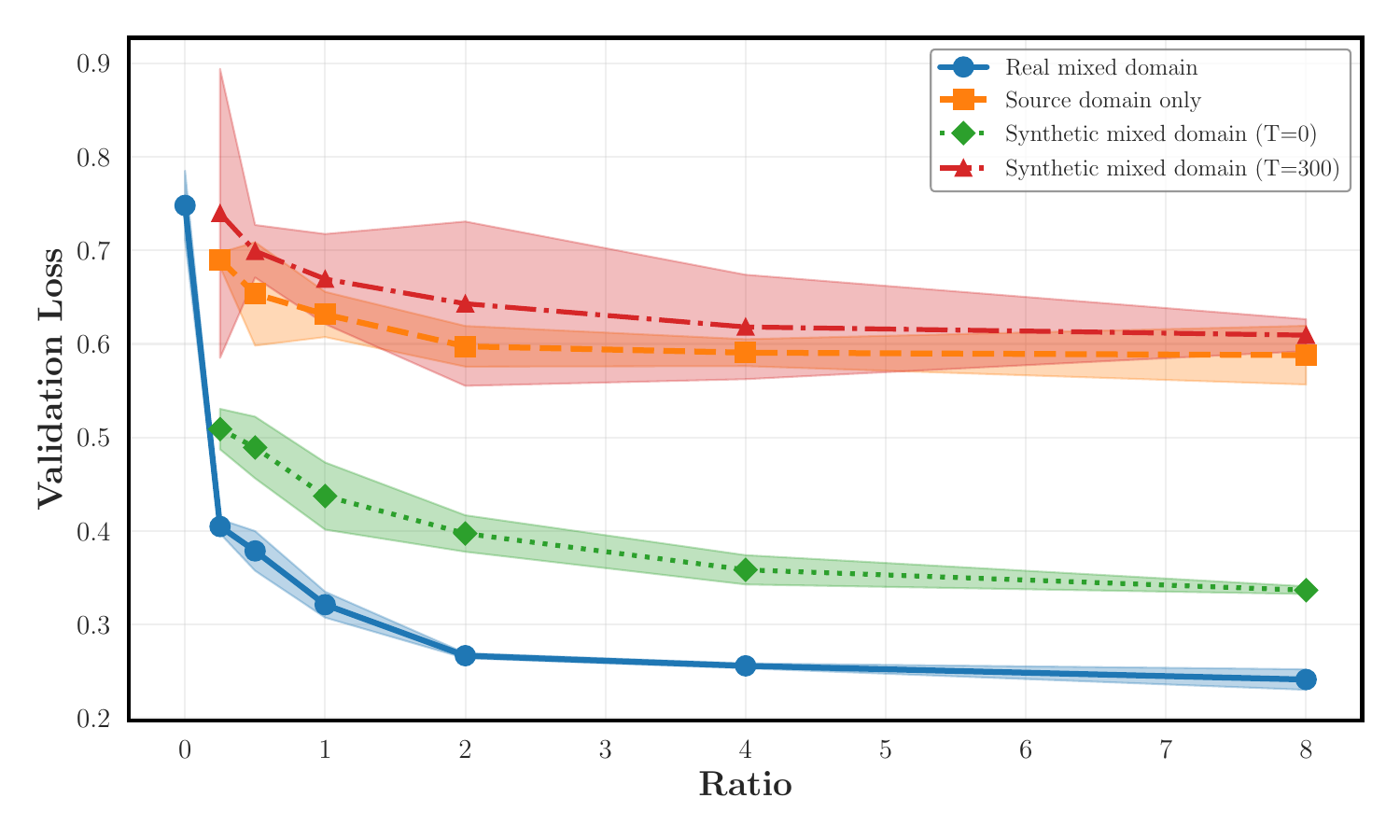}
         \caption{Validation loss vs ratio with domain shift.}
         \label{fig:mean-domain-shift}
     \end{subfigure}
    \caption{
        (a) Validation loss vs. synthetic-to-real data ratio across different sampling timesteps, representing varying distributional distances. We observe a sharper U-turn effect as the distributional distance increases. (b) The optimal synthetic-to-real data ratio decreases as synthetic samples become noisier or deviate further from the true distribution. (c) Incorporating synthetic data improves out-of-domain generalization when the synthetic data distribution is closer to the target than the source. All experiments are repeated with three different seeds. The results align with those shown in the main text for single experimental runs.
    }
    \label{fig:mean-plots}
\end{figure}

For reproducibility, we repeat each experiment using three different random seeds to account for variability introduced by stochastic elements in the training and sampling processes. The reported results in \Cref{fig:mean-plots} represent the mean performance across these runs, with corresponding confidence intervals to capture the variability. This approach ensures that our conclusions are not driven by a particular random initialization and provides a more robust estimate of model behavior.

\FloatBarrier

\subsection{Practical Insights}\label{app:practical-insight}

\begin{figure}[ht]
    \centering
     \begin{subfigure}[b]{0.99\textwidth}
         \centering
         \includegraphics[width=1.\textwidth]{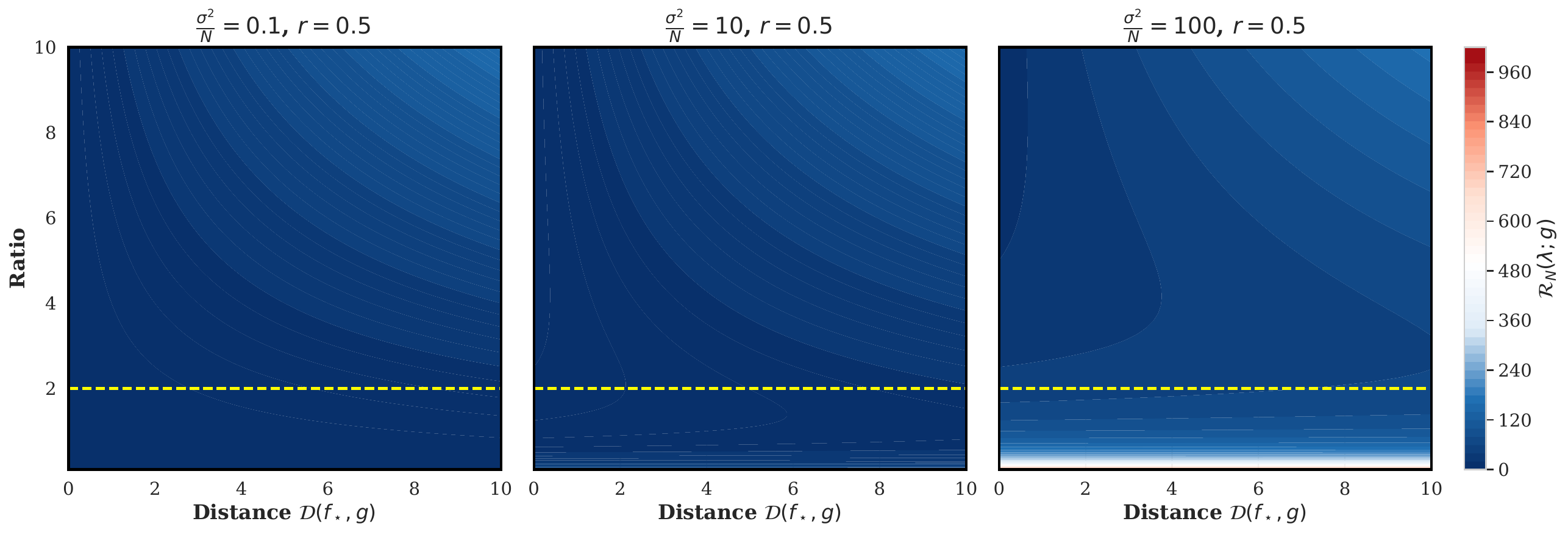}
         \label{fig:heatmap-r-0.5}
     \end{subfigure}
     \vfill
     \begin{subfigure}[b]{0.99\textwidth}
         \centering
         \includegraphics[width=1.\textwidth]{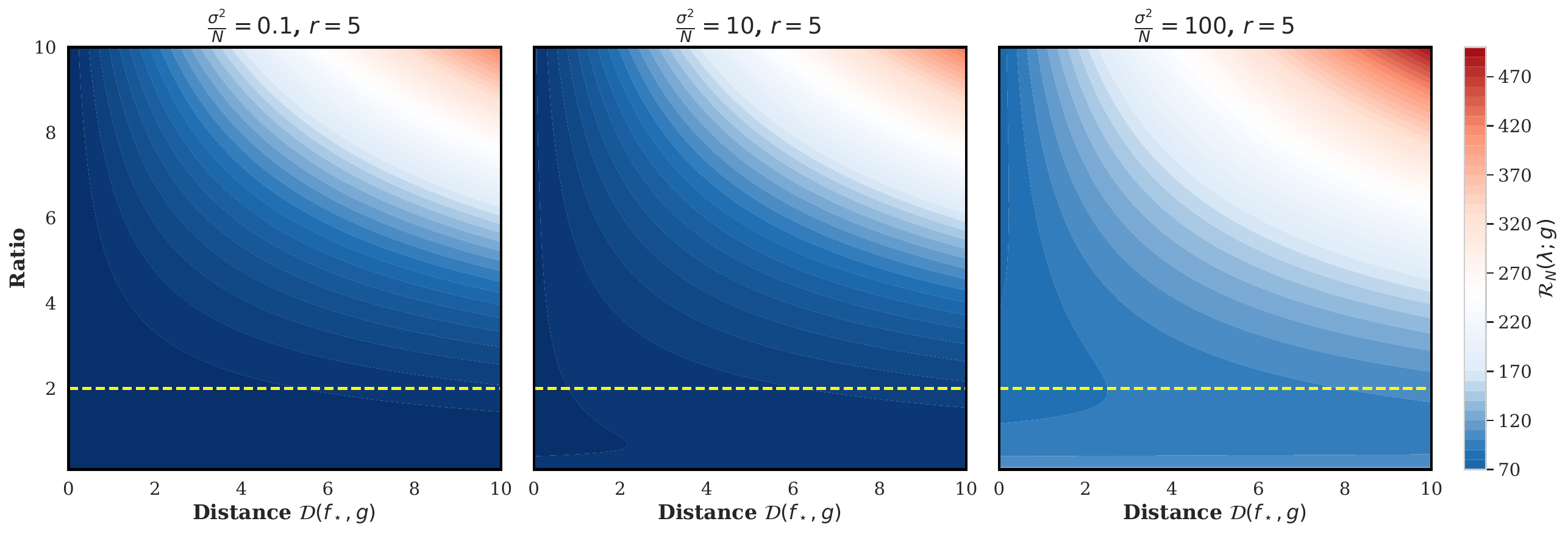}
         \label{fig:heatmap-r-5}
     \end{subfigure}
    \caption{
        Effect of signal-to-noise ratio on the choice of optimal synthetic-to-real data ratio, across two different values of $r \in \{0.5, 5\}$.
    }
    \label{fig:multiple-heatmap}
\end{figure}

In this section, we investigate the effects of signal-to-noise ratio, i.e., heterogeneity, and the regularity of the problem. As shown in \Cref{fig:multiple-heatmap}, varying the noise level across different values of $r$ exhibits a pattern consistent with our observations in \Cref{sec:practical}. We again find that a 1:2 ratio of real to synthetic data performs well across these scenarios. While changes in the regularity of the objectives (i.e.,~$r$) influence the scale of the test error, the overall behavior remains consistent.

\begin{figure}[ht]
    \centering
    \includegraphics[width=1.\textwidth]{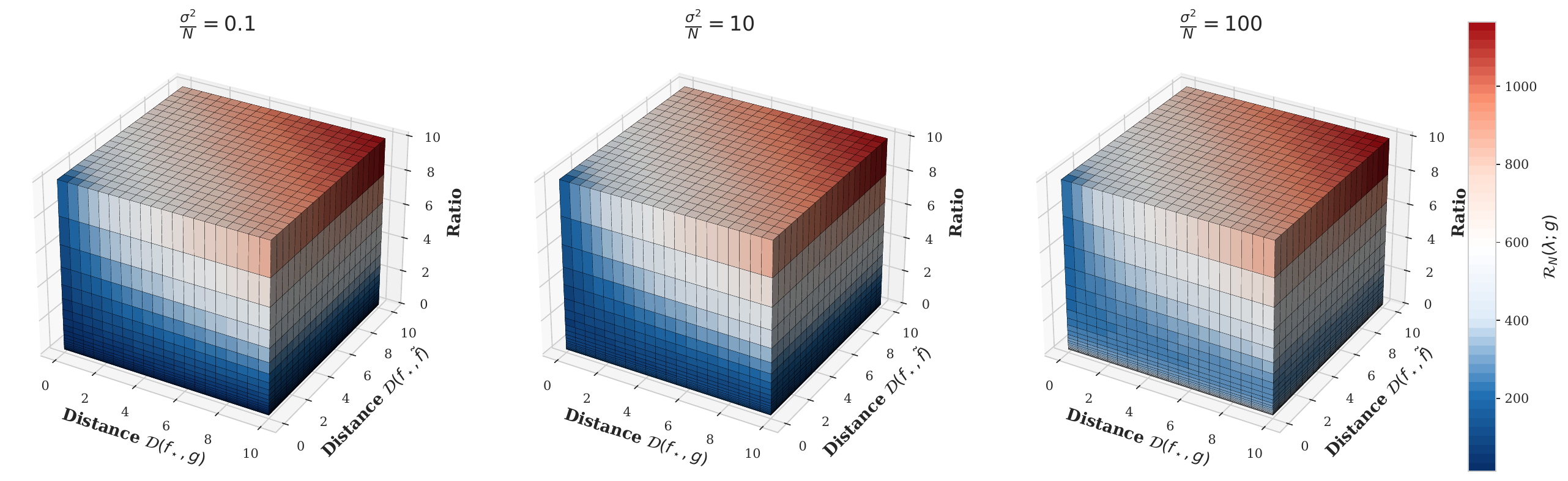}
    \caption{Effect of signal-to-noise ratio on the generalization error. All the plots are with $r = 1, \mu_\text{max} = 1$.
    }
    \label{fig:compare-3d}
\end{figure}

Similarly, under domain shift (see \Cref{fig:compare-3d}), the effect of the signal-to-noise ratio aligns with the in-domain behavior reported in \Cref{sec:practical}. Specifically, higher heterogeneity necessitates more careful selection of the real-to-synthetic ratio, as higher ratios can degrade performance when the distributional distance from the target domain is large.

\FloatBarrier

\section{EXPANDED RELATED WORK}\label{sec:expanded-related}

\paragraph{Synthetic Data} The rapid advancement of generative models has significantly improved data generation quality, making it increasingly difficult to distinguish between synthetic and real data. Many previous works \citep{DBLP:journals/corr/ZhangFZSA15,DBLP:journals/corr/abs-2410-22748,DBLP:conf/cvpr/ShrivastavaPTSW17,DBLP:journals/corr/abs-2408-14559,DBLP:journals/corr/abs-2007-08781,zhezherau2024hybrid} have demonstrated the effectiveness of synthetic data in enhancing the performance of deep learning methods and stabilizing training, both in supervised and unsupervised settings, through augmentation and various applications. \citet{DBLP:conf/iclr/AlemohammadCLHB24,DBLP:journals/nature/ShumailovSZPAG24,DBLP:journals/corr/abs-2311-16822} analyzed the effects of training generative models on synthetic data across multiple iterations, creating a self-consuming loop, and found that without sufficient fresh real data, model quality or diversity deteriorates over time, leading to model collapse. To further investigate this, \citet{DBLP:journals/corr/abs-2402-07712} studied model collapse theoretically in the regression case. Similarly, \citet{DBLP:conf/iclr/BertrandBDJG24,DBLP:journals/corr/abs-2404-01413} looked at iteratively training generative models on mixed datasets, concluding that stability is maintained if the initial model is accurate and the real data proportion is sufficiently high. \citet{DBLP:journals/corr/abs-2407-09499} expanded on this by investigating the impact of user-curated synthetic data on iterative retraining, framing it as an implicit preference optimization process and exploring its theoretical effects on model stability and quality. In contrast, our paper does not examine iterative learning but instead focuses on a single-step approach, framing synthetic data as a regularizer. Possibly closest to our work is \citet{DBLP:journals/corr/abs-2402-04376}, where the authors use a weighted empirical risk minimization approach to integrate surrogate data, reducing test error even when unrelated to the original data. However, their work differs from ours in two main aspects: (1) they focus on the scaling law of the test error, while we aim to determine the optimal synthetic-to-real data ratio or regularizer weight; (2) they do not account for the distance between synthetic and real data distributions, while our work specifically provides a bound based on this difference.

\paragraph{Optimal Mixing and Model Collapse} While the high-level U-shaped behavior of test error with respect to the proportion of synthetic data is intuitively expected, our contribution lies in deriving it formally via a distinct analysis route: (i) a tractable kernel-regression formulation yielding the optimal ratio, and (ii) a general stability-based bound incorporating Wasserstein distances, which to our knowledge has not appeared previously. Several concurrent or recent works study related phenomena from complementary perspectives. \citet{zhang2023mixture} focuses on self-consuming generative models and out-of-distribution generalization of mixture data; the setting differs from our mixed-data generalization analysis, which considers a single-step ERM scenario with a fixed pool of synthetic samples rather than recursive retraining. We highlight conceptual overlaps in studying the limits of synthetic data, but the assumptions differ substantially (e.g., their recursive training loop vs.\ our single-round ERM formulation). \citet{garg2025preventing} and \citet{he2025golden} both derive optimal mixing effects but rely on different technical frameworks (e.g., asymptotic kernel limits, hierarchies of synthetic corruption models). Crucially, these works do not provide general stability-based bounds, which is where our contribution is novel and complementary. We view these concurrent results as reinforcing the importance of the optimal-ratio question while highlighting the diversity of proof techniques that can be brought to bear on it.

\paragraph{Noise Injection, Data Augmentation, and Distributional Mismatch}
A natural question is how our framework relates to the classical literature on training with noise and data augmentation. Prior works \citep{bishop1995training,lin2022good,shen2023balancing} mainly analyze perturbations of inputs or labels under the assumption that the augmented data is sampled from the same underlying distribution. In contrast, our setting explicitly considers a distinct synthetic distribution, with a potentially non-zero Wasserstein distance from the real data. This structural domain shift materially changes the generalization behavior and is the key modelling distinction of our work. Concretely, while \citet{bishop1995training} shows that noise training induces Tikhonov regularisation, it does not derive an optimal mixing ratio nor account for distribution mismatch. Similarly, \citet{lin2022good} and \citet{shen2023balancing} study benign overfitting or domain adaptation but do not provide an explicit dependence on the distributional distance between sources. Among these, \citet{shen2023balancing} is the closest to our work. However, although that paper provides generalization bounds and discusses the bias--variance tradeoff, it focuses primarily on aligning multi-source data to the target distribution in the unsupervised setting, where the goal is to leverage additional sources of feature space and label them according to their similarity or dissimilarity to the source data. In contrast, we explicitly introduce a tradeoff that depends on the distributional distance and derive an explicit interior optimum for the mixing ratio. More broadly, while bias--variance ideas appear throughout the literature, our main contribution is a general algorithmic-stability bound that yields an analytic form for the tradeoff, explicit dependence on the Wasserstein distance, and applicability to general learning algorithms---not only kernel methods or linear models. This allows us, for example, to show that the U-curve sharpens with distributional mismatch, a phenomenon that does not arise in classical noise-injection analyses.

\paragraph{Mixing Data Sources} Recent work has investigated scaling laws for mixing data sources in pretraining objectives \citep{ye2025data,shukor2025scaling,pmlr-v139-hashimoto21a}, showing how performance depends on the proportions of different datasets. These studies typically require either running a large suite of experiments with varying source mixtures or access to massive datasets that can be systematically subsampled to empirically fit the law. In contrast, our work is motivated by applications such as healthcare and medical imaging, where data scarcity makes such large-scale experimentation infeasible. Methodologically, our approach provides an a priori estimate of the optimal synthetic-to-real data ratio, eliminating the need to empirically fit a scaling law. From a theoretical perspective, our analysis also departs substantially from existing work: whereas prior studies rely on empirical findings and use theoretical insights primarily for justification, we adopt a stability-theoretic framework that enables a precise characterisation of the trade-off and yields an explicit formula for the optimal ratio that practitioners can apply directly. Finding the optimal mixture of data sources is also closely related to identifying which sources are most \emph{useful} for downstream performance. For example, \cite{thudi2025mixmin} approaches the problem via a bi-level optimisation framework rather than empirical scaling laws, and analyses how the benefits of their method evolve with increasing model size. Similarly, \cite{firdoussi2025synthetic} applies data pruning based on a score function to select high-quality samples, thereby improving the effectiveness of mixed data. Our work differs from these approaches in that we do not attempt to select or filter the synthetic data based on its quality. Instead, we assume the synthetic data is given and focus on how to combine it with the available real data in an optimal way. This perspective is particularly relevant in practical scenarios where the synthetic data generation process is fixed or externally provided, and the key challenge lies in determining how best to exploit it in conjunction with scarce real data.

\paragraph{Domain Adaptation and Transfer Learning} Recent research has explored the intersection of domain adaptation and synthetic data, showing how synthetic data can bridge the gap between source and target domains, thereby enhancing model transferability and generalization across tasks \citep{DBLP:journals/corr/abs-2302-04149,DBLP:journals/corr/abs-1806-09755,DBLP:conf/case/ImbuschSB22,DBLP:journals/corr/abs-2010-06028}. A major challenge in transfer learning is the distribution gap, and several studies address this by using synthetic data to fine-tune models, improving generalizability \citep{DBLP:conf/cvpr/MishraPPCKSSF22,DBLP:conf/cvpr/KimSPC20,DBLP:conf/cvpr/SariyildizALK23}. \citet{DBLP:journals/mlst/GeraceSMSZ22} propose synthetic data as a framework for modeling correlations between datasets, showing improvements in generalization when transferring learned features from source to target tasks. On a more theoretical level, several works connect domain adaptation to distributionally robust learning, demonstrating that adding unlabeled or labeled data improves generalization; these setups can be easily extended to include synthetic data \citep{DBLP:journals/corr/abs-2410-14061,DBLP:conf/iclr/Saberi0HMMK24, wu2022generalization, DBLP:journals/jmlr/HouKK0R23}. Our perspective differs in that we do not view synthetic data as a replacement or proxy for domain adaptation, but rather as a complementary source whose utility depends critically on how it is balanced with real data. This focus on the optimal ratio naturally connects to knowledge distillation \citep{DBLP:journals/corr/HintonVD15,DBLP:conf/nips/StantonIKAW21,DBLP:journals/corr/abs-2005-10419,busbridge2025distillation}, where the objective similarly involves balancing two heterogeneous sources of supervision: the labelled data (analogous to real samples) and the teacher’s soft signal (analogous to synthetic information). Just as we provide an explicit characterisation of the optimal real-to-synthetic ratio, one can view distillation through the same lens of optimally weighting complementary signals, highlighting a deeper connection between synthetic data integration and distillation-based learning.

\paragraph{Generalization Bounds} The first generalization bounds were based on characterizations of the hypothesis space's complexity, such as the VC dimension or Rademacher complexity \citep{DBLP:conf/ac/BousquetBL03, DBLP:books/daglib/0026015, DBLP:books/daglib/0033642}. However, due to their algorithm-independent nature, these bounds must hold even for the worst algorithm within a given hypothesis space, making them often inadequate for modern over-parameterized neural networks, where the complexity measure typically scales exponentially with the architecture's depth \citep{DBLP:books/daglib/0025992, DBLP:conf/iclr/ZhangBHRV17, DBLP:conf/icml/BelkinMM18}. To address this issue, recent approaches focus on providing algorithm-dependent generalization bounds. The underlying intuition is that a hypothesis less dependent on the input dataset is less prone to overfitting and, therefore, generalizes better. Among the results building on this idea are bounds based on uniform stability \citep{DBLP:journals/jmlr/BousquetE02,DBLP:conf/colt/AttiaK22}, differential privacy \citep{DBLP:journals/fttcs/DworkR14}, PAC-Bayesian bounds \citep{DBLP:journals/corr/abs-1901-05353, DBLP:conf/colt/McAllester99}, information-theoretic bounds \citep{DBLP:journals/tit/RussoZ20,DBLP:journals/corr/abs-2010-10994,DBLP:conf/nips/HaghifamNK0D20}, and chained bounds \citep{DBLP:conf/colt/ClericoSDD22,DBLP:conf/nips/AsadiAV18}. Our work mainly uses the previously established ideas of generalization bounds for mixed of real and synthetic data when the synthetic data acts as regularizer. More related to our work and on the importance of regularization, \citet{DBLP:conf/nips/LiZ21} analyzes the generalization properties of fine-tuning in transfer learning and proposes a PAC-Bayes generalization bound, combining regularization and self-labeling. \citet{DBLP:conf/icml/MouZGW18} provides generalization guarantees for dropout training by bounding the error using offset Rademacher complexities, capturing data-dependent regularization and the effect of perturbation variance.

\section{NOTES ON THE CHOICE OF DISTANCE METRIC}

In \Cref{sec:practical} we discussed possible ways of estimating the Wasserstein distance for practical scenarios. However, one setup that happens in many practical cases particularly in healthcare domain is heterogeneous tabular data. Indeed, Wasserstein implicitly assumes a meaningful ground metric on the input space, which can be problematic when the data contains a mixture of categorical, ordinal, and continuous features, since arbitrary encodings or feature scalings can dominate the geometry. One possible direction is to first obtain a more homogeneous representation of tabular inputs using an unsupervised or semisupervised embedding (e.g., contrastive encoders \citep{DBLP:conf/icml/ChenK0H20,DBLP:conf/iccv/CaronTMJMBJ21,DBLP:conf/iclr/ShidaniHRWDB24} or variational embeddings \citep{DBLP:journals/corr/KingmaW13}) and then apply the Wasserstein-based bound in this learned feature space. Another direction is to replace Wasserstein with discrepancies better suited to tabular data. Total variation (TV) is appealing for categorical or mixed tabular features because it requires no ground metric and tightly controls differences in expectations. However, TV quickly becomes too coarse and especially hard to estimate. This motivates using alternatives like Maximum Mean Discrepancy (MMD) with characteristic kernels (robust to mixed feature types and statistically favorable in high dimension) \cite{DBLP:journals/jmlr/GrettonBRSS12}. These alternatives avoid imposing a single global geometry and often yield discrepancy measures more faithful to the structure of tabular data. Systematically studying how our stability-based generalization bound behaves under TV or these alternative divergences is an interesting future direction.

\end{document}